%% file: main.tex
\documentclass[10pt,twocolumn,letterpaper]{article}
\usepackage{cvpr}
\usepackage{xspace}
\input{preamble}

\usepackage{amsmath,amssymb}
\usepackage{booktabs}
\usepackage{multirow}
\usepackage{enumitem}
\usepackage{graphicx}
\usepackage{tikz}
\usepackage{colortbl}

\definecolor{cFus}{rgb}{0.56,0.24,0.44}
\definecolor{cJoint}{rgb}{0.00,0.40,0.64}
\definecolor{cOld}{rgb}{0.38,0.40,0.44}
\colorlet{oursbg}{cFus!10}
\colorlet{bandbg}{black!7}
\newcommand{\chip}[2]{\tikz[baseline=(c.base)]\node[fill=#1,text=white,rounded corners=1.4pt,inner xsep=2.6pt,inner ysep=1.1pt,font=\scriptsize\bfseries](c){#2};}
\newcommand{\lchip}[2]{\tikz[baseline=(c.base)]\node[draw=#1,line width=0.4pt,fill=white,text=#1,rounded corners=1.4pt,inner xsep=2.4pt,inner ysep=0.9pt,font=\scriptsize\bfseries](c){#2};}
\newcommand{\famsq}[1]{\textcolor{#1}{\rule[0.1ex]{0.85ex}{0.85ex}}\hspace{0.45em}}
\newcommand{\second}[1]{\underline{#1}}
\newcommand{\grule}{\arrayrulecolor{black!30}\specialrule{0.3pt}{1.0pt}{1.0pt}\arrayrulecolor{black}}

\definecolor{cvprblue}{rgb}{0.21,0.49,0.74}
\usepackage[pagebackref,breaklinks,colorlinks,allcolors=cvprblue]{hyperref}

\newcommand{\figfile}[3]{\IfFileExists{figures/#1.pdf}{\includegraphics[width=#2]{figures/#1.pdf}}{\fbox{\parbox[c][#3][c]{\dimexpr#2-2\fboxsep-2\fboxrule\relax}{\centering\footnotesize\texttt{figures/#1.pdf}}}}}

\def\confName{CVPR}
\def\confYear{2027}

\title{Send Probabilities, Not Answers:\\False Alarms Compound in Multi-Camera VLM Agent Teams}

\author{Yu Chang\textsuperscript{1} \quad Anzhe Cheng\textsuperscript{1} \quad Jiahao Chen\textsuperscript{2}\\
Shahin Nazarian\textsuperscript{1} \quad Paul Bogdan\textsuperscript{1}\\[3pt]
\textsuperscript{1}University of Southern California \\ \quad \textsuperscript{2}University of Toronto}

\begin{document}
\maketitle
\begin{strip}
\centering
\vspace{-16pt}
\figfile{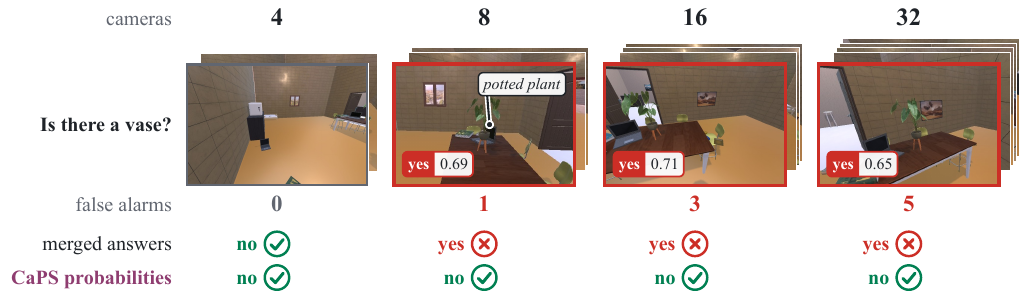}{\textwidth}{5.1cm}
\vspace{-6pt}
\captionsetup{hypcap=false}
\captionof{figure}{\textbf{More cameras, more false alarms.}
A team of VLM agents, one per camera, is asked ``Is there a vase?'' about a room that has none; each photograph tops a stack of the team's other views.
Some agents take the potted plant for a vase and say yes (red frame; the tag gives that agent's $P(\text{yes})$).
Merging the yes/no answers by an OR, the team wrongly answers yes from eight cameras on; if the agents send their probabilities (\method), these lukewarm yeses weigh little and it answers no at every size.
Over 260 rooms with 32 cameras, the OR of the answers reaches 73.3\% balanced accuracy, the best fitted rule on the same answers 92.2\% and \method 96.6\% (Qwen3-VL-8B).}
\captionsetup{hypcap=true}
\label{fig:teaser}%
\end{strip}
\input{sec/0_abstract}
\input{sec/1_intro}
\input{sec/2_related}
\input{sec/3_problem}
\input{sec/4_method}
\input{sec/5_experiments}
\input{sec/7_conclusion}
\input{sec/8_llm_statement}
{
    \small
    \bibliographystyle{ieeenat_fullname}
    \bibliography{main}
}

\input{supp}

\end{document}

%% file: preamble.tex
\usepackage{cuted} 

\usepackage{currfile} 

\usepackage{caption} 

\newcommand{\method}{CaPS\xspace}   



%% file: sec/0_abstract.tex
\begin{abstract}
In multi-view scene understanding, multi-agent systems built on vision-language models (VLMs) mostly exchange answers or text reports, and what their agents should communicate as cameras are added remains largely unexamined.
We show that with such messages, false alarms compound with every added camera, since any agent's mistaken yes can mislead the team.
In rendered rooms, a team that trusts any yes falls from 79.9\% to 46.6\% accuracy on absent objects between 4 and 32 cameras; a stricter prompt, debate, a verifier or an aggregator does not stop the fall.
The same model given all views in one prompt, a joint call, shows no such fall.
A yes/no answer is the agent's probability cut at one half, so the margin between a confident and a marginal yes is gone before the team sees it.
Following distributed detection theory, we propose Calibrated Probability Sharing (\method), in which each agent sends its probability of yes and the team fuses these probabilities with a threshold fitted for its own size, on labelled scenes or, without labels, on a joint call's answers.
In the rooms, a \method team of 32 cameras reaches 96.6\% balanced accuracy and 98.9\% accuracy on absent objects, and on object existence across the rooms, ScanNet scans, Argoverse~2 driving scenes and the public 3D-POPE benchmark, pooled over camera counts, \method comes within 1.2 points of the joint call for five VLMs and surpasses it for a sixth.
\end{abstract}

%% file: sec/1_intro.tex
\section{Introduction}
\label{sec:intro}

\begin{figure*}[t]
\centering
\figfile{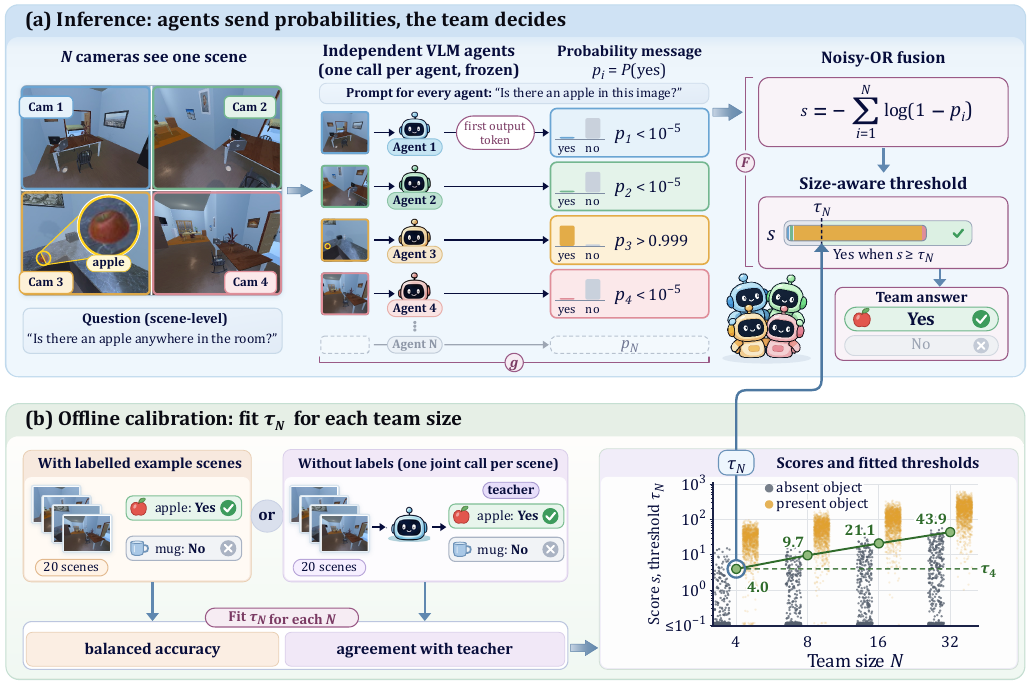}{\textwidth}{11.6cm}
\vspace{-12pt}
\caption{\textbf{Overview of \method: the team sets the cut for its size.}
(a)~Each camera's frozen VLM agent sends one number, its probability of yes $p_i$ read from the first output token (real outputs of Qwen3-VL-8B for an apple that only the third of four cameras sees); the team fuses them into a noisy-OR score $s$ and answers yes if $s \geq \tau_N$.
(b)~$\tau_N$ is fitted offline per team size, on labelled scenes or a joint call's answers (twenty scenes suffice); on Thor-PV, absent-object scores (grey) climb past a kept $\tau_4$ (dashed) as $N$ grows, and the fitted $\tau_N$ follows them.}
\label{fig:overview}
\vspace{-8pt}
\end{figure*}

Vision-language models (VLMs) have demonstrated strong capabilities in recognizing, grounding and reasoning about the content of an image~\cite{bai2025qwen3vl,li2024llavaov,zhu2025internvl3}.
However, many real-world scenes are observed by several cameras at once, and understanding them requires combining evidence across views~\cite{dai2017scannet,wilson2021argoverse2,hou2020mvdet}.
While one VLM can be given every view in a single prompt, such a \emph{joint call} grows with each added camera and requires all images to be gathered at one model.
Multi-agent systems address these challenges by giving each agent its own view and letting the agents exchange short messages instead of images~\cite{zhang2024coela,mandi2024roco,sikarwar2026communicating,parekh2026gptnt}.
In this work, we focus on teams of frozen VLM agents, one per camera, that decide whether an object is present in a scene and how many there are.

What the agents of such a team should communicate, however, has received little attention.
Existing systems exchange answers or text reports on the assumption that every added camera adds evidence, yet we find that with such messages a team becomes less accurate as cameras are added.
In a room without a vase, for instance, five of 32 agents answer yes, so a team that trusts any yes is wrong (Fig.~\ref{fig:teaser}), whereas the joint call of the same frozen model answers no.
Each added agent thus brings another chance of a mistaken yes, and false alarms compound with the size of the team, at a rate that a beta-binomial model fitted on four cameras forecasts (Fig.~\ref{fig:law}).

Existing designs~\cite{xu2026a4vl,chen2025lvagent,chen2026videochatm1,zhang2025gam,sivakumaran2025dart} add machinery on top of the answers, from votes and verification calls to text aggregation and debate, yet in our tests each of these message patterns falls short as the team grows.
\textbf{Voting} suppresses false alarms but silences a \emph{lone witness}, an object that only one camera sees.
\textbf{Verification} passes the views that said yes to a further call, which removes only part of the false alarms.
\textbf{Aggregation}~\cite{wang2025moa,tavella2026moreviews} lets a text-only model merge the agents' reports, but this model mostly follows any report that mentions the object.
\textbf{Debate}~\cite{du2024improving,liang2024divergent} lets agents revise their answers after reading their peers', yet the remaining false alarms still accumulate.

The main reason is that these designs operate on answers.
Since a yes/no answer is the agent's probability cut at one half, whatever the size of the team, the evidence that separates a confident yes from a marginal one is discarded before the team sees it.
A team is left with a tension between sensitivity and reliability, since trusting any yes keeps the lone witness but accumulates false alarms, whereas demanding agreement removes them but loses the lone witness.

We therefore let the agents communicate how confident they are and leave the decision to the team.
Each agent sends its probability of yes, read from its first output token at no additional cost, and the team fuses these probabilities under a threshold that it sets for its own size.
A threshold that rises with the team absorbs the growing number of marginal yeses.
This formulation follows distributed detection theory, in which soft local decisions preserve information that hard ones discard~\cite{tenney1981detection,chair1986optimal,viswanathan1997distributed,drakopoulos1991optimum,willett2000good}.

Building on this idea, we propose \emph{Calibrated Probability Sharing} (\method), which fuses the agents' probabilities by a noisy-OR and answers yes when the fused score reaches a threshold fitted for the team size.
The threshold is fitted on labelled scenes or, without labels, on a joint call's answers, and a one-bit variant, the \emph{calibrated bit}, applies it to each agent's probability, so that the team still exchanges one bit per camera and a single confident witness can still carry the decision.
\method requires no training and, for existence, no additional calls or tokens, and it works with VLMs from four model families.
ReConcile~\cite{chen2024reconcile} also sends a verbal confidence that weights a vote, whereas \method fuses first-token probabilities and moves the threshold to the team.

In rendered rooms with 4 to 32 cameras, a \method team stays level with the joint call as cameras are added, while teams that trust any yes fall (Table~\ref{tab:thor}), and it remains level with the joint call on real ScanNet scans, Argoverse~2 driving scenes and the public 3D-POPE benchmark~\cite{yang20253dgrand}.
Each of its calls reads a single image, 0.8k prompt tokens against the joint call's 24.7k at 32 cameras, and the calls can run in parallel.
Of the systems above, only COSMIC~\cite{sikarwar2026communicating}, with two views, compares a team with the joint call.

Our core contributions are:
\begin{itemize}[leftmargin=*,nosep]
\item A study showing that VLM agent teams that trust any yes accumulate false alarms with every added camera, with a beta-binomial model that forecasts the decay from four cameras.
\item \method, a probability-sharing protocol whose threshold is fitted for the team size, without training and optionally without human labels, with a one-bit variant that keeps the lone witness.
\item Benchmarking against the joint call on three testbeds and the public 3D-POPE benchmark, where \method comes within 1.2 points of it on object existence for five VLMs and surpasses it for a sixth, pooled over camera counts.
\end{itemize}

%% file: sec/2_related.tex
\section{Related Work}
\label{sec:related}

\paragraph{Multi-agent communication.}
LLM-based multi-agent systems let agents debate~\cite{du2024improving,liang2024divergent} or stack their answers in layers~\cite{wang2025moa}, and what the agents exchange is almost always text.
A few works send more than an answer, as ReConcile~\cite{chen2024reconcile} attaches a verbal confidence that weights a vote, DebUnc~\cite{yoffe2025debunc} passes uncertainty estimates through the prompt or the attention weights, and CIPHER~\cite{pham2024cipher} exchanges embeddings.
Later analyses find that majority voting accounts for most of the gain of debate~\cite{choi2025debate}, that agents abandon correct answers after reading their peers~\cite{wynn2025talk} and pool distributed facts poorly~\cite{li2026hiddenbench}, and that the gains are small against self-consistency~\cite{wang2023selfconsistency} once cost is counted~\cite{li2026m3mad,illusion2026}; more calls can help a vote~\cite{li2024moreagents} or hurt it~\cite{chen2024morecalls}.
In all of these studies the evidence reaches every agent as text, whereas in our setting each agent must extract it from its own image.

\paragraph{VLM agent teams.}
Video systems split a long video over agents that rate one another and prune the lowest-rated~\cite{xu2026a4vl,chen2025lvagent}, or that vote and hand their reports to an organizer~\cite{chen2026videochatm1,kugo2025videomultiagents}, while image systems show every agent the same picture and add uncertainty weights, reflectors, tools or evidence checks~\cite{zhang2025gam,cherian2025wise,sivakumaran2025dart,wang2026seeing}.
Closest to our setting, COSMIC~\cite{sikarwar2026communicating} gives two agents private views and finds one model with both views ahead of their dialogue, and Tavella~\etal~\cite{tavella2026moreviews} report that merging per-view captions with an LLM is brittle.
However, none of these works varies the number of views or asks what the agents should send.
We instead scale the team from 4 to 32 cameras and compare message types, with their decision rules as baselines.

\paragraph{Distributed detection.}
Fusing the local decisions of sensors is a classical problem~\cite{tenney1981detection,varshney1997distributed,viswanathan1997distributed}, for which Chair and Varshney~\cite{chair1986optimal} derived the optimal fusion of independent hard decisions and Tenney and Sandell~\cite{tenney1981detection} showed that the optimal local thresholds depend on the rest of the team, while correlation among the sensors changes both the rule and how performance scales with their number~\cite{drakopoulos1991optimum,willett2000good}.
Classifier combination draws the same line between fusing labels and fusing scores~\cite{kittler1998combining}, correlated voters weaken the gains of majority voting~\cite{ladha1992condorcet,kuncheva2003measures}, and the noisy-OR is the usual pooling in multiple-instance detection~\cite{viola2005milboost}.
\method builds on these results and measures what the theory leaves open: how much of a joint call's evidence the first-token probabilities of a frozen VLM retain, and whether VLM team designs inherit the failure of the OR.

\paragraph{Multi-image VLMs.}
Benchmarks of cross-view matching and counting find multi-image reasoning hard for a single model~\cite{yang2026mmsi,yeh2026allangles,shah2026crossview,zhang2026multiview,lee2025spatialmosaic,das2026mimic}, and VLMs over-answer yes when probed for objects~\cite{li2023pope,yang20253dgrand}, much as text-to-image models omit or confuse requested objects~\cite{chang2026maskattnsdxl}.
Multi-view recognition fuses per-view features, often with camera geometry~\cite{su2015mvcnn,hou2020mvdet}, and Coates and Ng~\cite{coates2010multicamera} fused per-camera detector probabilities by adding log odds; our agents get no geometry.

%% file: sec/3_problem.tex
\section{Problem Formulation}
\label{sec:problem}

We consider a scene that is observed by $N$ cameras and a question $q$ about the scene as a whole, such as ``Is there a mug anywhere in the room?''.

\noindent\textbf{Definition 1 (Camera team).}
A \emph{team} gives each image $x_i$ to its own agent, one call of a frozen VLM, which returns a message $m_i = g(x_i, q)$, and the team answers $\hat{y} = F(m_1, \dots, m_N)$ from the messages alone.
The \emph{joint call} instead gives all $N$ images to the same model in a single prompt, and it serves as our reference for what the model can extract from the scene.

A team is thus specified by its message $g$ and its fusion rule $F$.
Existing teams let every agent send an answer or a text report and merge these messages by an OR, a vote or a further model call.
Our goal is a team that matches the joint call while its calls remain small and independent, which raises three challenges:
\begin{itemize}[leftmargin=*,nosep]
\item \textbf{Compounding false alarms.} An agent occasionally answers yes for an object that is absent, and since every added camera brings another chance of such a mistake, a rule that trusts single agents degrades as $N$ grows.
\item \textbf{Lone witnesses.} An object may be visible to only one of the $N$ cameras, so a rule that demands agreement among agents discards the very evidence only a team can add.
\item \textbf{Communication cost.} The appeal of a team lies in short messages and independent calls, which dialogue rounds, further calls or longer reports give up.
\end{itemize}

\paragraph{A model of compounding false alarms.}
The first challenge can be quantified: if the $N$ views flagged an absent object independently with probability $f$, an OR team would be right with probability $(1-f)^N$.
In practice some objects and scenes are hallucinated more easily than others, so the flags of one question are correlated, which we capture with a beta-binomial model~\cite{skellam1948probability} in which the flag rate $p$ of a question is drawn from $\mathrm{Beta}(\alpha, \beta)$ and the views flag independently given $p$.
The team then raises no false alarm with probability
\begin{equation}
P(\text{no view flags}) = B(\alpha, \beta+N)\,/\,B(\alpha, \beta),
\label{eq:law}
\end{equation}
where $B$ is the beta function.
Eq.~\eqref{eq:law} decays more slowly than the exponential of independent views, yet it still decays, so that under this model correlation among the views delays the failure of an answer-merging team without preventing it.
Fitted on the flags of four cameras, it also forecasts larger teams (Sec.~\ref{sec:analysis}).

%% file: sec/4_method.tex
\section{Method}
\label{sec:method}

To address these challenges, we propose \emph{Calibrated Probability Sharing} (\method), a communication protocol in which every agent sends its probability of yes and the team makes the decision.
Rather than letting each agent commit to an answer at a threshold of one half, \method moves the threshold to the team, where it can follow the team size.

\subsection{Overview}
\label{sec:overview}
As illustrated in Fig.~\ref{fig:overview}, \method replaces the message $g$ and the fusion rule $F$ of a team and leaves the agents, their prompts and, for existence, the number of calls unchanged.
At inference it consists of a \textbf{probability message} from each agent, a \textbf{noisy-OR fusion} of these messages into one score, and a \textbf{size-aware threshold} on that score (Fig.~\ref{fig:overview}a), which an offline calibration fits for each team size (Fig.~\ref{fig:overview}b).
In the example of Fig.~\ref{fig:overview}, only the third of four cameras sees the apple, and its agent sends a probability above 0.999 while the other three send less than $10^{-5}$, so a rule that demands agreement drops the apple, whereas the noisy-OR score lies far above $\tau_4$.

\subsection{Probability Messages}
\label{sec:message}
Given its image and the question ``Is there a $K$ in this image?'', agent $i$ produces a distribution over its first output token, from which we read $p_i = P_i(\text{yes}) / (P_i(\text{yes}) + P_i(\text{no}))$, the probability mass on yes-tokens relative to the mass on yes- and no-tokens.
The greedy answer of the agent is in effect this probability cut at one half, so $p_i$ can be read from the same forward pass at no additional cost, and sending it instead of the answer keeps the margin by which the agent said yes or no.

\subsection{Fusion under a Size-Aware Threshold}
\label{sec:fusion}
\paragraph{Noisy-OR fusion.}
The team fuses the probabilities into the score
\begin{equation}
s = -\textstyle\sum_{i=1}^{N} \log(1-p_i),
\label{eq:caps}
\end{equation}
which is the noisy-OR of the agents in log form, and it answers yes when $s \geq \tau_N$.
Each agent adds a term that grows steeply as its probability approaches one, so a confident agent weighs far more than a marginal one, and the marginal yeses that an absent object attracts each add little.

\paragraph{Size-aware threshold.}
The threshold $\tau_N$ answers the compounding of false alarms.
Since an absent object attracts more marginal yeses as $N$ grows, its score rises with the team size, and we therefore fit $\tau_N$ separately for each $N$ by maximising balanced accuracy on labelled training scenes that are observed by $N$ cameras.
Distributed detection gives the same prescription for local decisions, whose optimal thresholds depend on the rest of the team~\cite{tenney1981detection}, whereas a greedy answer fixes the cut at one half.

\paragraph{Fitting without labels.}
When no labelled scenes are available, a joint call serves as the teacher: we run it offline on a small set of unlabelled scenes and choose the $\tau_N$ that maximises the team's balanced agreement with its answers, after which the team runs on its own.

\subsection{One-Bit and Counting Variants}
\label{sec:variants}
\paragraph{Calibrated bit.}
When the channel carries only one bit per agent, the threshold is applied at the agent, so that agent $i$ answers yes when $p_i \geq \tau^{\mathrm{b}}_N$ and the team takes the OR of these bits, which amounts to $\max_i p_i \geq \tau^{\mathrm{b}}_N$.
The message is still one bit, but unlike an answer it is the agent's probability cut where the team needs it rather than at one half, and because a single agent above the threshold decides, an object seen by one camera cannot be outvoted.
\emph{Calibrated} here means a fitted cut, not probability calibration~\cite{guo2017calibration}.

\paragraph{Counting.}
For questions of the form ``How many $K$s are in the room?'', each agent also reports a count in a second call, and the \emph{gated} team answers 0 when $s$ is below a fitted threshold and the maximum of the agents' counts (at least 1) otherwise, which suppresses the counts that single agents hallucinate for absent objects.
We also report a learned fusion, a classifier on the agents' counts and scores (supplement).

%% file: sec/5_experiments.tex
\section{Experiments}
\label{sec:experiments}

\input{tables/main_thor}

\subsection{Experimental Setup}
\label{sec:setup}

\paragraph{Testbeds.}
We use three multi-camera testbeds of our own and a public benchmark (details in the supplement).
\textbf{Thor-PV} consists of 260 ProcTHOR rooms~\cite{deitke2022procthor} rendered in AI2-THOR~\cite{kolve2017ai2thor}, each with at least 32 cameras, in which instances of some of eight object kinds are placed and all other small objects are removed, so that presence and per-camera visibility are known exactly; nested sets of $N=4$, 8, 16 and 32 cameras yield 2{,}080 questions.
\textbf{ScanNet}~\cite{dai2017scannet} provides 150 real indoor scans with 16 evenly spaced frames each ($N=4$, 8, 16) and 1{,}799 questions over 12 kinds.
\textbf{AV2-PV} has 300 real driving scenes from Argoverse~2~\cite{wilson2021argoverse2}, each seen by the seven ring cameras of one car ($N=4$, 7), with eight kinds fixed before any model was run.
\textbf{3D-POPE}~\cite{yang20253dgrand} is a public benchmark of yes/no object-existence questions on ScanNet rooms, answered from $N=4$, 8 and 16 frames.

\paragraph{Tasks and metrics.}
For \emph{existence} we report balanced accuracy, the mean of the accuracies on present and on absent kinds; a present kind counts if at least one of the $N$ cameras sees it and is \emph{exclusive} if exactly one does.
As exclusive kinds are rare in the rooms, we also build \emph{lone-witness sets} on Thor-PV, in which exactly one camera sees the object.
For \emph{counting}, on Thor-PV (and AV2-PV in the supplement), the truth is the number of instances that at least one of the $N$ cameras sees, and we report exact accuracy.

\paragraph{Baselines.}
We compare \method with four groups of methods, all of which use the same frozen model, images and questions; a baseline with a citation applies that work's message pattern or decision rule to our agents, without its tools, peer scoring or extra rounds (supplement).
\begin{itemize}[leftmargin=*,nosep]
\item \textbf{Joint call}, a single call that receives all $N$ images, also \emph{calibrated} (its own probability of yes $p_J$ cut at a fitted $\tau_N$) and, for counting, \emph{gated}.
\item \textbf{Answer-merging rules}, namely the OR of the agents' answers, majority voting~\cite{du2024improving}, ``at least $k$ of $N$ say yes'' with $k$ fitted per $N$, the optimal fusion of hard decisions~\cite{chair1986optimal} and self-consistency over five sampled answers per agent~\cite{wang2023selfconsistency}; for counting, the maximum of the agents' counts or a learned fusion of them.
\item \textbf{Multi-stage teams}: a hierarchy of $N/4$ four-view agents, route-then-verify, where a further call inspects the views that said yes, an aggregator whose text-only call merges per-view reports~\cite{wang2025moa,tavella2026moreviews}, and two-round debate~\cite{du2024improving}.
\item \textbf{Confidence and score fusion}, namely the confidence-weighted vote of ReConcile~\cite{chen2024reconcile}, applied to verbal confidences and to first-token probabilities, and the sum rule~\cite{kittler1998combining} with a fitted threshold.
\end{itemize}

\paragraph{Implementation.}
We use Qwen3-VL-8B~\cite{bai2025qwen3vl} throughout and, on subsets of the data, Qwen3-VL-4B and 32B, InternVL3-8B~\cite{zhu2025internvl3}, Idefics3-8B~\cite{laurencon2024building}, LLaVA-OneVision-7B~\cite{li2024llavaov} and Qwen2.5-VL-7B~\cite{bai2025qwen25vl} (supplement), served with vLLM~\cite{kwon2023vllm} or HF Transformers and decoded greedily.
Every threshold and weight, for the teams and for the joint call alike, is fitted per $N$ by five-fold cross-validation over rooms, scans or driving logs and scored on the held-out fold; the noisy-OR and the max rule were retained after comparing four fusion rules on these data (supplement).
For existence the joint call and the $N$ single-view calls see each image once, roughly matching tokens for the Qwen models.

\subsection{Main Results}
\label{sec:main}

\paragraph{\method matches the joint call as the team grows.}
With Qwen3-VL-8B at 32 cameras, \method reaches \textbf{96.6\%} balanced accuracy, level with the calibrated joint call and above the greedy one (95.9\%), and pooled over the four team sizes it is within 0.8 points of the joint call, greedy or calibrated, for both the 8B and the 4B model, and the calibrated bit within one point (Table~\ref{tab:thor}).
A team of single-view agents that each send one number thus recovers nearly all that the model extracts from all views at once.

\paragraph{Answer-merging teams fall behind.}
The OR of the agents' answers drops from 89.1\% to 73.3\% between 4 and 32 cameras with the 8B model, and route-then-verify, the aggregator and debate decline with it despite the calls or tokens they add.
The fitted vote and the optimal fusion of hard decisions stay roughly four to six points behind \method at every size, and majority voting about fifteen, pooled over $N$.

\begin{figure*}[t]
\centering
\figfile{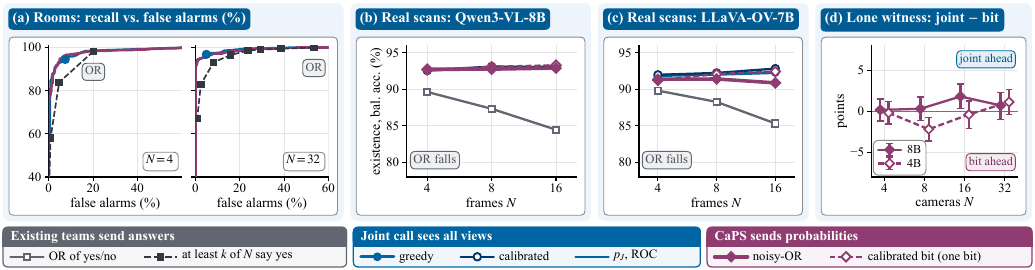}{\textwidth}{4.5cm}
\vspace{-14pt}
\caption{\textbf{Probabilities keep what yes/no answers throw away.}
(a)~Rendered rooms (8B): present objects found (y) against false alarms on absent ones (x); the agents' fused probabilities (purple) follow the joint call that sees all views (blue), while merged answers (grey, black) need more false alarms for the same recall at 32 cameras.
(b,~c)~Real ScanNet scans: as frames are added, the OR of the answers loses accuracy, while the probability teams stay with the joint call.
(d)~Objects seen by one camera only: joint call minus the one-bit team, at most about two points up to 32 cameras (bars: 95\% paired bootstrap intervals over rooms).}
\label{fig:compare}
\vspace{-8pt}
\end{figure*}

\paragraph{\method outperforms the fusion rules of prior work.}
Among the rules of prior work that use confidences or scores, the sum rule with a fitted threshold is the strongest and still trails \method by 3.5 to 4.0 points pooled over $N$, whereas the confidence-weighted vote trails by more than twelve.
Of all baselines, only the hierarchy comes close, as each of its agents is a small joint call over four views.

\input{tables/pope}

\paragraph{\method generalizes across model families.}
On Thor-PV, \method also stays within 1.2 points of the joint call, pooled over $N$, for Qwen3-VL-32B and InternVL3-8B and exceeds it by 4.6 points for Idefics3-8B, while every model's OR team declines with $N$ (supplement).

\paragraph{\method holds on real data and a public benchmark.}
On ScanNet, the OR team of all four models loses 12 to 16 points on absent kinds between 4 and 16 frames (supplement), whereas \method and the calibrated bit stay within 1.2 points of the joint call, pooled over $N$, and lead every prior rule for Qwen3-VL-8B and LLaVA-OneVision-7B, the two models with stored probabilities (Figs.~\ref{fig:compare}b,\,c and~\ref{fig:gallery}a; supplement).
On AV2-PV, whose cameras mostly see different objects, \method is 2.8 points above the greedy joint call at seven cameras and level with the calibrated one (Table~\ref{tab:av2}).
On the public 3D-POPE benchmark (Table~\ref{tab:pope}), the OR teams of Qwen3-VL-8B and 4B lose 17.6 and 18.6 points on absent objects between 4 and 16 frames (supplement), while \method stays level with the joint call and the calibrated bit exceeds it at 16.

\input{tables/av2}

\subsection{Ablations and Analysis}
\label{sec:analysis}

\begin{figure}[t]
\centering
\figfile{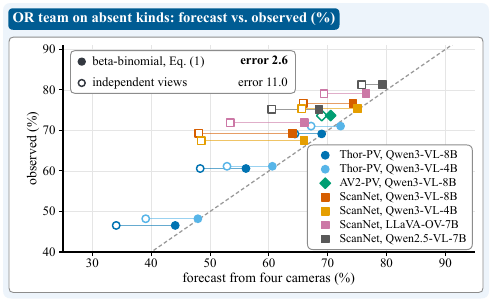}{\columnwidth}{5.1cm}
\vspace{-14pt}
\caption{\textbf{Four cameras are enough to forecast the OR team's false alarms at larger sizes.}
Each pair of points is one setting (model, testbed, more than four cameras): the OR team's observed accuracy on absent objects (y) against the forecast made from four cameras only (x).
Filled: our beta-binomial model, close to the diagonal (mean error 2.6 points); open: assuming independent views (11.0 points).}
\label{fig:law}
\vspace{-8pt}
\end{figure}

\begin{figure*}[t]
\centering
\ifdefined\withholdgallery
\fbox{\parbox[c][\dimexpr5.06cm-14pt-2\fboxsep-2\fboxrule\relax][c]{\dimexpr\textwidth-2\fboxsep-2\fboxrule\relax}{\centering\small This figure shows real ScanNet frames and is withheld in this reading copy; see the full PDF.}}\vspace{14pt}
\else
\figfile{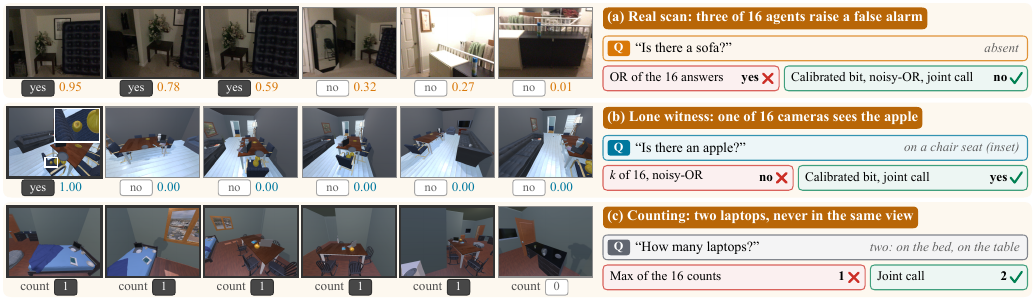}{\textwidth}{5.06cm}
\fi
\vspace{-14pt}
\caption{\textbf{Three examples with 16 cameras (Qwen3-VL-8B).} Under each view, the agent's answer with its $P(\text{yes})$, or its count.
(a)~A real scan without a sofa: three agents wrongly say yes, so the OR of the answers says yes, while the calibrated bit, the noisy-OR and the joint call say no.
(b)~An apple seen by one camera only: the calibrated bit and the joint call find it; the fitted vote and the noisy-OR miss it.
(c)~Two laptops never in the same view: each agent counts what it sees, so the max of their counts is 1; the joint call counts both.}
\label{fig:gallery}
\vspace{-8pt}
\end{figure*}

\paragraph{False alarms compound as the model predicts.}
A single agent is a good detector, which over the 32 cameras of Thor-PV answers yes for 91.7\% of the kinds its camera sees and for only 7.0\% of the absent kinds (8B).
The merge, however, multiplies the latter: the OR team's accuracy on absent kinds falls from 79.9\% at four cameras to 46.6\% at 32, whereas the joint call stays above 92\% and \method reaches 98.9\% (supplement).
The beta-binomial model of Eq.~\eqref{eq:law}, fitted on four cameras' flags in training scenes, forecasts this decay on held-out rooms, scans and driving scenes with a mean absolute error of 2.6 points over 15 settings, against 11.0 points under independence (Fig.~\ref{fig:law}).

\paragraph{Additional machinery does not stop the decay.}
The designs that process the answers further inherit the problem: verification rejects only part of the false alarms that reach it, and the aggregator agrees with the plain OR on 92 to 97\% of the questions.
A stricter prompt removes 34 to 47\% of the false alarms, yet its OR still loses more than twelve points between 4 and 32 cameras, and five sampled answers per agent agree on at least 96\% of the views, and their best rule stays 2.9 to 5.3 points below \method.
Nor does writing the probabilities into a prompt, as in DebUnc~\cite{yoffe2025debunc}: the best aggregator that reads them stays more than four points below \method, pooled over $N$, even with a 32B reader (supplement), so probabilities need a fitted rule, not a reader.\looseness=-1

\paragraph{Probabilities preserve the evidence.}
Before any threshold, the noisy-OR of the agents' probabilities ranks present above absent kinds nearly as well as the joint call's probability, within 0.2 AUC points at every team size for both models on Thor-PV (Fig.~\ref{fig:compare}a; supplement).
Splitting the scene thus loses almost nothing; what a greedy answer discards is the margin, lost once every agent cuts at one half: no rule on single-view greedy answers in Table~\ref{tab:thor} exceeds 92.4\%, whereas the calibrated bit and \method never fall below 94.6\%, and even a fitted vote over the answers to a stricter prompt stays 1.7 to 3.1 points below \method (supplement).

\paragraph{The threshold must follow the team size.}
The fitted thresholds rise steeply with $N$ (Fig.~\ref{fig:overview}b).
For the calibrated bit of the 8B model the cut moves from 0.97 at four cameras to 0.9999 at 32, so that fitting it per size is worth up to 2.6 points over the best single cut, and a threshold carried over from four cameras to 32 costs 10.2 points, whereas the message itself can be coarse: log-odds rounded to one decimal cost \method at most 0.1 points (supplement).\looseness=-1

\paragraph{Few labels, or a joint-call teacher, suffice.}
With only twenty labelled scenes the fit loses at most 1.1 points on average, and a threshold fitted to a joint call's answers on 20 unlabelled scenes keeps the team within 1.3 points of the joint call, pooled over $N$, for four models on Thor-PV and two on ScanNet (supplement).\looseness=-1

\input{tables/strata}

\paragraph{One calibrated bit keeps the lone witness.}
Rules that demand agreement suppress false alarms at the price of the objects that a single camera sees.
At 16 cameras the calibrated bit recalls 87.0\% of the exclusive kinds, more than the joint call (85.5\%), whereas the fitted vote recalls 52.2\% and the noisy-OR 72.5\% (Table~\ref{tab:strata}, 8B).
On lone-witness sets the bit is never more than two points below the joint call up to 32 cameras, for both models (Figs.~\ref{fig:compare}d and~\ref{fig:gallery}b).

\paragraph{Gating recovers most of the counting gap.}
A counting team falls behind mainly because one hallucinating agent makes the maximum of the counts non-zero for an absent kind; gating by the fused score removes almost all of this failure, so that the gated team matches or exceeds the ungated joint call for 8B (Table~\ref{tab:thor}).
The gated joint call stays ahead, as a total requires recognising the same object across views (Fig.~\ref{fig:gallery}c); a learned fusion of the agents' counts and scores closes this gap at 32 cameras for 8B.\looseness=-1

\paragraph{A team is preferable when the joint call is limited.}
The team is ahead when the joint call cannot be given every pixel, a constraint that also drives low-resolution segmentation~\cite{cheng2025maskattnunet}: at 32 cameras a joint call downscaled to the pixel budget of 4, 8 or 16 images stays below \method, falling to 82.7\% at the four-image budget against 96.6\% (8B).
It is also ahead of a greedy joint call that reads many images poorly or answers cautiously (Qwen2.5-VL-7B on ScanNet, Idefics3-8B, AV2-PV).

%% file: tables/main_thor.tex
\begin{table*}[t]
\centering
\footnotesize
\setlength{\tabcolsep}{4.0pt}
\renewcommand{\arraystretch}{0.96}
\begin{tabular}{llrrrrcrrrr}
\toprule
 & & \multicolumn{4}{c}{\textbf{Qwen3-VL-8B}} & & \multicolumn{4}{c}{\textbf{Qwen3-VL-4B}} \\
\cmidrule(lr){3-6}\cmidrule(l){8-11}
 & \textbf{Method} \hfill $N=$ & 4 & 8 & 16 & 32 & & 4 & 8 & 16 & 32 \\
\midrule
\rowcolor{bandbg}\multicolumn{11}{l}{\textbf{Existence} (balanced accuracy, \%)} \\
\multirow{2}{*}{\chip{cJoint}{All views}} & Joint call & \second{93.5} & 94.8 & 95.6 & \second{95.9} & & \second{95.3} & \textbf{96.9} & \textbf{97.4} & \second{97.4} \\
 & Calibrated joint call, $p_J \geq \tau_N$$^\dagger$ & 93.0 & \second{95.4} & \textbf{96.3} & \textbf{96.6} & & 95.2 & \second{96.4} & \textbf{97.4} & \textbf{98.0} \\
\grule
\multirow{8}{*}{\chip{cOld}{Answers}} & OR of yes/no & 89.1 & 84.3 & 80.2 & 73.3 & & 90.1 & 85.3 & 80.5 & 74.0 \\
 & Majority vote~\cite{du2024improving} & 78.7 & 80.3 & 81.0 & 80.4 & & 79.3 & 82.1 & 82.8 & 83.1 \\
 & $k$ of $N$ say yes$^\dagger$ & 88.4 & 89.8 & 91.1 & 91.9 & & 89.6 & 91.2 & 91.3 & 92.4 \\
 & Optimal hard fusion~\cite{chair1986optimal}$^\dagger$ & 89.6 & 90.7 & 91.5 & 92.2 & & 91.1 & 91.5 & 91.8 & 92.3 \\
 & Debate, two rounds~\cite{du2024improving} & 88.0 & -- & 83.3 & -- & & 84.1 & -- & 80.3 & -- \\
 & Four-view hierarchy, $k$ of $N/4$$^\dagger$ & \second{93.5} & 92.3 & 94.7 & 94.8 & & 95.2 & 95.9 & 96.3 & 96.9 \\
 & Route, then verify & 89.9 & 86.1 & 83.4 & 78.1 & & 91.6 & 88.7 & 86.8 & 82.9 \\
 & Reports + aggregator~\cite{wang2025moa,tavella2026moreviews} & 89.7 & 84.7 & 80.5 & 73.5 & & 87.6 & 84.5 & 81.4 & 75.9 \\
\grule
\multirow{3}{*}{\lchip{cOld}{Scores}} & Weighted vote, verbal confidences~\cite{chen2024reconcile} & 73.8 & 71.6 & 70.0 & 68.1 & & 61.5 & 57.9 & 56.1 & 55.2 \\
 & Weighted vote on $p_i$~\cite{chen2024reconcile} & 81.1 & 82.5 & 82.7 & 81.9 & & 81.5 & 83.9 & 84.0 & 84.3 \\
 & Sum rule on $p_i$~\cite{kittler1998combining}$^\dagger$ & 93.2 & 92.0 & 91.4 & 92.2 & & 93.4 & 92.5 & 91.3 & 91.1 \\
\grule
\rowcolor{oursbg} & Calibrated bit: OR of $p_i \geq \tau_N$ (max-P)$^\dagger$ & \textbf{94.6} & \textbf{95.7} & 94.9 & 94.8 & & 94.8 & \second{96.4} & 96.3 & 96.0 \\
\rowcolor{oursbg}\multirow{-2}{*}{\chip{cFus}{Ours}} & \textbf{\method}: noisy-OR $\geq \tau_N$$^\dagger$ & \textbf{94.6} & \textbf{95.7} & \second{96.0} & \textbf{96.6} & & \textbf{95.6} & 95.6 & \second{96.7} & 96.3 \\
\midrule
\rowcolor{bandbg}\multicolumn{11}{l}{\textbf{Counting} (exact accuracy, \%)} \\
\multirow{2}{*}{\chip{cJoint}{All views}} & Joint call & 57.3 & 56.5 & 58.4 & 58.0 & & \second{63.8} & \second{63.7} & \second{64.7} & 63.7 \\
 & Joint call, gated by $p_J$ (two calls)$^\dagger$ & \textbf{62.9} & \textbf{59.7} & \textbf{60.9} & \second{61.2} & & \textbf{64.6} & \textbf{65.1} & \textbf{66.8} & \textbf{66.4} \\
\grule
\multirow{4}{*}{\chip{cOld}{Answers}} & Max of counts & 53.5 & 50.8 & 49.9 & 48.9 & & 58.1 & 55.0 & 54.9 & 52.4 \\
 & Learned fusion of counts$^\dagger$ & 54.6 & 53.5 & 53.2 & 54.0 & & 57.7 & 54.0 & 55.4 & 57.6 \\
 & Route, then verify & 55.9 & 53.9 & 53.9 & 53.4 & & 60.8 & 61.3 & 61.4 & 60.1 \\
 & Reports + aggregator & 59.1 & 51.4 & 34.6 & 27.5 & & 54.8 & 31.3 & 17.5 & 11.9 \\
\grule
\rowcolor{oursbg} & \textbf{\method}: max of counts, gated by noisy-OR$^\dagger$ & \second{60.7} & 58.6 & 59.0 & 59.4 & & 63.4 & 60.3 & 61.3 & 61.2 \\
\rowcolor{oursbg}\multirow{-2}{*}{\chip{cFus}{Ours}} & Learned fusion of counts + scores$^\dagger$ & 59.5 & \second{58.8} & \second{59.8} & \textbf{62.7} & & 61.3 & 61.1 & 63.5 & \second{64.2} \\
\bottomrule
\end{tabular}
\vspace{-4pt}
\caption{\textbf{Single-view teams that merge greedy answers stay behind the joint call, and those that trust any yes fall further as cameras are added; \method comes within 0.8 points of it on existence, pooled over $N$.}
Thor-PV, 260 rooms, the same frozen model in every row.
$^\dagger$Threshold or weights fitted on training rooms and scored on held-out rooms (5-fold CV by room).
Best per column in bold, second best underlined; our rows are shaded.}
\label{tab:thor}
\vspace{-6pt}
\end{table*}

%% file: tables/pope.tex
\begin{table}[t]
\centering
\footnotesize
\setlength{\tabcolsep}{3.4pt}
\begin{tabular}{@{}lcccccc@{}}
\toprule
 & \multicolumn{3}{c}{\textbf{Qwen3-VL-8B}} & \multicolumn{3}{c}{\textbf{Qwen3-VL-4B}} \\
\cmidrule(lr){2-4}\cmidrule(l){5-7}
\textbf{Method} \hfill $N=$ & 4 & 8 & 16 & 4 & 8 & 16 \\
\midrule
\famsq{cJoint}Joint call & \second{74.2} & 77.8 & \second{80.6} & 74.1 & 77.9 & \second{80.4} \\
\grule
\famsq{cOld}OR of yes/no & 73.7 & 75.3 & 73.5 & 75.1 & 75.7 & 73.8 \\
\famsq{cOld}$k$ of $N$ say yes$^\dagger$ & 73.7 & 75.3 & 75.2 & 75.1 & 75.7 & 77.3 \\
\famsq{cOld}Self-consistency~\cite{wang2023selfconsistency}$^\dagger$ & 73.8 & 76.4 & 76.1 & 75.0 & 76.0 & 77.7 \\
\famsq{cOld}Sum rule on $p_i$$^\dagger$ & \textbf{74.3} & 76.7 & 76.4 & 75.6 & 77.6 & 78.2 \\
\grule
\rowcolor{oursbg}\famsq{cFus}Calibrated bit$^\dagger$ & 73.9 & \second{78.1} & \textbf{82.3} & \textbf{77.0} & \textbf{79.0} & \textbf{81.8} \\
\rowcolor{oursbg}\famsq{cFus}\textbf{\method}$^\dagger$ & \second{74.2} & \textbf{79.2} & 80.5 & \second{75.7} & \second{78.8} & 80.2 \\
\bottomrule
\end{tabular}
\vspace{-4pt}
\caption{\textbf{On the public 3D-POPE benchmark, rules on answers fall behind the joint call as frames are added; \method and the calibrated bit keep pace.}
Balanced accuracy (\%), 61 ScanNet scenes with 16 frames (1{,}880 questions).
$^\dagger$Fitted per $N$ by 5-fold CV.
Self-consistency: best of four rules on five sampled answers.}
\label{tab:pope}
\vspace{-6pt}
\end{table}

%% file: tables/av2.tex
\begin{table}[t]
\centering
\footnotesize
\setlength{\tabcolsep}{2.4pt}
\begin{tabular}{lcccccc}
\toprule
 & \multicolumn{2}{c}{\textbf{Balanced acc.}} & \multicolumn{2}{c}{\textbf{Exclusive recall}} & \multicolumn{2}{c}{\textbf{Absent acc.}} \\
\cmidrule(lr){2-3}\cmidrule(lr){4-5}\cmidrule(l){6-7}
\textbf{Method} \hfill $N=$ & 4 & 7 & 4 & 7 & 4 & 7 \\
\midrule
\famsq{cJoint}Joint call             & 81.8 & 79.2 & 68.1 & 60.0 & \textbf{92.1} & \textbf{91.6} \\
\famsq{cJoint}Calibrated joint call  & 82.2 & 81.2 & \second{79.4} & 69.1 & 83.9 & \second{86.4} \\
\grule
\famsq{cOld}OR of yes/no           & \second{82.7} & 80.8 & \textbf{84.5} & \textbf{83.1} & 79.2 & 73.7 \\
\grule
\rowcolor{oursbg}\famsq{cFus}Calibrated bit (max-P) & 82.6 & \second{81.9} & 76.4 & 73.1 & 86.0 & 84.2 \\
\rowcolor{oursbg}\famsq{cFus}\textbf{\method} (noisy-OR) & \textbf{83.1} & \textbf{82.0} & 77.3 & \second{75.6} & \second{86.3} & 82.6 \\
\bottomrule
\end{tabular}
\vspace{-4pt}
\caption{\textbf{On a real seven-camera car rig, \method beats the greedy joint call and is level with the calibrated one.}
AV2-PV, 300 driving scenes, Qwen3-VL-8B; at seven cameras, 598 present object kinds (320 seen by one camera only) and 1{,}284 absent ones.}
\label{tab:av2}
\vspace{-6pt}
\end{table}

%% file: tables/strata.tex
\begin{table}[t]
\centering
\footnotesize
\setlength{\tabcolsep}{1.7pt}
\begin{tabular}{lrrrrcrrrr}
\toprule
 & \multicolumn{4}{c}{\textbf{Qwen3-VL-8B}} & & \multicolumn{4}{c}{\textbf{Qwen3-VL-4B}} \\
\cmidrule(lr){2-5}\cmidrule(l){7-10}
\textbf{Method} & excl. & shar. & abs. & lone & & excl. & shar. & abs. & lone \\
\midrule
\famsq{cJoint}Joint call         & 85.5 & 96.4 & 95.3 & 88.0 & & 76.8 & 98.0 & 97.8 & 87.8 \\
\famsq{cJoint}Calibrated joint   & 76.8 & 94.7 & 98.6 & 88.0 & & 73.9 & 97.4 & 98.3 & 89.9 \\
\grule
\famsq{cOld}OR of yes/no       & 100  & 99.8 & 60.6 & 79.1     & & 98.6 & 99.9 & 61.2 & 79.7 \\
\famsq{cOld}$k$ of $N$         & 52.2 & 93.6 & 90.4 & ---      & & 46.4 & 93.9 & 90.6 & --- \\
\grule
\rowcolor{oursbg}\famsq{cFus}Calibrated bit     & 87.0 & 96.3 & 93.9 & 87.4     & & 89.9 & 96.7 & 96.1 & 89.4 \\
\rowcolor{oursbg}\famsq{cFus}\textbf{\method} (noisy-OR) & 72.5 & 95.8 & 97.2 & 85.3     & & 76.8 & 97.0 & 97.2 & 87.4 \\
\bottomrule
\end{tabular}
\vspace{-4pt}
\caption{\textbf{The fitted vote loses the lone witness; the calibrated bit keeps it.}
Thor-PV, $N=16$: recall on exclusive (one of 16 cameras sees the kind; 69 questions) and shared kinds (1{,}612), accuracy on absent kinds (363), and balanced accuracy on lone-witness sets built from the first 32 cameras (one camera sees the kind, 15 do not; 1{,}019 present questions), with refitted thresholds.}
\label{tab:strata}
\vspace{-6pt}
\end{table}

%% file: sec/7_conclusion.tex
\section{Conclusion}
\label{sec:conclusion}

We presented \method, a protocol in which the frozen VLM agents of a multi-camera team share their probability of yes and the team sets the threshold for its own size.
On rendered rooms, real scans and driving scenes, merging answers lets false alarms compound with every added camera, a decay that a beta-binomial model forecasts and that verification, aggregation and debate do not stop.
On object existence, sharing probabilities counters it without training or extra calls, bringing a team within 1.2 points of the joint call for five VLMs and above it for a sixth, also on the public 3D-POPE benchmark, and within 1.3 points when the threshold is fitted to a joint call's answers instead of labels, while one calibrated bit per agent preserves the lone witness.
The broader lesson is that the decision threshold belongs to the team, which knows its size, not to each agent.

\paragraph{Limitations.}
\method needs first-token log-probabilities and a threshold fitted for each team size and camera layout; Thor-PV removes all other small objects, so clutter is tested on the real scans and 3D-POPE; and counting and other cross-view reasoning may need richer messages.
\label{lastbody}

%% file: sec/8_llm_statement.tex
\section*{LLM Usage Statement}
We used a large language model, Claude (Anthropic), only at the assist level, \ie, to edit or refine content produced by the authors.
This applied to writing (language, restructuring, and prose), related work, code and implementation, the design, execution, and analysis of experiments, figures and visual material, data, and the presentation of research ideas, hypotheses, claims, and conclusions.
We also provided the full manuscript to Claude to obtain reviewer-style feedback.
All research ideas, methods, code, data, experiments, and results originate from the authors, who checked all LLM-assisted content and take full responsibility for the submitted paper.

%% file: supp.tex
\clearpage
\section*{Supplementary Material}
\setcounter{section}{0}
\setcounter{table}{0}
\renewcommand{\theHsection}{supp.\arabic{section}}
\renewcommand{\theHtable}{supp.\arabic{table}}
\renewcommand{\dblfloatpagefraction}{0.8}
\renewcommand{\thesection}{\Alph{section}}
\renewcommand{\thetable}{S\arabic{table}}

\section{Prompts}
\label{sec:s-prompts}
All calls are greedy and answer-only except the per-view report calls, which allow up to 160 new tokens and carry no answer-only suffix, and the reasoning reader of Sec.~\ref{sec:s-more} (up to 256 new tokens); $\langle K\rangle$ is the kind with its article, $\langle n\rangle$ the number of images in words.
\begin{itemize}\setlength\itemsep{1pt}
\item Single view, existence: ``Is there $\langle K\rangle$ in this image? Answer yes or no.''
\item Single view, counting: ``How many $\langle K\rangle$s are in this image? Answer with a number.''
\item Joint call, existence: ``These $\langle n\rangle$ images show different parts of one room. Is there $\langle K\rangle$ anywhere in the room? Answer yes or no.''
\item Joint call, counting: ``These $\langle n\rangle$ images show different parts of one room. Some objects may appear in more than one image. How many $\langle K\rangle$s are in the room in total? Count each object once. Answer with a number.''
\item Route, then verify: the views whose single-view answer flags the kind (yes, or a count above 0) go to one call with the joint-call wording for that many images (``This image shows part of one room. \dots'' for one view); with no flagged view the team answers no (0) without a call.
\item Report: ``How many $\langle K\rangle$s can you see in this image? If there is none, reply `none'. Otherwise give the number, then describe each $\langle K\rangle$ once, one short line each: its color, its size, where it is in the image (left, center or right; top or bottom) and what it is on or next to.''
\item Aggregator: ``$\langle n\rangle$ cameras in one room each reported how many $\langle K\rangle$s they can see and described them. Different cameras may see the same object from different angles.'' then ``Camera $j$: $\langle$report$\rangle$'' for each camera, then the joint-call question.
\item Hierarchy: agent $g$ sees cameras $4g$ to $4g{+}3$ with the joint-call wording for four images; the $N/4$ group answers are merged by the vote rules.
\item AV2-PV: ``show different parts of one room'' becomes ``were taken at the same moment by the cameras of one car'', ``anywhere in the room'' becomes ``in any of these images'' and ``are in the room in total'' becomes ``appear in these images in total'' (Sec.~\ref{sec:s-av2} reports the first wording, ``anywhere around the car'').
\end{itemize}
$P(\text{yes})$ is the probability mass on yes-tokens over the mass on yes- and no-tokens among the top 20 first-token log-probabilities, logged in a separate call with one new token (the forward pass that produces the answer yields the same quantity); the greedy answer agrees with $P(\text{yes}) \geq 0.5$ on 99.6\% of Thor-PV view calls. The counting gates use these scores but a different fitting target from the existence thresholds: they are fitted on all training questions to separate kinds with at least one instance in view from the rest, so present kinds no camera sees count as negatives, whereas the existence thresholds drop those questions. At $N=4$ the gated joint call therefore answers 0 on 97.2\% of absent kinds (8B; 4B 98.1\%), against 93.9\% (96.7\%) for the calibrated joint call. A gated count needs a count call and a $P(\text{yes})$ call over the same images: per question at $N=32$, 49.4k tokens for the gated joint call and 51.1k for the gated max and for the learned fusion with scores (6.3k and 6.4k at $N=4$).

\section{Testbeds}
\label{sec:s-testbeds}
\paragraph{Thor-PV.} Rooms are ProcTHOR houses rendered in AI2-THOR at $1024{\times}768$; models see the images at that size.
Each room keeps up to 64 cameras whose view shows at least one instance of the eight kinds; the order of cameras is random and fixed, and the $N$-camera set is its first $N$ entries, so the sets are nested.
Of 300 generated rooms, 260 have at least 32 cameras.
Each room contributes one question per kind (2{,}080 per $N$); 363 are absent kinds and the number of present-and-seen kinds is 1{,}469, 1{,}621, 1{,}681 and 1{,}709 at $N=4$ to 32 (exclusive 378, 179, 69 and 25).
\textbf{No overlays.} An earlier render of the same rooms drew a numbered tag on every object; numbers were consistent across cameras.
Tags raise the per-view false-alarm rate (Qwen3-VL-8B 10.3\% against 7.0\%) and lower the joint call's accuracy on absent kinds (87.6 to 91.7\% against 92.6 to 95.3\%), and they give the joint call a cross-view identity cue.
All main-text Thor-PV numbers use the untagged re-render; the tagged runs appear only in Table~\ref{tab:s-law}.

\paragraph{Needle protocol.} For each present kind the camera set is one camera that sees it plus $N-1$ cameras of the same room that do not, shuffled; absent kinds get $N$ random cameras of the room. Questions without enough non-seeing cameras are dropped (1{,}632, 1{,}554, 1{,}293 and 671 present questions at $N=4$ to 32).

\paragraph{ScanNet.} 150 scans with at least 16 frames, one per space (the lowest rescan index), drawn with a fixed seed from the public 25k-frame subset; 16 frames evenly spaced along each scan. A frame sees a kind if an instance covers at least 1{,}000 pixels at $1296{\times}968$; a kind is present if some frame of the whole scan sees it and absent if no frame of the scan has a pixel of it, and kinds with pixels but never 1{,}000 in a frame are dropped. Images are shown at $1024{\times}768$ and were processed only on our own machines by open-weight models. The 12 NYU40 kinds are chair, table or desk, sofa, bed, pillow, lamp, television, sink, toilet, refrigerator, towel and bookshelf. LLaVA-OneVision tiles a lone image but not the images of a multi-image prompt, so its agents read about 6.5 times the prompt tokens of its joint call (measured at 4 and 16 frames).

\paragraph{AV2-PV.} Sweeps are taken 8\,s apart from 150 validation logs; images are resized so the long side is 1{,}024 pixels. An instance is visible to a camera if the centre of its 3D box lies in front of the camera, within 60\,m of the car, holds at least five lidar points, and projects to at least 2{,}400 pixels with at least 30\% of its area inside the image; occlusion is not modelled. $N=4$ uses the front, rear-left, side-left and side-right cameras. Prevalence decided the eight kinds before any model call.

\section{Detailed results for the main text}
\label{sec:s-detail}
The main text states each finding with one or two headline numbers. This section gives the full numbers and 95\% bootstrap intervals behind those statements; a lead is the joint call minus the team, in points. Figure and equation numbers in this section refer to the main text.

\paragraph{Setup details.} We render ProcTHOR~\cite{deitke2022procthor} rooms in AI2-THOR~\cite{kolve2017ai2thor} at $1024{\times}768$.
Rooms hold instances of some of eight kinds (apple, book, bowl, laptop, mug, remote control, vase, plate); all other pickupable objects are removed, so the truth for ``is there a $K$?'' is the generator's list, and the generator records every instance's visibility per camera (at least 150 pixels at $512{\times}384$).
Each room has up to 64 cameras, each seeing some instance; the $N$-camera sets are the first $N$ of a random order, hence nested.
The 260 rooms with at least 32 cameras give 2{,}080 questions at $N=4, 8, 16, 32$; renders carry no overlays.
150 ScanNet~\cite{dai2017scannet} scans and 12 NYU40 kinds give 1{,}799 questions, 565 present.
Sixteen frames are evenly spaced along each scan ($N=4$ and 8 take every fourth and every second frame); a frame sees a kind if an instance covers at least 1{,}000 pixels, a kind is present if some frame of the scan sees it, absent if no frame has a pixel of it, and otherwise dropped.
From the Argoverse~2 sensor validation split~\cite{wilson2021argoverse2} we take two annotated sweeps, 8\,s apart, from each of 150 logs: 300 scenes, each with its seven ring cameras.
Eight kinds were fixed from annotation prevalence before any model was run: bus, truck, bicycle, motorcycle, traffic cone, stop sign, bollard and trailer.
An instance is visible to a camera if its 3D box lies in front of the camera and near the car, holds enough lidar points and projects to a large enough area of the image; occlusion is not modelled.
$N=4$ uses four of the seven cameras, and the prompts ask about the images of one car instead of a room.
\emph{Existence}: a present kind is \emph{exclusive} if one of the $N$ cameras sees it and \emph{shared} if several do; present kinds that none of the $N$ cameras sees are left out.
We report balanced accuracy, the mean of the accuracies on present and on absent kinds, and both parts.
\emph{Counting} (Thor-PV and AV2-PV): ``How many $K$s are in the room in total? Count each object once.''
The truth is the number of instances seen by at least one of the $N$ cameras; we report exact accuracy.
CIs are 95\% paired bootstrap intervals over rooms and scans, and on AV2-PV the cross-validation folds and the bootstrap keep the two sweeps of a log together. A lead is the joint call minus the team (points).
For existence, the joint call and the $N$ single-view calls see each image once, so for the Qwen models they are token-matched: 3.1k against 3.2k prompt and completion tokens per question at $N=4$, 24.7k against 25.6k at $N=32$.
LLaVA-OneVision tiles a lone image but not the images of a multi-image prompt, so its agents read about 6.5 times the prompt tokens of its joint call and its comparison is not token-matched.
The models are Qwen3-VL-8B~\cite{bai2025qwen3vl} on all three testbeds, Qwen3-VL-4B on Thor-PV and ScanNet, Qwen3-VL-32B, InternVL3-8B~\cite{zhu2025internvl3} and Idefics3-8B~\cite{laurencon2024building} on Thor-PV, and LLaVA-OneVision-7B~\cite{li2024llavaov} and Qwen2.5-VL-7B~\cite{bai2025qwen25vl} on ScanNet, run with vLLM~\cite{kwon2023vllm} (ScanNet: HF Transformers), greedy and answer-only except the aggregator team's per-view reports, which allow up to 160 new tokens, and one reasoning reader.
For counting, each agent gives a count and the team takes the maximum or applies a gradient-boosted classifier to the kind, the eight largest counts, and the sum, maximum and number of nonzero counts.
The gated joint call answers 0 when $p_J < \tau'_N$ and otherwise its count (at least 1), where $\tau'_N$, like the team's gate, is fitted on all training questions to tell kinds some camera sees from the rest.
In Table~1 of the main text, 1{,}832 to 2{,}072 of the 2{,}080 questions enter the existence metric per $N$.

\paragraph{Answer-merging teams against the joint call.} In the rendered rooms (Table~1 of the main text), even the vote threshold fitted for each $N$ trails the joint call, by $+5.1$ $[+3.4, +6.6]$ points at four cameras and $+4.1$ $[+2.5, +5.6]$ at 32 with Qwen3-VL-8B, and by $+5.0$ to $+6.2$ with 4B.
Fixed rules and richer designs fall further, as the joint call leads route-then-verify by $+17.8$ $[+15.5, +20.3]$ and the aggregator by $+22.5$ $[+20.0, +25.0]$ at $N=32$ (8B; OR and 4B in Table~1 of the main text).
On counting, the max of counts likewise trails by $+3.7$ at $N=4$ and $+9.0$ $[+6.9, +11.0]$ at $N=32$ (8B), and the aggregator's accuracy falls from 59.1 to 27.5\%.
Only the hierarchy of $N/4$ four-view agents, which sends fewer messages, stays within 1.2 points of the joint call at $N \geq 16$ (it trails by $+2.5$ $[+1.1, +4.0]$ at $N=8$ for 8B).

\paragraph{False alarms.} A single agent is a good detector, which over the 32 cameras of Thor-PV says yes for 91.7\% of the kinds its camera sees and for only 7.0\% of the absent kinds (8B), but the merge multiplies the latter.
An absent kind is flagged by 0.26 views on average at $N=4$ and by 2.25 at $N=32$, so the OR team's accuracy on absent kinds falls from 79.9 to 46.6\% (Table~\ref{tab:s-law}).
Route-then-verify filters these flags but removes only part of them, rejecting 57.0\% of absent kinds at $N=32$ against 46.6\% for OR alone.
The aggregator reads separate text reports and mostly ORs them, agreeing with the OR of the view answers on 92 to 97\% of questions (85 to 92\% of those about absent kinds), so its accuracy on absent kinds falls from 82.6 to 51.2\% between $N=4$ and 32.
Nor does anything else a practitioner would try first stop the fall.
Asked to say yes only when they clearly see the object, agents raise about 40\% fewer false alarms, yet their OR still declines from 92.9 to 80.3\% balanced accuracy (8B) and from 92.7 to 78.4\% (4B) between 4 and 32 cameras, and two rounds of debate reach only 88.0 and 83.3\% (8B) and 84.1 and 80.3\% (4B) at 4 and 16 cameras.
Among aggregators that are handed the agents' probabilities, told how to read them, shown labelled examples or asked to reason, the best stays 8.6 (8B) and 9.2 (4B) points below \method pooled over $N$, and a 32B reader 4.8, which shows that probabilities need a fitted rule, not a reader.
Higher vote thresholds avoid the false alarms but lose the kinds that one camera sees, since at $N=16$ the fitted threshold keeps 52.2\% of exclusive kinds against 85.5\% for the joint call (Table~4 of the main text), and on sets rebuilt so that exactly one of $N$ cameras sees each present kind the joint call leads the fitted vote by $+8.4$ $[+5.9, +10.9]$ at $N=32$ (8B).
The joint call, by contrast, accumulates far less, with an accuracy on absent kinds of 92.6, 94.2, 95.3 and 95.0\% at $N=4$ to 32 on Thor-PV.

\paragraph{The beta-binomial forecast.} We fit $(\alpha, \beta)$ per kind by maximum likelihood on the flag counts of the first four cameras of the training rooms and forecast every larger $N$ on the held-out rooms (Fig.~4).
Over all 15 settings with $N > 4$ (Thor-PV, two models; ScanNet, four; AV2-PV, one), the mean absolute error is 2.6 points, against 11.0 for independent views.
At $N=32$ on Thor-PV it forecasts 44.1\% against the observed 46.6\% for 8B and 47.9 against 48.2\% for 4B, where independence gives 34.0 and 39.1\%; on AV2-PV it forecasts 70.6 against 73.7\% at seven cameras.
\input{tables/s_scannet}
On ScanNet (Table~\ref{tab:s-scannet}) each of the four models' OR teams loses 12.3 to 15.6 points on absent kinds from 4 to 16 frames, while the joint call moves by $-2.2$ to $+1.4$; the forecast at $N=16$ is 1.5 to 6.7 points low.
The forecast errs on the pessimistic side in 14 of these 15 settings and is off by at most 6.7 points; for three further room models (Qwen3-VL-32B, InternVL3-8B and Idefics3-8B) it is within 2.1 points at every $N$.

\paragraph{Ranking by fused probabilities.} On Thor-PV, without any threshold (Fig.~3a), the noisy-OR of the agents' probabilities ranks present-and-seen kinds above absent ones nearly as well as the joint call's $p_J$, with an AUC of 98.5, 99.0, 99.2 and 99.2 against 98.4, 99.0, 99.3 and 99.4 at $N=4$ to 32 for 8B, and within 0.2 points at every $N$ for 4B.
The team's AUC thus does not fall with more cameras (on AV2-PV the noisy-OR's falls from 91.2 to 89.9 between four and seven), but a greedy yes/no cannot use this ranking, because it fixes each agent's threshold at 0.5 for any team size.

\paragraph{\method against the joint call.} Once its threshold is fitted for the team size, the noisy-OR comes within one point of the joint call in the rooms, pooled over $N$ (Table~1 of the main text), at $+0.7$ $[+0.1, +1.4]$ for 8B and $-0.7$ $[-1.4, +0.1]$ for 4B, and at $+0.4$ $[-0.2, +0.9]$ and $-0.7$ $[-1.4, -0.0]$ against the calibrated joint call.
The same holds on real scans (Fig.~3b,\,c; one scan in Fig.~5a), where against the calibrated joint call and pooled over $N=4$ to 16 \method sits at $-0.1$ $[-0.8, +0.6]$ for Qwen3-VL-8B and $-1.1$ $[-1.9, -0.4]$ for LLaVA-OneVision-7B and the calibrated bit at $+0.1$ $[-0.5, +0.7]$ and $-0.4$ $[-1.3, +0.3]$.
The pattern also holds beyond Qwen3-VL. For InternVL3-8B the OR team falls from 74.1 to 58.9 between 4 and 32 cameras while \method holds 82.6 to 84.2, level with its joint call ($-0.2$ $[-1.5, +1.1]$ pooled), and for Idefics3-8B, whose joint call is cautious, the OR team falls from 81.8 to 68.1 while \method reaches 84.0 to 86.5, $+4.6$ $[+3.2, +5.8]$ points above the joint call and $+3.5$ $[+2.4, +4.6]$ above its calibrated version. Nor does a model four times larger escape, since Qwen3-VL-32B agents flag absent kinds as often as 8B agents (7.5 against 7.0\% of views), their OR team falls from 89.4 to 74.7, and \method holds 94.0 to 95.0, within 1.2 points of the 32B joint call pooled over $N$.
For 8B the noisy-OR does give up some lone witnesses (Table~4 of the main text: at $N=16$ it recalls 72.5\% of the kinds only one camera sees, the joint call 85.5\%), and these are what the calibrated bit keeps.

\paragraph{The calibrated bit and the lone witness.} An OR team whose agents answer yes only when $p_i \geq \tau^{\mathrm{b}}_N$ sends one bit per agent, and its balanced accuracy at $N=32$ rises from 73.3 to 94.8 for 8B and from 74.0 to 96.0 for 4B (Table~1 of the main text).
Pooled over $N$, it is $+0.0$ $[-0.7, +0.8]$ and $-0.9$ $[-1.6, -0.1]$ against the joint call.
Unlike vote thresholds, it keeps the lone witness (Fig.~5b), and it does so at scale.
On lone-witness sets drawn from 64 cameras, with one seer among $N$ (1,632 to 671 present questions for $N=4$ to 32), the bit with the thresholds of the ordinary questions reaches 88.7, 88.4, 86.4 and 87.2\% balanced accuracy against 88.9, 88.7, 88.1 and 87.9\% for the joint call (8B), and 90.2, 88.8, 88.2 and 87.0\% against 90.0, 86.6, 87.8 and 88.1\% (4B), so that at 32 cameras one bit per agent is level with a call that sees every view (Fig.~3d).
The fitted threshold rises with the team, from 0.97 at four cameras to 0.9999 at 32 for 8B, which is why it is fitted per team size; for 8B this is worth up to 2.6 points over the best single cut for all sizes, and a threshold carried from four cameras to 32 loses 10.2.

\paragraph{Counting.} The counting team falls behind mainly because one hallucinating view makes the max of counts non-zero on an absent kind, and answering 0 whenever the noisy-OR says absent removes this.
At $N=32$ the gated team answers 0 for 98.7\% of kinds with no instance in view (8B; 25.1\% for the max of counts, 65.8\% for the joint call), and for 8B it matches or beats the ungated joint call (Table~1 of the main text).
The joint call can, however, be gated by its own $p_J$, and it then leads the gated max by $+1.1$ to $+2.2$ points for 8B (resolved at 4 and 16 cameras) and by up to $+5.5$ for 4B.
This lead sits on present kinds, where a total needs the same object recognised across views and a gate cannot help, and there the gated joint call leads the gated max by 2.3 to 2.8 points (8B; Fig.~5c).
A learned fusion of the agents' counts and scores (Table~1 of the main text) narrows the overall gap at $N=32$ to $-1.5$ $[-4.3, +1.1]$ for 8B and $+2.3$ $[-0.3, +4.9]$ for 4B.

\paragraph{The driving rig.} AV2-PV has few cameras and little overlap, and the per-view hit rate of Qwen3-VL-8B is lower than in the rooms (74.9\% of the kinds a camera sees, with 6.5\% false alarms on absent kinds).
Here the greedy joint call is conservative, finding 60.0\% of the kinds that one of seven cameras sees against 83.1\% for OR (Table~3 of the main text).
With seven agents OR's false alarms stay moderate (its accuracy on absent kinds falls from 79.2 to 73.7\% between four and seven cameras), and \method beats the greedy joint call by $+2.8$ $[+1.2, +4.3]$ at $N=7$.

\paragraph{When a team is ahead.} A team's calls are small and independent. At $N=32$ one agent call reads 0.8k prompt tokens where the joint call needs 24.7k in one context, the agents can run in parallel or on the cameras' own devices, and each sends one number instead of an image.
When the joint call cannot have every pixel, it loses more than the split costs, since restricted to the pixel budget of 4, 8 or 16 images, with its $N$ images downscaled to fit, it falls to 82.7, 90.4 and 94.1 on existence at $N=32$ (8B; 85.6 for 4B at four images), while \method holds 96.6 and 96.3.
A team can also be ahead when the model reads many images poorly (on ScanNet the joint call of Qwen2.5-VL-7B falls from 90.6 to 86.3 between 4 and 16 frames, and the fitted vote threshold leads it by 3.7 at $N=16$, Table~\ref{tab:s-scannet}), or on a few-camera rig that mostly sees different objects, where the team's higher single-camera recall makes up for the joint call's caution (Table~3 of the main text).

\paragraph{What a team should send.} On existence in the rooms, a team that sends probabilities leads OR, route-then-verify and the aggregator by margins that grow with $N$ (23.3 points over OR at $N=32$, 8B) and the fitted vote by 4.7 to 6.1 points (8B), and it leads the fusion rules of prior multi-agent work (majority voting, the optimal fusion of hard decisions, the sum rule, a confidence-weighted vote and two rounds of debate; Table~1 of the main text) by 3.5 to 38 points pooled over $N$, and on real scans by 1.7 to 24 points for both models, while against the four-view hierarchy, which also sends answers, \method leads by 1.1 to 3.4 points for 8B and is within 0.6 points either way for 4B.
On real scans it leads the fitted vote by 3.1 to 3.9 points for Qwen3-VL-8B and by 0.5 to 3.8 for LLaVA-OneVision-7B, whereas on AV2-PV its lead over OR is not resolved.
These results suggest four rules.
(1)~Have agents report the log-odds of yes (one decimal suffices, two decimals of a probability do not), or answer yes only above a threshold fitted for the number of agents; do not OR greedy answers at large $N$ (at small $N$ on a low-overlap rig, OR can be competitive, Table~3 of the main text).
(2)~Fit the threshold at the deployed $N$; no human labels are needed. Fitted to the joint call's answers on twenty unlabelled scenes, the team stays within 1.3 points of the joint call, pooled over $N$, in every setting (four VLMs in the rooms, two on real scans), and the teacher can be cheap: taught by a joint call that sees the 32 images at an 8-image pixel budget, a one-bit team scores 93.3, 2.9 points above its teacher. A four-camera pilot forecasts the OR team's false alarms (Eq.~(1)).
(3)~Gate counts with the fused probability.
(4)~Compare the team with one call that sees every view, calibrated and gated the same way, at matched tokens, and report accuracy on absent objects as cameras are added.

\paragraph{Further limitations.} A joint call free of accumulation is a property of the model, not a guarantee, and the probability results on AV2-PV rest on Qwen3-VL-8B.
Thor-PV removes all other small objects, so false-alarm rates are likely optimistic in clutter.
We test existence and counting, where the fusion rule is explicit; cross-view reasoning may need other messages.
\method needs first-token probabilities at full precision and a threshold per team size and camera layout. At the joint call's false-alarm rate a large one-bit team recalls up to 5.6 points fewer lone witnesses, the noisy-OR more.

\paragraph{The example of Fig.~2.} The values in Fig.~2 of the main text are stored outputs of Qwen3-VL-8B for room 291 of Thor-PV and its four-camera set. For ``apple'' (one instance, seen by the third camera only) the four agents give $P(\text{yes})$ of $<10^{-5}$, $<10^{-5}$, $1-1.7\times10^{-10}$ and $<10^{-5}$ and answer no, no, yes, no; the noisy-OR score is 22.50, far above the threshold of 4.62 fitted on the other four folds at $N=4$, so \method answers yes. The room holds no mug, the second label in panel (b): for ``mug'' the agents give 0.003, $0.732$, $<10^{-4}$ and $<10^{-4}$ and answer no, yes, no, no, the score is 1.32 and \method answers no. The joint call over the four images answers yes for the apple and no for the mug, the teacher's answers in panel (b). Panel (b) shows the fold means of the fitted noisy-OR thresholds, 4.0, 9.7, 21.1 and 43.9 at $N=4$, 8, 16 and 32. Its dots are the noisy-OR scores of all 363 absent questions and of the 1{,}469 to 1{,}709 present questions that at least one camera sees, at each $N$ (scores below 0.1 drawn at 0.1). Against the thresholds fitted on the other four folds, a cut kept at $\tau_4$ is exceeded by 6.1, 14.9, 27.3 and 39.7\% of the absent questions at $N=4$, 8, 16 and 32, whereas the cut fitted at each $N$ is exceeded by 6.1, 3.6, 2.8 and 1.1\%.

\section{Full results}
\label{sec:s-thor}
Table~\ref{tab:s-rules} gives the fixed rules of the main text (no picking among rules on held-out data), by evidence type and with CIs against the joint call and the calibrated joint call, on Thor-PV and AV2-PV.
Tables~\ref{tab:s-thor-8b} and \ref{tab:s-thor-4b} add the richer designs and, for reference, the best of several fusion and vote rules picked on held-out rooms (which favours the team), with token costs.
Table~\ref{tab:s-count} splits counting into kinds with an instance in view of at least one of the $N$ cameras (present) and zero-visible questions (no instance in any of the $N$ views: the kinds the room does not have plus the kinds it has that none of the $N$ cameras sees), and adds the classifier-based teams and the gated joint call.
Three patterns recur.
Answer-sending teams that keep the single witness (OR, route, aggregator) do so at the price of false alarms, and vote thresholds that avoid the false alarms drop exclusive kinds.
On the needle sets (Sec.~\ref{sec:s-testbeds}; exactly one of the $N$ cameras sees each present kind, 1{,}632 present questions at $N=4$ and 671 at $N=32$, with the 363 absent ones) the joint call finds the single witness in 87.4\% of questions at $N=4$ and 82.1\% at $N=32$ while rejecting 90.4 and 93.7\% of absent kinds (8B).
At $N=32$ the OR team keeps 99\% recall but rejects only 46.6\% of absent kinds, and the joint call's lead over it grows from $+1.6$ $[-0.1, +3.3]$ to $+15.2$ $[+12.6, +18.1]$; over the fitted vote threshold it is $+8.4$ $[+5.9, +10.9]$ at $N=32$ (4B: $+7.1$).
The calibrated bit keeps exclusive kinds at every $N$; the noisy-OR gives some of them up (exclusive recall 56.0\% for 8B and 64.0\% for 4B at $N=32$, on 25 questions), in exchange for fewer false alarms for 8B at $N \geq 16$ and for 4B at $N=4$ and 16.
On counting, a gate removes most of the false counts on zero-visible kinds, and against the gated max of the agents' counts the joint call keeps its lead on present kinds.
Against the learned count-and-score classifier, which has no explicit gate, the gated joint call's lead for 8B falls from $+3.4$ at $N=4$ to $-1.5$ $[-4.3, +1.1]$ at 32, and at 16 and 32 cameras it sits on zero-visible kinds ($+6.5$ and $+5.1$) rather than present ones ($-0.1$ and $-3.0$).

\input{tables/s_rules}
\input{tables/s_thor_8b}
\input{tables/s_thor_4b}
\input{tables/s_count}

\paragraph{Other fusion rules.}
Calibrated joint call minus each fusion rule at $N=4, 8, 16, 32$ (all fitted by the same 5-fold room cross-validation).
Qwen3-VL-8B: LO-sum $+3.8, +6.2, +6.4, +7.6$; max-P $-1.6, -0.2, +1.4, +1.8$; noisy-OR $-1.5, -0.3, +0.2, 0.0$; LO-logistic $-0.9, +0.3, +1.2, +0.2$.
Qwen3-VL-4B: LO-sum $+4.7, +7.5, +8.5, +8.9$; max-P $+0.4, 0.0, +1.1, +1.9$; noisy-OR $-0.4, +0.8, +0.7, +1.6$; LO-logistic $+0.5, +0.8, +1.1, +1.4$.
The sum of log-odds, which treats the views as independent evidence, is the worst rule at every $N$: it adds the strongly negative log-odds of the many views that do not see the kind.
Stacking the joint call's log-odds with the agents' (logistic regression, CV by room) adds $+0.80$ $[+0.42, +1.20]$ to the calibrated joint call for 8B and $+0.14$ $[-0.35, +0.62]$ for 4B, pooled over $N$.

\paragraph{Thresholds across $N$.}
The calibrated joint call's threshold rises with $N$ (8B: 0.80, 0.86, 0.94, 0.97; 4B: 0.49, 0.53, 0.77, 0.83), and a threshold fitted at $N=4$ loses at most 0.7 points when applied at larger $N$.
The team's thresholds do not transfer: fitted at $N=4$ and applied at $N=8$, 16 and 32, max-P loses 2.4, 5.3 and 10.2 points (8B; 4B 1.4, 3.6, 6.6) against a threshold fitted at that $N$, and the noisy-OR, whose scale grows with $N$, loses 4.0, 10.0 and 16.5 (4B: gains 0.4, then loses 5.1 and 11.9).
On AV2-PV, from $N=4$ to 7, both lose under one point (max-P 0.2, noisy-OR 0.4, folds grouped by driving log).

\paragraph{How many labelled scenes a threshold needs.}
We refit each threshold on $m$ training rooms drawn at random (40 draws per fold, the five folds of the main text) and score the held-out fold.
Mean loss of balanced accuracy against the fit on all 208 training rooms, for the noisy-OR at $N=4$, 8, 16, 32: with five rooms 3.0, 3.6, 3.2, 3.6 (8B) and 2.7, 2.2, 3.1, 2.5 (4B); with ten 1.3, 1.7, 1.5, 1.9 and 1.6, 0.7, 1.7, 0.7; with twenty 0.8, 1.1, 0.9, 1.0 and 0.8, 0.0, 1.1, 0.3; with forty at most 0.6.
With twenty rooms the calibrated bit loses at most 1.0 points and the calibrated joint call, whose threshold is fitted the same way, at most 0.9.
On ScanNet twenty scans cost the noisy-OR at most 0.5 points and the bit at most 0.6 (Qwen3-VL-8B and LLaVA-OneVision-7B).
With twenty labelled rooms the noisy-OR at 32 cameras scores 95.5 on average (8B; 94.5 to 96.3 between the 5th and 95th percentile of the draws), against 73.3 for the OR team, which needs no labels.

\paragraph{A threshold from the four-camera model.}
We also tried to avoid labels at the deployed $N$: fit the beta-binomial of Eq.~1 of the main text (pooled over kinds here) to the number of views with $p_i \geq \tau$ among the first four cameras, for each candidate $\tau$, and take for each $N$ the smallest $\tau$ whose predicted accuracy on absent kinds reaches that of the four-camera pilot.
On ScanNet the calibrated bit with this threshold stays within 1.2 points of the per-$N$ fit (8B: $-0.5$ and $-1.2$ $[-2.5, +0.0]$ at 8 and 16 frames; LLaVA-OneVision: $-0.2$ and $-0.3$).
On Thor-PV it does not: the threshold becomes too strict at large $N$ and loses 7.0 points at $N=32$ for 8B and 13.6 at $N=16$ and 32 for 4B, where the beta-binomial and the independence prediction pick the same threshold, which suggests that the four-camera fit over-predicts false alarms in the far tail of the scores.
The paper therefore fits every threshold at the deployed $N$, on labelled scenes unless stated (Sec.~\ref{sec:s-more} fits it without labels).

\paragraph{Message precision.}
How coarse can the message be? We quantise each $p_i$ before fusion and refit the thresholds (Thor-PV, both models, the folds of the main text).
Log-odds rounded to one decimal cost \method at most 0.1 points and the calibrated bit at most 0.3 at every $N$; rounded to an integer, at most 0.5 and 0.7.
A probability written to two decimals is too coarse, because at large $N$ the fitted thresholds sit above 0.99: at $N=32$ the calibrated bit loses 4.9 $[3.3, 6.5]$ points for 8B and 9.0 $[6.9, 11.3]$ for 4B, and \method 0.9 for 8B (1.5 for 4B at $N=16$); with three decimals the bit still loses 1.4 and 4.5 at $N=32$.
An agent should therefore send its log-odds, for which one decimal is enough.
Differences of about half a point between quantisations are threshold-fitting noise (with four decimals the 4B noisy-OR gains 1.0 at $N=32$).

\paragraph{Lone witness for the probability team.} At $N=16$ (Table~4 of the main text): on the main sets the noisy-OR's threshold, fitted where most present kinds are shared, misses lone witnesses; refitted on lone-witness sets it recovers their recall (86.0\%, 8B) at the cost of false alarms (absent 84.6\%).
The needle sets of the main text draw cameras from all 64 of a room, but per-view probabilities were at first stored only for the first 32 (Sec.~\ref{sec:s-more} adds the rest), so here we rebuild lone-witness sets from those 32: for each present kind one camera that sees it and $N-1$ that do not, and $N$ random cameras for absent kinds (1{,}590, 1{,}444, 1{,}019 and 25 present questions at $N=4$, 8, 16 and 32), and run the joint call on the same sets (Table~\ref{tab:s-lone}).
For 8B the joint call and the calibrated bit stay within a point of each other up to 16 cameras; at 32 the joint call leads ($+4.5$ $[+1.3, +9.3]$, 25 present questions), while its calibrated version does not ($+0.6$ $[-5.2, +6.4]$); on the 671 present questions of Sec.~\ref{sec:s-more} the lead is not resolved ($+0.7$ $[-1.0, +2.3]$).
Against the noisy-OR the 8B joint call leads from eight cameras on, and at the noisy-OR's own false-alarm rate its recall is higher at every $N$.
For 4B the joint call is conservative on lone witnesses (recall 76 to 82\%): it trails the calibrated bit at four and eight cameras ($-2.3$ $[-3.9, -0.6]$ and $-3.9$ $[-5.6, -2.3]$) and leads the noisy-OR only at 32 ($+6.1$ $[+0.4, +12.1]$; at 16 cameras $+0.4$ $[-1.3, +2.3]$).

\input{tables/s_lone}

\paragraph{Probabilities on real scans.}
For Qwen3-VL-8B and LLaVA-OneVision-7B we re-collected per-view and joint probabilities on the ScanNet questions with the prompts of the answer runs (Table~\ref{tab:s-scanprob}).
Between 4 and 16 frames the OR team's accuracy on absent kinds falls from 83.1 to 69.3\% (8B) and from 84.7 to 71.9\% (LLaVA-OneVision).
The calibrated bit stays within one point of the calibrated joint call at every $N$ for both models; so does the noisy-OR for 8B (pooled $-0.1$ $[-0.8, +0.6]$) and, for LLaVA-OneVision, at 4 and 8 frames, while at 16 frames it trails by 2.0 $[-3.1, -0.8]$ (pooled $-1.1$ $[-1.9, -0.4]$).

\input{tables/s_scanprob}

\input{tables/s_auc}
The joint call's AUC lead on the 25 exclusive questions at $N=32$ (Table~\ref{tab:s-auc}) is larger on the earlier tagged renders, where the tags give the joint call a cross-view identity cue: $+0.5$ $[+0.2, +0.8]$ on all present kinds and $+3.9$ $[+0.6, +7.7]$ on exclusive ones at $N=32$ for 8B.

\section{The model of Eq.~(1)}
\label{sec:s-law}
Let $X_1, \dots, X_N$ be the flags of the $N$ views for one absent kind in one room.
If $X_i \mid p \sim \mathrm{Bernoulli}(p)$ independently and $p \sim \mathrm{Beta}(\alpha, \beta)$, the number of flags is beta-binomial and
$P(\sum_i X_i = 0) = \int_0^1 (1-p)^N \,\mathrm{Beta}(p; \alpha, \beta)\, dp = B(\alpha, \beta+N)/B(\alpha, \beta) = \Gamma(\alpha+\beta)\Gamma(\beta+N)/(\Gamma(\beta)\Gamma(\alpha+\beta+N))$, which behaves as $\Gamma(\alpha+\beta)/\Gamma(\beta) \cdot N^{-\alpha}$ only once $N \gg \beta$ (with the pooled 8B fit, $\beta = 7.5$, the asymptotic form is far off at $N \leq 32$; we always use the exact form).
The mean flag rate is $\alpha/(\alpha+\beta)$ and the correlation between two views' flags is $1/(\alpha+\beta+1)$.
We fit $(\alpha, \beta)$ per kind by maximum likelihood (L-BFGS-B on log-parameters, bounds $e^{-6}$ to $e^{6}$ for $\alpha$ and $e^{8}$ for $\beta$) on the flag counts out of the first four cameras of the training rooms' absent questions, and forecast the held-out rooms' accuracy at every $N$ (5-fold CV by room; by driving log on AV2-PV, which keeps the two sweeps of a log together).
With 27 to 66 absent questions per kind and four flags each, the fit is weakly identified: on Thor-PV, fitting all absent questions of a kind puts 3 of 8 kinds (8B) and 4 of 8 (4B) at the bound, and the five training-fold fits that produce the forecasts put 2, 3, 2, 2, 2 (8B) and 2, 2, 4, 4, 2 (4B) there; at the bound the beta-binomial approaches the binomial, and the forecast, which we do not replace by the independence one, stays within half a point of it.
The independence model uses the same data and predicts $(1-f)^N$ with $f$ the per-kind flag rate.
Pooled over kinds, Thor-PV gives $\alpha = 0.51$, $\beta = 7.5$ for Qwen3-VL-8B (flag rate 6.4\% on the first four cameras, correlation 0.11) and $\alpha = 0.34$, $\beta = 5.2$ for 4B (6.1\%, 0.15); over all 32 cameras the 8B flag rate is 7.0\%.
Table~\ref{tab:s-law} lists every run.
Over the 15 untagged settings with $N > 4$ the mean absolute error is 2.6 points (independence 11.0), the largest error 6.7, and the observed accuracy is above the forecast in 14 of 15: the model slightly overstates the decay.
On ScanNet the four fitting frames are spread over the scan, while the frames added at $N=8$ and 16 lie between them and may share more of the view, which would make their flags more correlated than the fit assumes; we do not measure that dependence directly.
On the tagged renders the model fits Idefics3 and Mistral within 2.4 points; for GPT-5.4-mini, whose single views flag 45\% of absent kinds, the OR team is near the floor, the model over-forecasts by 6, and the independence model is closer.

\input{tables/s_law}

\section{AV2-PV: wording, calibrated baselines and counting}
\label{sec:s-av2}
\paragraph{Wording.} Our first AV2-PV joint call asked ``Is there $\langle K\rangle$ anywhere around the car?''.
That scope is wider than the single views' ``in this image'', and the joint call was more conservative: balanced accuracy 80.8 and 77.8 at $N=4$ and 7 (exclusive recall 64.8 and 54.1\%), against 81.8 and 79.2 with ``in any of these images''; its counting accuracy was 71.0 and 64.4\%, against 66.5 and 59.8\%.
The order of the existence results is the same: probability fusion (noisy-OR at both $N$, the best of four fusion rules by held-out balanced accuracy, with folds and bootstrap grouped by driving log) leads that joint call by $+2.3$ $[+0.6, +4.1]$ and $+4.2$ $[+2.6, +5.9]$, and its calibrated version by $+0.5$ $[-0.8, +1.9]$ and $+0.6$ $[-0.7, +1.8]$.
The main text uses the second wording, whose scope matches the single views; the single-view prompts were not changed.

\paragraph{Calibrated baselines.} With the second wording (Table~\ref{tab:s-rules}), the noisy-OR minus the calibrated joint call is $+0.9$ $[-0.7, +2.4]$ and $+0.9$ $[-0.3, +2.1]$ at $N=4$ and 7, pooled $+0.85$ $[-0.24, +2.04]$; minus the OR team it is $+0.4$ $[-1.1, +1.7]$ and $+1.2$ $[-0.1, +2.5]$ (folds and bootstrap group the two sweeps of a log, as in the main text).
Threshold-free, the noisy-OR's AUC is 91.2 and 89.9 against 91.2 and 89.3 for the joint call.

\paragraph{Counting.} At $N=7$, 75\% of AV2-PV counting questions have no instance in view (truth 0), so exact counting accuracy is dominated by saying 0: a kind-wise modal count, fitted on training logs, scores 75.1\% at $N=7$, ahead of the joint call (59.8\%), the max of counts (45.8\%) and a gradient-boosted count classifier (65.7\%, folds grouped by log).
With a gate on each side (the gated joint call against the gated max of the agents' counts), the interval does not resolve the difference at $N=4$ ($-0.3$ $[-1.6, +1.1]$) and the joint call leads at $N=7$ ($+2.2$ $[+1.0, +3.5]$; 68.6 against 66.3\%); on present kinds the max of the agents' counts is right for 48.5 and 46.5\% at $N=4$ and 7, against 42.1 and 38.0\% for the joint call.

\section{Additional experiments}
\label{sec:s-more}
\input{tables/s_new}

\paragraph{What a practitioner would try first.}
Table~\ref{tab:s-protocols} adds four families of alternatives for both Qwen3-VL models, with the cameras, questions and folds of the main text.
(i)~A conservative prompt (``answer yes only if you can clearly see one''), merged by OR or by a vote threshold fitted per $N$: agents raise about 40\% fewer false alarms and the OR still declines with $N$.
(ii)~A verbal confidence between 0 and 100 as the message, fused by the noisy-OR or the max under a fitted threshold.
(iii)~Two rounds of debate with one view per agent (majority after the second round; $N=4$ and 16).
(iv)~A text aggregator that reads one line per camera: the answers, the probabilities, both, or the log-odds; for the probabilities also a reader that is told how to read them, is shown eight labelled examples from the other folds, or reasons in three sentences before answering. Writing the agents' uncertainty into the prompt of a language model is the pattern of DebUnc~\cite{yoffe2025debunc}, applied here without its debate rounds.
None of them reaches the fitted rules. For 8B the fitted vote over conservative answers comes closest, but a vote threshold above one cannot keep a lone witness; for 4B the noisy-OR of verbal confidences comes closest, within three points of \method.
A reader four times larger does not close the gap: Qwen3-VL-32B reading the 8B agents' probabilities scores 88.3, 80.9, 76.4 and 70.8\% at $N=4$ to 32 as a plain reader, 92.0, 87.8, 82.9 and 77.8\% when told how to read them, and 91.2, 88.4, 89.0 and 88.2\% with eight labelled examples; with a threshold fitted on its own probability, its best arm stays 4.8 points below \method pooled over $N$ ($[-5.6, -4.0]$).

\paragraph{Fusion rules of prior work.}
\input{tables/s_prior}
Table~\ref{tab:s-prior} applies the decision rules of prior multi-agent and classifier-combination work to the same single-view agents, with no new model call beyond those of Table~\ref{tab:s-protocols}.
\emph{Majority vote}: yes when more than half of the agents answer yes, the baseline of multi-agent debate~\cite{du2024improving} and the readout of several agent systems.
\emph{Optimal fusion of hard decisions}~\cite{chair1986optimal}: the log-likelihood-ratio sum of the agents' answers under independence, with the per-view hit and false-alarm rates estimated on the training folds at each $N$ and equal priors.
\emph{Debate}~\cite{du2024improving}: two rounds in which every agent sees its image and the other agents' replies, read out by majority (the paper states no readout rule and summarises the other agents when there are five or more; we pass every reply).
\emph{Confidence-weighted vote}~\cite{chen2024reconcile}: each agent's verbal confidence is recalibrated with ReConcile's table (1.0 to 1.0; $[0.9, 1.0)$ to 0.8; $[0.8, 0.9)$ to 0.5; $(0.6, 0.8)$ to 0.3; otherwise 0.1) and the answer with the larger total weight wins; this is ReConcile's final rule with one model family and without its discussion rounds and convincing samples.
\emph{Sum, product, median, max and min rules}~\cite{kittler1998combining} on the agents' first-token probabilities, with the usual cut at 0.5 and with a threshold fitted per $N$.
Rules that count agreeing agents keep the accuracy on absent kinds near 99\% and lose a third or more of the present kinds, since most cameras do not see the object; the max rule with the cut 0.5 is the OR.
Every rule is below \method pooled over $N$: by 3.5 and 4.0 points for the best of them, the sum rule with a fitted threshold, and by 4.7 and 4.4 for the optimal fusion of hard decisions.
\input{tables/s_prior_real}
Table~\ref{tab:s-prior-real} repeats the comparison on the real testbeds for the runs that stored per-view probabilities (no verbal confidences were collected there, so the weighted vote uses first-token probabilities).
On ScanNet every prior rule is below \method for both models, by 1.7 to 24 points pooled over $N$; the closest is again the sum rule with a fitted threshold ($-2.4$ $[-3.3, -1.6]$ for Qwen3-VL-8B, $-1.7$ $[-2.3, -1.0]$ for LLaVA-OneVision-7B).
On the seven-camera rig, where few cameras share an object, \method is ahead of every prior rule in the point estimate: the fitted sum rule and the OR are within one point ($-0.7$ $[-1.3, +0.1]$ and $-0.8$ $[-1.9, +0.4]$), and the other rules are 4 to 24 points below.
Intervals for Table~1 of the main text: the joint call minus $k$ of $N$ at $N=32$ is $+5.0$ $[+3.5, +6.6]$ (4B), and the gated joint call minus the gated max at $N=32$ is $+1.8$ $[0.0, +3.6]$ (8B) and $+5.3$ $[+3.1, +7.3]$ (4B); at $N=32$ route-then-verify and the aggregator use 38.0k and 28.0k tokens per question (8B).

\paragraph{Fixed, common and per-$N$ cuts.}
Table~\ref{tab:s-cuts} applies the max rule with fixed cuts, with one cut optimised for all $N$ on the training folds, and with the per-$N$ fit. The cut 0.5 is nearly the OR of greedy answers (the two agree on 99.6\% of view calls); no fixed cut is right at every team size, and for Qwen3-VL the per-$N$ fit is worth up to 2.6 points over the best common cut (8B, $N=32$). The fitted cut itself rises from $1-2.9\times10^{-2}$ at four cameras to $1-9.7\times10^{-5}$ at 32 for 8B and from $1-9.1\times10^{-4}$ to $1-1.2\times10^{-6}$ for 4B (median over folds).

\paragraph{Thresholds without labels.}
In Table~\ref{tab:s-labelfree} the threshold is fitted to the joint call's answers on the training scenes instead of labels, with no ground truth on the calibration side: on all training scenes, on twenty scenes (40 draws), and, for 8B at $N=32$, on twenty scenes with a joint call that sees the images at a 16-, 8- or 4-image pixel budget. Taught by the 8-image-budget call (90.4), the one-bit team scores 93.3. The pixel-budget comparison of the main text measures what the joint call loses at low resolution, not an advantage of the team at equal cost.
Pooled over $N$, the team fitted on twenty unlabelled scenes stays within 1.3 points of the greedy joint call in all six settings (noisy-OR: $+0.1$, $-1.2$, $-1.2$ and $+0.4$ for Qwen3-VL-8B, 4B, InternVL3-8B and Idefics3-8B in the rooms, $-0.8$ and $-1.2$ on real scans; bit: $-0.4$, $-1.3$, $+0.3$, $+0.9$, $-0.5$ and $-0.6$).

\paragraph{Lone witness at scale.}
With per-view probabilities collected for all 64 cameras of a room, the lone-witness sets of the main text can be scored for the probability team at every $N$ (Table~\ref{tab:s-needle}; 1{,}632 to 671 present questions, against 25 at $N=32$ in Table~\ref{tab:s-lone}). With the thresholds of the ordinary questions the calibrated bit is never more than 1.8 points below the joint call, for both models. The last columns give the recall of lone witnesses when each rule's threshold is matched to the joint call's false-alarm rate on the training folds.

\paragraph{Other model families.}
Table~\ref{tab:s-families} repeats the rooms with InternVL3-8B (one 448-pixel tile per image, 256 visual tokens, for agents and joint call alike) and Idefics3-8B-Llama3 (longest edge 728 pixels), with the questions, cameras and folds of the main text. For both, the OR team decays, while the noisy-OR stays between 82.6 and 84.2 for InternVL3 and rises from 84.0 to 86.5 for Idefics3, where the fitted rules are above the joint call at every $N$.
The table also holds Qwen3-VL-32B, four times the size of 8B (tensor-parallel on two GPUs; a 32-image joint call is a 24{,}679-token prompt): its agents say yes on 7.5\% of absent-kind views (8B: 7.0\%), its OR team falls from 89.4 to 74.7\%, and \method holds 94.0 to 95.0, 12.6 points above the OR team pooled over $N$ ($[+10.7, +14.6]$) and $-1.2$ $[-2.1, -0.3]$ against the 32B joint call.
Table~\ref{tab:s-law9} scores the beta-binomial forecast of the main text on these three models: fitted on four cameras, it is within 2.1 points of the OR team's accuracy on absent kinds at every $N$ (mean absolute error 1.3, against 8.9 for independent views). Together with the 15 settings of the main text, which share models, scenes and nested camera sets and are therefore not independent replications, this makes 24; in 14 of those 15 the forecast is below the observed value, consistent with added views sharing more of the scene than the four it was fitted on (we do not measure that dependence directly), while on these nine points the sign is mixed.

\paragraph{A public benchmark.}
\input{tables/s_pope}
\input{tables/s_pope_n}
The testbeds of the main text are our own question sets. To test \method on questions that we did not write, we use 3D-POPE~\cite{yang20253dgrand}, which asks whether an object is in a ScanNet room (``Are there any desks in the room?'') for 140 validation scenes, with as many absent as present objects and with absent objects drawn at random, by popularity or adversarially.
The benchmark targets 3D-LLMs that read the reconstructed scene; we answer its questions from frames, with the protocol of our ScanNet testbed: evenly spaced frames of the public 25k-frame subset in nested sets of 4, 8 and 16, one agent per frame that is asked the benchmark's question about its image (``\dots\ in this image?''), and a joint call that receives all $N$ frames and the question verbatim.
Of the 140 scenes, 139 have at least four frames in the subset, 127 at least eight and 61 at least sixteen; thresholds are fitted per $N$ with five folds by scene. We run Qwen3-VL-8B and 4B.

Table~\ref{tab:s-pope-n} gives the 61 scenes that have all three team sizes (its balanced accuracies are Table~2 of the main text). The joint call gains 6.4 points of balanced accuracy between 4 and 16 frames for 8B and 6.3 for 4B, whereas the OR team gains nothing, because its accuracy on absent objects falls from 66.9 to 49.3\% (4B: 68.1 to 49.5\%).
\method follows the joint call ($-0.0$ $[-1.6, +1.5]$, $+1.4$ $[+0.1, +2.7]$ and $-0.1$ $[-1.5, +1.4]$ at $N=4$, 8 and 16 for 8B; $+1.6$ $[+0.5, +2.6]$, $+0.9$ $[-0.4, +2.0]$ and $-0.2$ $[-1.3, +1.0]$ for 4B) and at sixteen frames leads the OR team by 7.0 points $[+5.2, +8.5]$ and the fitted vote by 5.3 $[+3.8, +6.8]$ (4B: 6.3 $[+4.7, +8.1]$ and 2.9 $[+1.2, +4.5]$); the calibrated bit is 1.7 points above the joint call there $[+0.2, +3.2]$ (4B: 1.5 $[+0.1, +2.8]$).
A threshold fitted to the joint call's answers instead of labels gives 74.5, 78.0 and 80.6\% (4B: 74.9, 78.9 and 80.4\%).
On this benchmark the joint call's own accuracy on absent objects declines with the number of frames (73.2 to 68.0\% for 8B, 74.2 to 69.0\% for 4B) and that of \method declines with it (75.9 to 68.3\% and 83.2 to 71.8\%), so the result here is parity with the joint call.

In the benchmark's own metrics (Table~\ref{tab:s-pope}), a team of eight single-frame agents of a frozen model reaches 87.2\% (8B) and 87.1\% (4B) accuracy on the ScanNet200 questions with random absent objects (127 scenes), against 50 to 53\% for 3D-LLM, 3D-VisTA and LEO and 89.1\% for the 3D-GRAND model, which is trained on 3D data and reads the whole scene (140 scenes).
The two groups differ in input and in scene coverage, so the comparison is indicative only.

\paragraph{Sampled answers.}
\input{tables/s_samp}
Self-consistency~\cite{wang2023selfconsistency} is the usual way to obtain a confidence from answers alone. Table~\ref{tab:s-samp} samples five answers per agent at temperature 0.7 for the first 16 cameras of the rooms of Table~1 of the main text.
The samples rarely differ: all five agree on 96\% of the views for Qwen3-VL-8B and on 97\% for 4B, and an agent's majority answer is yes for 6.7 to 7.5\% of the absent-kind views, against 6.1 to 6.8\% for its greedy answer, so that the false alarms are stable errors of the model rather than sampling noise.
Accordingly the OR of the majority answers declines like the OR of the greedy answers, and the best rule on sampled answers, the noisy-OR of the agents' yes shares with a fitted threshold, reaches 91.6, 90.4 and 92.5\% at $N=4$, 8 and 16 for 8B and 90.9, 92.0 and 92.9\% for 4B, which is 2.9 $[+1.5, +4.5]$, 5.3 $[+3.9, +6.7]$ and 3.5 $[+2.5, +4.5]$ points below \method for 8B and 4.7 $[+2.9, +6.5]$, 3.6 $[+2.2, +5.0]$ and 3.8 $[+2.6, +5.2]$ for 4B, at five times the calls.
A majority over three sampled joint calls changes the 8B joint call by less than one point.
On 3D-POPE the picture is the same (Table~\ref{tab:s-pope-n}): at sixteen frames the best rule on sampled answers reaches 76.1\% (8B) and 77.7\% (4B), against 80.5 and 80.2\% for \method and 82.3 and 81.8\% for the calibrated bit.

\section{Analysis protocol}
\label{sec:s-protocol}
The comparisons of the joint call with yes/no teams, richer teams and probability fusion on Thor-PV and AV2-PV were specified before the model runs.
The model of the decay (Sec.~5.3 of the main text), the threshold-free comparison, the calibrated bit, the gated counting teams, the lone-witness sets and the fixed-rule tables were run afterwards on the saved model outputs, with no new model calls; before each round we wrote down the expected outcome.
Two later rounds made new model calls with unchanged prompts: per-view and joint probabilities on ScanNet for two models (Table~\ref{tab:s-scanprob}) and the joint call on the lone-witness sets (Table~\ref{tab:s-lone}); their expected outcomes were also written down before the runs, and two expectations failed in the team's favour. We expected the joint call to be within three points of the calibrated bit at 16 cameras (it is: $+0.7$ for 8B, $-1.5$ for 4B), to recall at least as many lone witnesses as the bit at 32 cameras, and the noisy-OR to recall about half of them there. For 4B, on those sets, the joint call recalls fewer than the bit at 32 (76.0 against 88.0\%; it also trails the bit by 3.9 at eight cameras), and the noisy-OR recalls 88.0\% (8B) and 76.0\% (4B).
A final round (Sec.~\ref{sec:s-more}) made new model calls with new prompts (the conservative prompt, verbal confidence, debate and the text readers), collected per-view probabilities for all 64 cameras and the joint call under pixel budgets, and ran two further model families and a larger model; the expected outcome of each of these model runs was written down before it.
Per-view counts are clipped to $[0, 10]$ in the gated and learned counting teams. Three expectations failed and changed the paper: we expected the team's AUC to fall with $N$, and on Thor-PV it does not (Table~\ref{tab:s-auc}; on AV2-PV the noisy-OR's falls from 91.2 to 89.9 between four and seven cameras, as the joint call's falls from 91.2 to 89.3); we expected the independence model to describe the false alarms within 8 points, and it misses by 8 to 13 at $N \geq 16$, which led to the beta-binomial; and we expected a max-P threshold fitted at four cameras to lose at most 8 points at 32, and it loses 10.2 for 8B.
After a first draft we fixed every rule before further analyses, gated the joint call's counts as well as the team's, and reported AV2-PV against the calibrated joint call; this lowered the reported advantage of the probability team on counting and on AV2-PV.
Intervals are 95\% paired bootstrap intervals over rooms and scans with 2{,}000 resamples (1{,}000 for counting, the paired lone-witness sets, single-$N$ intervals on ScanNet and absent-kind changes; 300 for AUC).
On AV2-PV the two sweeps of a log are one cluster: every cross-validation fold, fitted threshold, count classifier and beta-binomial fit in the main text and this supplement keeps them together, and every interval resamples logs.
The ScanNet analyses of the model and all lone-witness numbers are exploratory.
The noisy-OR and the maximum were retained after comparing four fusion rules on these data (Sec.~\ref{sec:s-thor}), so the choice of rule is not a held-out test; every threshold within a rule is fitted on training rooms only.

Per-view probabilities were stored for all runs on Thor-PV and for Qwen3-VL-8B on AV2-PV and re-collected for Qwen3-VL-8B and LLaVA-OneVision-7B on ScanNet; the other ScanNet runs kept answers only.

\paragraph{Probability clip.}
All probabilities were first clipped to $[10^{-4}, 1-10^{-4}]$.
A later check found that 28\% (Qwen3-VL-8B) and 37\% (4B) of the Thor-PV view scores, and 65 to 66\% of the joint call's scores (both models), lie above that ceiling, so the clip tied the most confident scores of both sides.
We recomputed every probability result in the paper with a numerical floor of $10^{-12}$ only, with the same folds, seeds and fitting.
Pooled over $N$, the team's thresholded results rose (single cells moved by up to 0.4 points either way): the noisy-OR minus the joint call, pooled over $N$ on Thor-PV, went from $+0.4$ $[-0.3, +1.1]$ to $+0.7$ $[+0.1, +1.4]$ for 8B and from $-1.3$ $[-2.1, -0.4]$ to $-0.7$ $[-1.4, +0.1]$ for 4B, the 4B calibrated bit at $N=32$ from 93.3 to 96.0, and the 8B noisy-OR on ScanNet at 16 frames from 91.4 to 92.9.
The calibrated joint call's thresholded results did not move (its thresholds lie between 0.49 and 0.97), and the threshold-free AUCs changed by under 0.3 points on present-and-seen kinds in the rooms and rose more for the joint call than for the team on ScanNet (8B) and AV2-PV.

%% file: tables/s_scannet.tex
\begin{table}[t]
\centering
\footnotesize
\setlength{\tabcolsep}{2.2pt}
\begin{tabular}{@{}lccccc@{}}
\toprule
 & \multicolumn{3}{c}{OR team, absent} & Joint call & Joint $-$ $k$ of $N$ \\
\cmidrule(lr){2-4}
\textbf{Model} & $N{=}4$ & $N{=}16$ & law & absent & at $N{=}16$ \\
\midrule
Qwen3-VL-8B   & 83.1 & 69.3 & 64.0 & 91.4\,$\to$\,89.2 & $+3.4$ {\scriptsize$[+2.0, +4.9]$} \\
Qwen3-VL-4B   & 83.1 & 67.5 & 66.0 & 90.8\,$\to$\,90.2 & $+3.3$ {\scriptsize$[+2.0, +4.7]$} \\
LLaVA-OV-7B   & 84.7 & 71.9 & 66.0 & 91.6\,$\to$\,89.4 & $+2.0$ {\scriptsize$[+0.6, +3.2]$} \\
Qwen2.5-VL-7B & 87.5 & 75.2 & 68.5 & 95.3\,$\to$\,96.7 & $-3.7$ {\scriptsize$[-6.1, -1.5]$} \\
\bottomrule
\end{tabular}
\vspace{-4pt}
\caption{\textbf{On real scans, every model's OR team loses 12 to 16 points on absent kinds from 4 to 16 frames; the joint call moves by at most 2.2.}
1{,}799 generated questions.
``Law'': Eq.~(1) of the main text fitted on four-frame flags of training scans, predicting $N=16$ on held-out scans.
Last column: balanced accuracy, joint call minus ``at least $k$ of $N$ say yes'' ($k$ fitted by 5-fold CV by scan; 95\% scan-bootstrap CI).}
\label{tab:s-scannet}
\vspace{-6pt}
\end{table}

%% file: tables/s_rules.tex
\begin{table*}[p]
\centering
\scriptsize
\setlength{\tabcolsep}{3.2pt}
\renewcommand{\arraystretch}{0.9}
\begin{tabular}{rlrrrrll}
\toprule
$N$ & Rule (fixed) & Balanced & Exclusive & Shared & Absent & minus joint call & minus calibrated joint call \\
\midrule
\multicolumn{8}{l}{\emph{Thor-PV, Qwen3-VL-8B}} \\
4 & Joint call & 93.5 & 86.8 & 97.2 & 92.6 & --- & $+0.5$ {\scriptsize[$-0.5$, $+1.4$]} \\
 & Calibrated joint call & 93.0 & 82.0 & 95.6 & 93.9 & $-0.5$ {\scriptsize[$-1.4$, $+0.5$]} & --- \\
 & OR of yes/no & 89.1 & 94.2 & 99.6 & 79.9 & $-4.5$ {\scriptsize[$-6.4$, $-2.6$]} & $-4.0$ {\scriptsize[$-5.9$, $-2.0$]} \\
 & $k$ of $N$ & 88.4 & 59.0 & 95.7 & 90.6 & $-5.1$ {\scriptsize[$-6.6$, $-3.4$]} & $-4.6$ {\scriptsize[$-6.0$, $-3.0$]} \\
\rowcolor{oursbg} & Calibrated bit (max-P) & 94.6 & 88.1 & 97.1 & 94.5 & $+1.1$ {\scriptsize[$-0.0$, $+2.3$]} & $+1.6$ {\scriptsize[$+0.6$, $+2.7$]} \\
\rowcolor{oursbg} & Noisy-OR & 94.6 & 87.3 & 97.9 & 93.9 & $+1.0$ {\scriptsize[$+0.0$, $+2.1$]} & $+1.5$ {\scriptsize[$+0.5$, $+2.6$]} \\
\addlinespace[2pt]
8 & Joint call & 94.8 & 87.2 & 96.3 & 94.2 & --- & $-0.7$ {\scriptsize[$-1.6$, $+0.1$]} \\
 & Calibrated joint call & 95.4 & 86.0 & 95.5 & 96.4 & $+0.7$ {\scriptsize[$-0.1$, $+1.6$]} & --- \\
 & OR of yes/no & 84.3 & 98.3 & 99.5 & 69.1 & $-10.5$ {\scriptsize[$-12.7$, $-8.3$]} & $-11.2$ {\scriptsize[$-13.3$, $-8.9$]} \\
 & $k$ of $N$ & 89.8 & 55.3 & 94.5 & 89.5 & $-4.9$ {\scriptsize[$-6.6$, $-3.2$]} & $-5.6$ {\scriptsize[$-7.3$, $-3.9$]} \\
\rowcolor{oursbg} & Calibrated bit (max-P) & 95.7 & 84.9 & 96.2 & 96.4 & $+0.9$ {\scriptsize[$-0.2$, $+2.1$]} & $+0.2$ {\scriptsize[$-0.6$, $+1.1$]} \\
\rowcolor{oursbg} & Noisy-OR & 95.7 & 80.4 & 96.8 & 96.4 & $+0.9$ {\scriptsize[$-0.1$, $+2.1$]} & $+0.3$ {\scriptsize[$-0.5$, $+1.1$]} \\
\addlinespace[2pt]
16 & Joint call & 95.6 & 85.5 & 96.4 & 95.3 & --- & $-0.6$ {\scriptsize[$-1.6$, $+0.2$]} \\
 & Calibrated joint call & 96.3 & 76.8 & 94.7 & 98.6 & $+0.6$ {\scriptsize[$-0.2$, $+1.6$]} & --- \\
 & OR of yes/no & 80.2 & 100.0 & 99.8 & 60.6 & $-15.4$ {\scriptsize[$-17.7$, $-13.2$]} & $-16.1$ {\scriptsize[$-18.3$, $-13.8$]} \\
 & $k$ of $N$ & 91.1 & 52.2 & 93.6 & 90.4 & $-4.5$ {\scriptsize[$-6.2$, $-2.7$]} & $-5.1$ {\scriptsize[$-6.7$, $-3.6$]} \\
\rowcolor{oursbg} & Calibrated bit (max-P) & 94.9 & 87.0 & 96.3 & 93.9 & $-0.7$ {\scriptsize[$-1.8$, $+0.4$]} & $-1.4$ {\scriptsize[$-2.5$, $-0.2$]} \\
\rowcolor{oursbg} & Noisy-OR & 96.0 & 72.5 & 95.8 & 97.2 & $+0.4$ {\scriptsize[$-0.7$, $+1.5$]} & $-0.2$ {\scriptsize[$-1.1$, $+0.6$]} \\
\addlinespace[2pt]
32 & Joint call & 95.9 & 88.0 & 97.0 & 95.0 & --- & $-0.6$ {\scriptsize[$-1.6$, $+0.3$]} \\
 & Calibrated joint call & 96.6 & 84.0 & 94.4 & 98.9 & $+0.6$ {\scriptsize[$-0.3$, $+1.6$]} & --- \\
 & OR of yes/no & 73.3 & 100.0 & 100.0 & 46.6 & $-22.7$ {\scriptsize[$-25.2$, $-20.2$]} & $-23.3$ {\scriptsize[$-25.7$, $-20.9$]} \\
 & $k$ of $N$ & 91.9 & 48.0 & 91.5 & 92.8 & $-4.1$ {\scriptsize[$-5.6$, $-2.5$]} & $-4.7$ {\scriptsize[$-6.4$, $-3.1$]} \\
\rowcolor{oursbg} & Calibrated bit (max-P) & 94.8 & 84.0 & 94.5 & 95.3 & $-1.1$ {\scriptsize[$-2.4$, $+0.2$]} & $-1.8$ {\scriptsize[$-2.9$, $-0.5$]} \\
\rowcolor{oursbg} & Noisy-OR & 96.6 & 56.0 & 94.8 & 98.9 & $+0.6$ {\scriptsize[$-0.4$, $+1.8$]} & $-0.0$ {\scriptsize[$-0.9$, $+0.9$]} \\
\addlinespace[2pt]
\multicolumn{8}{@{}p{0.97\textwidth}@{}}{\scriptsize Pooled over $N$ (95\% / 90\% CI): noisy-OR $-$ joint $+0.75$ [$+0.06$, $+1.41$] / [$+0.19$, $+1.30$]; noisy-OR $-$ calibrated joint $+0.38$ [$-0.16$, $+0.91$] / [$-0.05$, $+0.83$]; bit $-$ joint $+0.04$ [$-0.67$, $+0.77$] / [$-0.54$, $+0.67$]; bit $-$ calibrated joint $-0.32$ [$-1.00$, $+0.30$] / [$-0.88$, $+0.22$]; $k$ of $N$ $-$ joint $-4.65$ [$-5.81$, $-3.44$] / [$-5.59$, $-3.65$]} \\
\midrule
\multicolumn{8}{l}{\emph{Thor-PV, Qwen3-VL-4B}} \\
4 & Joint call & 95.3 & 80.2 & 98.2 & 97.0 & --- & $+0.1$ {\scriptsize[$-0.1$, $+0.4$]} \\
 & Calibrated joint call & 95.2 & 80.7 & 98.2 & 96.7 & $-0.1$ {\scriptsize[$-0.4$, $+0.1$]} & --- \\
 & OR of yes/no & 90.1 & 95.5 & 99.7 & 81.5 & $-5.2$ {\scriptsize[$-7.1$, $-3.3$]} & $-5.1$ {\scriptsize[$-7.0$, $-3.3$]} \\
 & $k$ of $N$ & 89.6 & 63.5 & 97.0 & 90.9 & $-5.6$ {\scriptsize[$-7.2$, $-4.0$]} & $-5.5$ {\scriptsize[$-7.1$, $-3.9$]} \\
\rowcolor{oursbg} & Calibrated bit (max-P) & 94.8 & 85.4 & 97.3 & 95.3 & $-0.5$ {\scriptsize[$-1.8$, $+0.8$]} & $-0.4$ {\scriptsize[$-1.6$, $+0.8$]} \\
\rowcolor{oursbg} & Noisy-OR & 95.6 & 81.7 & 97.5 & 97.8 & $+0.4$ {\scriptsize[$-0.6$, $+1.5$]} & $+0.4$ {\scriptsize[$-0.6$, $+1.5$]} \\
\addlinespace[2pt]
8 & Joint call & 96.9 & 80.4 & 97.4 & 98.3 & --- & $+0.5$ {\scriptsize[$+0.1$, $+1.0$]} \\
 & Calibrated joint call & 96.4 & 80.4 & 97.2 & 97.5 & $-0.5$ {\scriptsize[$-1.0$, $-0.1$]} & --- \\
 & OR of yes/no & 85.3 & 96.6 & 99.9 & 71.1 & $-11.6$ {\scriptsize[$-13.8$, $-9.3$]} & $-11.2$ {\scriptsize[$-13.4$, $-8.7$]} \\
 & $k$ of $N$ & 91.2 & 57.0 & 95.1 & 91.5 & $-5.8$ {\scriptsize[$-7.3$, $-4.2$]} & $-5.3$ {\scriptsize[$-6.9$, $-3.7$]} \\
\rowcolor{oursbg} & Calibrated bit (max-P) & 96.4 & 86.0 & 96.5 & 97.5 & $-0.5$ {\scriptsize[$-1.5$, $+0.5$]} & $+0.0$ {\scriptsize[$-1.1$, $+1.1$]} \\
\rowcolor{oursbg} & Noisy-OR & 95.6 & 74.3 & 97.0 & 96.7 & $-1.3$ {\scriptsize[$-2.4$, $-0.3$]} & $-0.8$ {\scriptsize[$-2.0$, $+0.2$]} \\
\addlinespace[2pt]
16 & Joint call & 97.4 & 76.8 & 98.0 & 97.8 & --- & $+0.1$ {\scriptsize[$-0.4$, $+0.4$]} \\
 & Calibrated joint call & 97.4 & 73.9 & 97.4 & 98.3 & $-0.1$ {\scriptsize[$-0.4$, $+0.4$]} & --- \\
 & OR of yes/no & 80.5 & 98.6 & 99.9 & 61.2 & $-17.0$ {\scriptsize[$-19.2$, $-14.7$]} & $-16.9$ {\scriptsize[$-19.2$, $-14.6$]} \\
 & $k$ of $N$ & 91.3 & 46.4 & 93.9 & 90.6 & $-6.2$ {\scriptsize[$-7.7$, $-4.7$]} & $-6.1$ {\scriptsize[$-7.6$, $-4.6$]} \\
\rowcolor{oursbg} & Calibrated bit (max-P) & 96.3 & 89.9 & 96.7 & 96.1 & $-1.2$ {\scriptsize[$-2.3$, $-0.0$]} & $-1.1$ {\scriptsize[$-2.2$, $+0.0$]} \\
\rowcolor{oursbg} & Noisy-OR & 96.7 & 76.8 & 97.0 & 97.2 & $-0.8$ {\scriptsize[$-1.8$, $+0.3$]} & $-0.7$ {\scriptsize[$-1.7$, $+0.3$]} \\
\addlinespace[2pt]
32 & Joint call & 97.4 & 80.0 & 97.7 & 97.2 & --- & $-0.6$ {\scriptsize[$-1.5$, $+0.1$]} \\
 & Calibrated joint call & 98.0 & 80.0 & 97.3 & 98.9 & $+0.6$ {\scriptsize[$-0.1$, $+1.5$]} & --- \\
 & OR of yes/no & 74.0 & 96.0 & 99.9 & 48.2 & $-23.3$ {\scriptsize[$-25.7$, $-20.8$]} & $-23.9$ {\scriptsize[$-26.3$, $-21.5$]} \\
 & $k$ of $N$ & 92.4 & 48.0 & 94.8 & 90.6 & $-5.0$ {\scriptsize[$-6.6$, $-3.5$]} & $-5.6$ {\scriptsize[$-7.2$, $-4.1$]} \\
\rowcolor{oursbg} & Calibrated bit (max-P) & 96.0 & 88.0 & 94.9 & 97.2 & $-1.3$ {\scriptsize[$-2.5$, $-0.2$]} & $-1.9$ {\scriptsize[$-3.0$, $-0.9$]} \\
\rowcolor{oursbg} & Noisy-OR & 96.3 & 64.0 & 96.1 & 97.0 & $-1.0$ {\scriptsize[$-2.1$, $+0.1$]} & $-1.6$ {\scriptsize[$-2.6$, $-0.7$]} \\
\addlinespace[2pt]
\multicolumn{8}{@{}p{0.97\textwidth}@{}}{\scriptsize Pooled over $N$ (95\% / 90\% CI): noisy-OR $-$ joint $-0.68$ [$-1.43$, $+0.09$] / [$-1.32$, $-0.05$]; noisy-OR $-$ calibrated joint $-0.68$ [$-1.43$, $-0.04$] / [$-1.29$, $-0.13$]; bit $-$ joint $-0.87$ [$-1.63$, $-0.14$] / [$-1.50$, $-0.25$]; bit $-$ calibrated joint $-0.86$ [$-1.61$, $-0.14$] / [$-1.50$, $-0.25$]; $k$ of $N$ $-$ joint $-5.64$ [$-6.76$, $-4.54$] / [$-6.59$, $-4.71$]} \\
\midrule
\multicolumn{8}{l}{\emph{AV2-PV, Qwen3-VL-8B}} \\
4 & Joint call & 81.8 & 68.1 & 84.1 & 92.1 & --- & $-0.5$ {\scriptsize[$-2.2$, $+1.2$]} \\
 & Calibrated joint call & 82.2 & 79.4 & 85.2 & 83.9 & $+0.5$ {\scriptsize[$-1.2$, $+2.2$]} & --- \\
 & OR of yes/no & 82.7 & 84.5 & 93.2 & 79.2 & $+1.0$ {\scriptsize[$-1.1$, $+3.0$]} & $+0.5$ {\scriptsize[$-1.3$, $+2.3$]} \\
 & $k$ of $N$ & 82.7 & 84.5 & 93.2 & 79.2 & $+1.0$ {\scriptsize[$-1.1$, $+3.0$]} & $+0.5$ {\scriptsize[$-1.3$, $+2.3$]} \\
\rowcolor{oursbg} & Calibrated bit (max-P) & 82.6 & 76.4 & 89.8 & 86.0 & $+0.8$ {\scriptsize[$-0.7$, $+2.3$]} & $+0.3$ {\scriptsize[$-1.4$, $+2.1$]} \\
\rowcolor{oursbg} & Noisy-OR & 83.1 & 77.3 & 89.8 & 86.3 & $+1.3$ {\scriptsize[$-0.3$, $+2.9$]} & $+0.9$ {\scriptsize[$-0.7$, $+2.4$]} \\
\addlinespace[2pt]
7 & Joint call & 79.2 & 60.0 & 74.8 & 91.6 & --- & $-1.9$ {\scriptsize[$-3.2$, $-0.7$]} \\
 & Calibrated joint call & 81.2 & 69.1 & 83.8 & 86.4 & $+1.9$ {\scriptsize[$+0.7$, $+3.2$]} & --- \\
 & OR of yes/no & 80.8 & 83.1 & 93.5 & 73.7 & $+1.6$ {\scriptsize[$-0.4$, $+3.5$]} & $-0.4$ {\scriptsize[$-1.9$, $+1.2$]} \\
 & $k$ of $N$ & 80.8 & 83.1 & 93.5 & 73.7 & $+1.6$ {\scriptsize[$-0.4$, $+3.5$]} & $-0.4$ {\scriptsize[$-1.9$, $+1.2$]} \\
\rowcolor{oursbg} & Calibrated bit (max-P) & 81.9 & 73.1 & 87.1 & 84.2 & $+2.7$ {\scriptsize[$+1.1$, $+4.2$]} & $+0.7$ {\scriptsize[$-0.5$, $+2.0$]} \\
\rowcolor{oursbg} & Noisy-OR & 82.0 & 75.6 & 88.1 & 82.6 & $+2.8$ {\scriptsize[$+1.2$, $+4.3$]} & $+0.9$ {\scriptsize[$-0.3$, $+2.1$]} \\
\addlinespace[2pt]
\multicolumn{8}{@{}p{0.97\textwidth}@{}}{\scriptsize Pooled over $N$ (95\% / 90\% CI): noisy-OR $-$ joint $+2.07$ [$+0.79$, $+3.37$] / [$+0.97$, $+3.17$]; noisy-OR $-$ calibrated joint $+0.85$ [$-0.24$, $+2.04$] / [$-0.08$, $+1.83$]; bit $-$ joint $+1.74$ [$+0.48$, $+3.01$] / [$+0.68$, $+2.85$]; bit $-$ calibrated joint $+0.53$ [$-0.69$, $+1.69$] / [$-0.45$, $+1.52$]; $k$ of $N$ $-$ joint $+1.28$ [$-0.38$, $+2.96$] / [$-0.12$, $+2.70$]} \\
\bottomrule
\end{tabular}
\caption{\textbf{Existence with fixed rules} (no picking among rules on held-out data), by evidence type, with leads against the joint call and the
calibrated joint call (95\% bootstrap over rooms on Thor-PV and over driving logs on AV2-PV, 2{,}000 resamples). Thresholds fitted per $N$ by
5-fold CV by room on Thor-PV and by driving log on AV2-PV (the two sweeps of a log stay together).
On AV2-PV the fitted $k$ of $N$ is OR at both $N$.}
\label{tab:s-rules}
\end{table*}

%% file: tables/s_thor_8b.tex
\begin{table*}[t]
\centering
\footnotesize
\setlength{\tabcolsep}{4pt}
\begin{tabular}{llrrrrrl}
\toprule
$N$ & Method (rule picked on held-out rooms) & Balanced & Exclusive & Shared & Absent & Tokens & Joint call minus this \\
\midrule
$4$ & Joint call & 93.5 & 86.8 & 97.2 & 92.6 & 3.1k & --- \\
 & Calibrated joint call & 93.0 & 82.0 & 95.6 & 93.9 & --- & ${+}0.5$ {\scriptsize[$-$0.5, $+$1.4]} \\
 & Best vote rule (2+) & 89.6 & 51.6 & 95.1 & 95.3 & 3.2k & ${+}3.9$ {\scriptsize[$+$2.6, $+$5.2]} \\
\rowcolor{oursbg} & Calibrated fusion (max-P) & 94.6 & 88.1 & 97.1 & 94.5 & 3.2k & ${-}1.1$ {\scriptsize[$-$2.3, $+$0.0]} \\
 & Four-view hierarchy (OR) & 93.5 & 86.2 & 97.3 & 92.6 & 3.1k & ${+}0.0$ {\scriptsize[$-$0.1, $+$0.2]} \\
 & Route, then verify & 89.9 & 93.1 & 99.2 & 82.1 & 4.8k & ${+}3.7$ {\scriptsize[$+$1.9, $+$5.5]} \\
 & Reports + aggregator & 89.7 & 92.9 & 98.0 & 82.6 & 3.6k & ${+}3.9$ {\scriptsize[$+$2.0, $+$5.7]} \\
\midrule
$8$ & Joint call & 94.8 & 87.2 & 96.3 & 94.2 & 6.2k & --- \\
 & Calibrated joint call & 95.4 & 86.0 & 95.5 & 96.4 & --- & ${-}0.7$ {\scriptsize[$-$1.6, $+$0.1]} \\
 & Best vote rule (2+) & 90.6 & 65.9 & 98.5 & 86.2 & 6.4k & ${+}4.2$ {\scriptsize[$+$2.6, $+$5.8]} \\
\rowcolor{oursbg} & Calibrated fusion (noisy-OR) & 95.7 & 80.4 & 96.8 & 96.4 & 6.4k & ${-}0.9$ {\scriptsize[$-$2.1, $+$0.1]} \\
 & Four-view hierarchy (OR) & 92.3 & 95.0 & 97.8 & 87.1 & 6.2k & ${+}2.5$ {\scriptsize[$+$1.1, $+$4.0]} \\
 & Route, then verify & 86.1 & 97.8 & 99.0 & 73.3 & 9.6k & ${+}8.7$ {\scriptsize[$+$6.6, $+$10.9]} \\
 & Reports + aggregator & 84.7 & 97.2 & 97.0 & 72.5 & 7.1k & ${+}10.0$ {\scriptsize[$+$7.8, $+$12.2]} \\
\midrule
$16$ & Joint call & 95.6 & 85.5 & 96.4 & 95.3 & 12.4k & --- \\
 & Calibrated joint call & 96.3 & 76.8 & 94.7 & 98.6 & --- & ${-}0.6$ {\scriptsize[$-$1.6, $+$0.2]} \\
 & Best vote rule (k-of-N) & 91.1 & 52.2 & 93.6 & 90.4 & 12.8k & ${+}4.5$ {\scriptsize[$+$2.7, $+$6.2]} \\
\rowcolor{oursbg} & Calibrated fusion (noisy-OR) & 96.0 & 72.5 & 95.8 & 97.2 & 12.8k & ${-}0.4$ {\scriptsize[$-$1.5, $+$0.7]} \\
 & Four-view hierarchy (2+) & 94.7 & 75.4 & 96.4 & 93.9 & 12.5k & ${+}0.9$ {\scriptsize[$-$0.2, $+$2.0]} \\
 & Route, then verify & 83.4 & 100.0 & 99.5 & 67.2 & 19.1k & ${+}12.3$ {\scriptsize[$+$10.1, $+$14.6]} \\
 & Reports + aggregator & 80.5 & 100.0 & 98.4 & 62.5 & 14.1k & ${+}15.1$ {\scriptsize[$+$12.8, $+$17.4]} \\
\midrule
$32$ & Joint call & 95.9 & 88.0 & 97.0 & 95.0 & 24.7k & --- \\
 & Calibrated joint call & 96.6 & 84.0 & 94.4 & 98.9 & --- & ${-}0.6$ {\scriptsize[$-$1.6, $+$0.3]} \\
 & Best vote rule (k-of-N) & 91.9 & 48.0 & 91.5 & 92.8 & 25.6k & ${+}4.1$ {\scriptsize[$+$2.5, $+$5.6]} \\
\rowcolor{oursbg} & Calibrated fusion (noisy-OR) & 96.6 & 56.0 & 94.8 & 98.9 & 25.5k & ${-}0.6$ {\scriptsize[$-$1.8, $+$0.4]} \\
 & Four-view hierarchy (major) & 95.5 & 56.0 & 92.3 & 99.2 & 25.0k & ${+}0.4$ {\scriptsize[$-$0.8, $+$1.7]} \\
 & Route, then verify & 78.1 & 96.0 & 99.3 & 57.0 & 38.0k & ${+}17.8$ {\scriptsize[$+$15.5, $+$20.3]} \\
 & Reports + aggregator & 73.5 & 100.0 & 95.7 & 51.2 & 28.0k & ${+}22.5$ {\scriptsize[$+$20.0, $+$25.0]} \\
\bottomrule
\end{tabular}
\caption{\textbf{Existence on Thor-PV, Qwen3-VL-8B, by evidence type.} 260 rooms; exclusive: one of the $N$ cameras sees the kind;
shared: several do; absent: the room has none. Tokens: prompt and completion per question, summed over every call the
method makes. CIs: 95\% paired room bootstrap.}
\label{tab:s-thor-8b}
\end{table*}

%% file: tables/s_thor_4b.tex
\begin{table*}[t]
\centering
\footnotesize
\setlength{\tabcolsep}{4pt}
\begin{tabular}{llrrrrrl}
\toprule
$N$ & Method (rule picked on held-out rooms) & Balanced & Exclusive & Shared & Absent & Tokens & Joint call minus this \\
\midrule
$4$ & Joint call & 95.3 & 80.2 & 98.2 & 97.0 & 3.1k & --- \\
 & Calibrated joint call & 95.2 & 80.7 & 98.2 & 96.7 & --- & ${+}0.1$ {\scriptsize[$-$0.1, $+$0.4]} \\
 & Best vote rule (2+) & 91.1 & 58.2 & 96.3 & 95.6 & 3.2k & ${+}4.2$ {\scriptsize[$+$2.9, $+$5.5]} \\
\rowcolor{oursbg} & Calibrated fusion (noisy-OR) & 95.6 & 81.7 & 97.5 & 97.8 & 3.2k & ${-}0.4$ {\scriptsize[$-$1.5, $+$0.6]} \\
 & Four-view hierarchy (OR) & 95.2 & 79.9 & 98.2 & 97.0 & 3.1k & ${+}0.0$ {\scriptsize[$+$0.0, $+$0.1]} \\
 & Route, then verify & 91.6 & 93.9 & 99.5 & 85.1 & 4.9k & ${+}3.7$ {\scriptsize[$+$2.0, $+$5.4]} \\
 & Reports + aggregator & 87.6 & 92.9 & 94.8 & 81.0 & 3.6k & ${+}7.6$ {\scriptsize[$+$5.4, $+$9.6]} \\
\midrule
$8$ & Joint call & 96.9 & 80.4 & 97.4 & 98.3 & 6.2k & --- \\
 & Calibrated joint call & 96.4 & 80.4 & 97.2 & 97.5 & --- & ${+}0.5$ {\scriptsize[$+$0.1, $+$1.0]} \\
 & Best vote rule (2+) & 91.2 & 72.1 & 98.9 & 86.5 & 6.4k & ${+}5.7$ {\scriptsize[$+$3.9, $+$7.6]} \\
\rowcolor{oursbg} & Calibrated fusion (max-P) & 96.4 & 86.0 & 96.5 & 97.5 & 6.4k & ${+}0.5$ {\scriptsize[$-$0.5, $+$1.5]} \\
 & Four-view hierarchy (OR) & 95.9 & 85.5 & 98.3 & 95.0 & 6.2k & ${+}1.0$ {\scriptsize[$-$0.0, $+$2.1]} \\
 & Route, then verify & 88.7 & 95.0 & 99.2 & 78.8 & 9.7k & ${+}8.2$ {\scriptsize[$+$6.2, $+$10.1]} \\
 & Reports + aggregator & 84.5 & 91.6 & 92.5 & 76.6 & 7.2k & ${+}12.4$ {\scriptsize[$+$10.2, $+$14.5]} \\
\midrule
$16$ & Joint call & 97.4 & 76.8 & 98.0 & 97.8 & 12.4k & --- \\
 & Calibrated joint call & 97.4 & 73.9 & 97.4 & 98.3 & --- & ${+}0.1$ {\scriptsize[$-$0.4, $+$0.4]} \\
 & Best vote rule (k-of-N) & 91.3 & 46.4 & 93.9 & 90.6 & 12.8k & ${+}6.2$ {\scriptsize[$+$4.7, $+$7.7]} \\
\rowcolor{oursbg} & Calibrated fusion (noisy-OR) & 96.7 & 76.8 & 97.0 & 97.2 & 12.8k & ${+}0.8$ {\scriptsize[$-$0.3, $+$1.8]} \\
 & Four-view hierarchy (2+) & 96.3 & 71.0 & 95.6 & 98.1 & 12.5k & ${+}1.1$ {\scriptsize[$+$0.1, $+$2.1]} \\
 & Route, then verify & 86.8 & 97.1 & 99.5 & 74.1 & 19.3k & ${+}10.7$ {\scriptsize[$+$8.5, $+$12.8]} \\
 & Reports + aggregator & 81.4 & 95.7 & 95.2 & 67.5 & 14.3k & ${+}16.1$ {\scriptsize[$+$13.8, $+$18.4]} \\
\midrule
$32$ & Joint call & 97.4 & 80.0 & 97.7 & 97.2 & 24.7k & --- \\
 & Calibrated joint call & 98.0 & 80.0 & 97.3 & 98.9 & --- & ${-}0.6$ {\scriptsize[$-$1.5, $+$0.1]} \\
 & Best vote rule (k-of-N) & 92.4 & 48.0 & 94.8 & 90.6 & 25.6k & ${+}5.0$ {\scriptsize[$+$3.5, $+$6.6]} \\
\rowcolor{oursbg} & Calibrated fusion (LO-logistic) & 96.5 & 56.0 & 96.1 & 97.5 & 25.5k & ${+}0.9$ {\scriptsize[$-$0.3, $+$2.0]} \\
 & Four-view hierarchy (k-of-N) & 96.9 & 64.0 & 96.4 & 97.8 & 25.0k & ${+}0.5$ {\scriptsize[$-$0.7, $+$1.7]} \\
 & Route, then verify & 82.9 & 96.0 & 99.2 & 66.7 & 38.4k & ${+}14.4$ {\scriptsize[$+$12.1, $+$16.8]} \\
 & Reports + aggregator & 75.9 & 96.0 & 97.0 & 54.8 & 28.5k & ${+}21.5$ {\scriptsize[$+$19.1, $+$24.0]} \\
\bottomrule
\end{tabular}
\caption{\textbf{Existence on Thor-PV, Qwen3-VL-4B, by evidence type.} 260 rooms; exclusive: one of the $N$ cameras sees the kind;
shared: several do; absent: the room has none. Tokens: prompt and completion per question, summed over every call the
method makes. CIs: 95\% paired room bootstrap.}
\label{tab:s-thor-4b}
\end{table*}

%% file: tables/s_count.tex
\begin{table*}[t]
\centering
\scriptsize
\setlength{\tabcolsep}{3pt}
\begin{tabular}{@{}lrlrrrrrrrll@{}}
\toprule
Model & $N$ & Kinds & Joint & Max & Learned & Gated max & Gated learned & Stacked & Gated joint & Joint $-$ gated max & Gated joint $-$ stacked \\
\midrule
Qwen3-VL-8B & 4 & all & 57.3 & 53.5 & 54.6 & 60.7 & 58.7 & 59.5 & 62.9 & $-3.5$ {\scriptsize[$-5.3$, $-1.4$]} & $+3.4$ {\scriptsize[$+1.0$, $+5.6$]} \\
 &  & present & 59.5 & 57.8 & 48.0 & 52.1 & 49.2 & 51.0 & 54.9 & $+7.4$ {\scriptsize[$+5.2$, $+9.6$]} & $+3.9$ {\scriptsize[$+0.7$, $+6.9$]} \\
 &  & zero-visible & 51.9 & 43.2 & 70.5 & 81.3 & 81.3 & 80.0 & 82.0 & $-29.5$ {\scriptsize[$-32.8$, $-26.0$]} & $+2.0$ {\scriptsize[$-0.8$, $+4.7$]} \\
\addlinespace[1pt]
 & 8 & all & 56.5 & 50.8 & 53.5 & 58.6 & 59.0 & 58.8 & 59.7 & $-2.0$ {\scriptsize[$-4.1$, $+0.1$]} & $+0.8$ {\scriptsize[$-1.3$, $+3.0$]} \\
 &  & present & 56.1 & 54.3 & 49.2 & 49.8 & 50.4 & 51.9 & 52.3 & $+6.3$ {\scriptsize[$+4.1$, $+8.4$]} & $+0.4$ {\scriptsize[$-2.2$, $+3.1$]} \\
 &  & zero-visible & 58.0 & 38.1 & 68.6 & 89.3 & 89.3 & 83.4 & 85.6 & $-31.4$ {\scriptsize[$-35.2$, $-27.9$]} & $+2.2$ {\scriptsize[$-0.9$, $+5.3$]} \\
\addlinespace[1pt]
 & 16 & all & 58.4 & 49.9 & 53.2 & 59.0 & 57.9 & 59.8 & 60.9 & $-0.7$ {\scriptsize[$-2.7$, $+1.3$]} & $+1.2$ {\scriptsize[$-1.3$, $+3.6$]} \\
 &  & present & 57.5 & 54.3 & 49.0 & 50.7 & 49.3 & 53.3 & 53.2 & $+6.8$ {\scriptsize[$+4.8$, $+8.8$]} & $-0.1$ {\scriptsize[$-3.1$, $+2.7$]} \\
 &  & zero-visible & 61.9 & 31.3 & 70.7 & 94.2 & 94.2 & 87.0 & 93.5 & $-32.3$ {\scriptsize[$-36.5$, $-27.8$]} & $+6.5$ {\scriptsize[$+3.7$, $+9.4$]} \\
\addlinespace[1pt]
 & 32 & all & 58.0 & 48.9 & 54.0 & 59.4 & 59.3 & 62.7 & 61.2 & $-1.4$ {\scriptsize[$-3.4$, $+0.4$]} & $-1.5$ {\scriptsize[$-4.3$, $+1.1$]} \\
 &  & present & 56.3 & 54.1 & 50.0 & 50.8 & 50.7 & 56.1 & 53.1 & $+5.4$ {\scriptsize[$+3.3$, $+7.4$]} & $-3.0$ {\scriptsize[$-6.3$, $+0.2$]} \\
 &  & zero-visible & 65.8 & 25.1 & 72.2 & 98.7 & 98.7 & 93.3 & 98.4 & $-32.9$ {\scriptsize[$-37.4$, $-28.4$]} & $+5.1$ {\scriptsize[$+2.8$, $+7.6$]} \\
\addlinespace[1pt]
\midrule
Qwen3-VL-4B & 4 & all & 63.8 & 58.1 & 57.7 & 63.4 & 61.0 & 61.3 & 64.6 & $+0.4$ {\scriptsize[$-1.5$, $+2.1$]} & $+3.3$ {\scriptsize[$+0.9$, $+5.7$]} \\
 &  & present & 62.7 & 59.5 & 52.3 & 56.0 & 52.6 & 52.6 & 57.0 & $+6.7$ {\scriptsize[$+4.3$, $+8.7$]} & $+4.4$ {\scriptsize[$+1.4$, $+7.6$]} \\
 &  & zero-visible & 66.4 & 54.7 & 70.5 & 81.2 & 81.2 & 82.3 & 83.0 & $-14.7$ {\scriptsize[$-17.4$, $-11.8$]} & $+0.7$ {\scriptsize[$-1.8$, $+3.1$]} \\
\addlinespace[1pt]
 & 8 & all & 63.7 & 55.0 & 54.0 & 60.3 & 58.5 & 61.1 & 65.1 & $+3.4$ {\scriptsize[$+1.4$, $+5.4$]} & $+4.0$ {\scriptsize[$+1.5$, $+6.5$]} \\
 &  & present & 62.1 & 56.3 & 49.7 & 52.3 & 49.9 & 54.5 & 58.7 & $+9.9$ {\scriptsize[$+7.5$, $+12.2$]} & $+4.2$ {\scriptsize[$+1.2$, $+7.3$]} \\
 &  & zero-visible & 69.5 & 50.3 & 69.3 & 88.9 & 88.9 & 84.3 & 87.6 & $-19.4$ {\scriptsize[$-22.6$, $-16.1$]} & $+3.3$ {\scriptsize[$+0.7$, $+5.9$]} \\
\addlinespace[1pt]
 & 16 & all & 64.7 & 54.9 & 55.4 & 61.3 & 59.1 & 63.5 & 66.8 & $+3.4$ {\scriptsize[$+1.1$, $+5.4$]} & $+3.3$ {\scriptsize[$+0.8$, $+5.9$]} \\
 &  & present & 62.5 & 56.9 & 51.5 & 53.7 & 51.0 & 57.0 & 60.3 & $+8.7$ {\scriptsize[$+6.2$, $+11.2$]} & $+3.2$ {\scriptsize[$+0.1$, $+6.3$]} \\
 &  & zero-visible & 73.9 & 46.1 & 71.9 & 93.2 & 93.2 & 90.7 & 94.5 & $-19.3$ {\scriptsize[$-22.9$, $-15.6$]} & $+3.8$ {\scriptsize[$+1.6$, $+6.1$]} \\
\addlinespace[1pt]
 & 32 & all & 63.7 & 52.4 & 57.6 & 61.2 & 61.6 & 64.2 & 66.4 & $+2.5$ {\scriptsize[$+0.3$, $+4.6$]} & $+2.3$ {\scriptsize[$-0.3$, $+4.9$]} \\
 &  & present & 60.7 & 55.8 & 53.1 & 53.3 & 53.9 & 57.8 & 59.4 & $+7.4$ {\scriptsize[$+4.9$, $+9.8$]} & $+1.6$ {\scriptsize[$-1.5$, $+4.7$]} \\
 &  & zero-visible & 77.4 & 36.7 & 78.4 & 97.3 & 97.3 & 93.5 & 98.9 & $-19.9$ {\scriptsize[$-23.7$, $-16.3$]} & $+5.4$ {\scriptsize[$+3.3$, $+7.7$]} \\
\addlinespace[1pt]
\bottomrule
\end{tabular}
\caption{\textbf{Counting on Thor-PV by kind type} (exact accuracy, \%). Present: kinds with an instance in view of at least one
of the $N$ cameras. Zero-visible: questions with no instance in any of the $N$ views, that is, the 363 kinds the room does not have plus
the kinds it has but none of the $N$ cameras sees (248, 96, 36 and 8 questions at $N=4$, 8, 16 and 32; 611, 459, 399 and 371 in all).
Max: max of the agents' counts. Learned: gradient-boosted classifier on the sorted counts, sum, max, views reporting any and the
kind. Gated max: 0 when the noisy-OR of the agents' P(yes) is below its threshold, else the max of counts (at least 1). Gated
learned: the same gate, then the classifier trained on present kinds. Stacked: one classifier on the counts plus the sorted view
log-odds and the noisy-OR, with no explicit gate. Gated joint: the joint call's count, 0 when its $p_J$ is below its threshold,
else at least 1. Every gate threshold is fitted on all training questions with ``an instance in view'' as the positive class.
All fitted on training rooms (5-fold CV by room); CIs: 95\% paired room bootstrap.}
\label{tab:s-count}
\end{table*}

%% file: tables/s_lone.tex
\begin{table*}[t]
\centering
\scriptsize
\setlength{\tabcolsep}{3.4pt}
\renewcommand{\arraystretch}{0.9}
\begin{tabular}{llrrrlll}
\toprule
$N$ {\scriptsize(present)} & Rule (fixed) & Balanced & Lone recall & Absent & joint $-$ rule & calibrated joint $-$ rule & joint recall at the rule's FPR \\
\midrule
\multicolumn{8}{l}{\emph{Thor-PV lone-witness sets, Qwen3-VL-8B}} \\
4 {\scriptsize(1590)} & Joint call & 89.3 & 87.2 & 91.5 & --- & --- & --- \\
 & Calibrated joint call & 89.2 & 87.2 & 91.2 & --- & --- & --- \\
 & OR of yes/no & 86.3 & 94.6 & 78.0 & --- & --- & --- \\
\rowcolor{oursbg} & Calibrated bit (max-P) & 89.3 & 88.3 & 90.4 & $+0.0$ {\scriptsize[$-1.6$, $+1.5$]} & $-0.1$ {\scriptsize[$-1.6$, $+1.3$]} & 88.9 at 9.6 \\
\rowcolor{oursbg} & Noisy-OR & 89.1 & 86.4 & 91.7 & $+0.3$ {\scriptsize[$-1.3$, $+1.7$]} & $+0.1$ {\scriptsize[$-1.3$, $+1.4$]} & 87.1 at 8.3 \\
\addlinespace[2pt]
8 {\scriptsize(1444)} & Joint call & 89.1 & 84.5 & 93.7 & --- & --- & --- \\
 & Calibrated joint call & 89.5 & 87.5 & 91.5 & --- & --- & --- \\
 & OR of yes/no & 82.7 & 96.5 & 68.9 & --- & --- & --- \\
\rowcolor{oursbg} & Calibrated bit (max-P) & 89.5 & 86.9 & 92.0 & $-0.4$ {\scriptsize[$-1.8$, $+0.9$]} & $+0.0$ {\scriptsize[$-1.4$, $+1.4$]} & 87.6 at 8.0 \\
\rowcolor{oursbg} & Noisy-OR & 85.9 & 87.0 & 84.8 & $+3.2$ {\scriptsize[$+1.5$, $+5.0$]} & $+3.6$ {\scriptsize[$+1.8$, $+5.4$]} & 91.6 at 15.2 \\
\addlinespace[2pt]
16 {\scriptsize(1019)} & Joint call & 88.0 & 83.2 & 92.8 & --- & --- & --- \\
 & Calibrated joint call & 88.0 & 87.5 & 88.4 & --- & --- & --- \\
 & OR of yes/no & 79.1 & 97.6 & 60.6 & --- & --- & --- \\
\rowcolor{oursbg} & Calibrated bit (max-P) & 87.4 & 84.4 & 90.4 & $+0.7$ {\scriptsize[$-0.8$, $+2.1$]} & $+0.6$ {\scriptsize[$-1.0$, $+2.3$]} & 86.0 at 9.6 \\
\rowcolor{oursbg} & Noisy-OR & 85.3 & 86.0 & 84.6 & $+2.8$ {\scriptsize[$+1.1$, $+4.6$]} & $+2.7$ {\scriptsize[$+1.1$, $+4.6$]} & 90.3 at 15.4 \\
\addlinespace[2pt]
32 {\scriptsize(25)} & Joint call & 90.8 & 88.0 & 93.7 & --- & --- & --- \\
 & Calibrated joint call & 86.9 & 84.0 & 89.8 & --- & --- & --- \\
 & OR of yes/no & 73.3 & 100.0 & 46.6 & --- & --- & --- \\
\rowcolor{oursbg} & Calibrated bit (max-P) & 86.4 & 84.0 & 88.7 & $+4.5$ {\scriptsize[$+1.3$, $+9.3$]} & $+0.6$ {\scriptsize[$-5.2$, $+6.4$]} & 88.0 at 11.3 \\
\rowcolor{oursbg} & Noisy-OR & 83.1 & 88.0 & 78.2 & $+7.7$ {\scriptsize[$+1.9$, $+13.3$]} & $+3.8$ {\scriptsize[$-3.5$, $+10.7$]} & 100.0 at 21.8 \\
\addlinespace[2pt]
\midrule
\multicolumn{8}{l}{\emph{Thor-PV lone-witness sets, Qwen3-VL-4B}} \\
4 {\scriptsize(1590)} & Joint call & 89.1 & 82.1 & 96.1 & --- & --- & --- \\
 & Calibrated joint call & 90.1 & 91.2 & 89.0 & --- & --- & --- \\
 & OR of yes/no & 88.2 & 96.0 & 80.4 & --- & --- & --- \\
\rowcolor{oursbg} & Calibrated bit (max-P) & 91.4 & 91.3 & 91.5 & $-2.3$ {\scriptsize[$-3.9$, $-0.6$]} & $-1.3$ {\scriptsize[$-3.1$, $+0.4$]} & 89.1 at 8.5 \\
\rowcolor{oursbg} & Noisy-OR & 91.6 & 87.0 & 96.1 & $-2.5$ {\scriptsize[$-3.6$, $-1.3$]} & $-1.5$ {\scriptsize[$-3.2$, $+0.2$]} & 81.9 at 3.9 \\
\addlinespace[2pt]
8 {\scriptsize(1444)} & Joint call & 86.8 & 76.5 & 97.2 & --- & --- & --- \\
 & Calibrated joint call & 88.3 & 86.7 & 89.8 & --- & --- & --- \\
 & OR of yes/no & 83.1 & 97.2 & 68.9 & --- & --- & --- \\
\rowcolor{oursbg} & Calibrated bit (max-P) & 90.8 & 88.4 & 93.1 & $-3.9$ {\scriptsize[$-5.6$, $-2.3$]} & $-2.5$ {\scriptsize[$-4.2$, $-1.0$]} & 82.9 at 6.9 \\
\rowcolor{oursbg} & Noisy-OR & 88.8 & 88.6 & 89.0 & $-1.9$ {\scriptsize[$-3.6$, $-0.3$]} & $-0.5$ {\scriptsize[$-2.0$, $+1.0$]} & 87.7 at 11.0 \\
\addlinespace[2pt]
16 {\scriptsize(1019)} & Joint call & 87.8 & 78.1 & 97.5 & --- & --- & --- \\
 & Calibrated joint call & 89.9 & 85.3 & 94.5 & --- & --- & --- \\
 & OR of yes/no & 79.7 & 98.0 & 61.4 & --- & --- & --- \\
\rowcolor{oursbg} & Calibrated bit (max-P) & 89.4 & 87.5 & 91.2 & $-1.5$ {\scriptsize[$-3.3$, $+0.2$]} & $+0.5$ {\scriptsize[$-1.2$, $+2.2$]} & 88.0 at 8.8 \\
\rowcolor{oursbg} & Noisy-OR & 87.4 & 86.1 & 88.7 & $+0.4$ {\scriptsize[$-1.3$, $+2.3$]} & $+2.5$ {\scriptsize[$+0.9$, $+4.0$]} & 89.1 at 11.3 \\
\addlinespace[2pt]
32 {\scriptsize(25)} & Joint call & 86.6 & 76.0 & 97.2 & --- & --- & --- \\
 & Calibrated joint call & 87.3 & 88.0 & 86.5 & --- & --- & --- \\
 & OR of yes/no & 72.1 & 96.0 & 48.2 & --- & --- & --- \\
\rowcolor{oursbg} & Calibrated bit (max-P) & 92.9 & 88.0 & 97.8 & $-6.3$ {\scriptsize[$-15.0$, $+2.5$]} & $-5.6$ {\scriptsize[$-12.8$, $+2.7$]} & 76.0 at 2.2 \\
\rowcolor{oursbg} & Noisy-OR & 80.6 & 76.0 & 85.1 & $+6.1$ {\scriptsize[$+0.4$, $+12.1$]} & $+6.7$ {\scriptsize[$+0.7$, $+13.2$]} & 92.0 at 14.9 \\
\addlinespace[2pt]
\bottomrule
\end{tabular}
\caption{\textbf{The lone witness, paired.} Sets with exactly one of the $N$ cameras seeing the kind (and $N$ random
cameras for 363 absent questions), rebuilt from the 32 cameras with stored probabilities; the joint call is run on the
same sets. Thresholds fitted per $N$ by 5-fold CV by room; 95\% bootstrap over rooms, 1{,}000 resamples. Last column: the recall of the
joint call's $p_J$ when thresholded at the rule's false-alarm rate (\%). At $N=32$ only 25 questions are present.}
\label{tab:s-lone}
\end{table*}

%% file: tables/s_scanprob.tex
\begin{table*}[tp]
\centering
\scriptsize
\setlength{\tabcolsep}{3.6pt}
\renewcommand{\arraystretch}{0.9}
\begin{tabular}{rlrrrll}
\toprule
$N$ & Rule (fixed) & Balanced & Present & Absent & rule $-$ joint call & rule $-$ calibrated joint call \\
\midrule
\multicolumn{7}{l}{\emph{ScanNet, Qwen3-VL-8B}} \\
4 & Joint call & 92.6 & 93.9 & 91.4 & --- & --- \\
 & Calibrated joint call & 92.6 & 92.4 & 92.7 & --- & --- \\
 & OR of yes/no & 89.6 & 96.2 & 83.1 & --- & --- \\
 & $k$ of $N$ & 89.6 & 96.2 & 83.1 & --- & --- \\
\rowcolor{oursbg} & Calibrated bit (max-P) & 92.7 & 92.6 & 92.7 & $+0.0$ {\scriptsize[$-0.8$, $+0.8$]} & $+0.1$ {\scriptsize[$-0.8$, $+1.1$]} \\
\rowcolor{oursbg} & Noisy-OR & 92.7 & 92.6 & 92.9 & $+0.1$ {\scriptsize[$-0.8$, $+1.0$]} & $+0.2$ {\scriptsize[$-0.8$, $+1.2$]} \\
 & \multicolumn{6}{l}{\scriptsize AUC: noisy-OR 97.5, $\max_i p_i$ 97.5, joint call's $p_J$ 97.4} \\
\addlinespace[2pt]
8 & Joint call & 92.8 & 95.3 & 90.3 & --- & --- \\
 & Calibrated joint call & 93.1 & 94.0 & 92.2 & --- & --- \\
 & OR of yes/no & 87.3 & 98.0 & 76.7 & --- & --- \\
 & $k$ of $N$ & 88.8 & 87.6 & 90.0 & --- & --- \\
\rowcolor{oursbg} & Calibrated bit (max-P) & 93.1 & 93.3 & 92.9 & $+0.3$ {\scriptsize[$-0.5$, $+1.1$]} & $-0.0$ {\scriptsize[$-0.7$, $+0.7$]} \\
\rowcolor{oursbg} & Noisy-OR & 92.8 & 92.7 & 92.8 & $-0.0$ {\scriptsize[$-1.0$, $+0.9$]} & $-0.4$ {\scriptsize[$-1.2$, $+0.5$]} \\
 & \multicolumn{6}{l}{\scriptsize AUC: noisy-OR 97.7, $\max_i p_i$ 97.7, joint call's $p_J$ 97.6} \\
\addlinespace[2pt]
16 & Joint call & 93.1 & 97.0 & 89.2 & --- & --- \\
 & Calibrated joint call & 93.0 & 94.1 & 91.9 & --- & --- \\
 & OR of yes/no & 84.5 & 99.6 & 69.3 & --- & --- \\
 & $k$ of $N$ & 89.7 & 90.9 & 88.4 & --- & --- \\
\rowcolor{oursbg} & Calibrated bit (max-P) & 93.3 & 95.7 & 90.8 & $+0.2$ {\scriptsize[$-0.7$, $+1.0$]} & $+0.3$ {\scriptsize[$-0.7$, $+1.2$]} \\
\rowcolor{oursbg} & Noisy-OR & 92.9 & 94.1 & 91.7 & $-0.2$ {\scriptsize[$-1.3$, $+0.9$]} & $-0.1$ {\scriptsize[$-1.2$, $+1.1$]} \\
 & \multicolumn{6}{l}{\scriptsize AUC: noisy-OR 98.1, $\max_i p_i$ 98.2, joint call's $p_J$ 98.0} \\
\addlinespace[2pt]
\multicolumn{7}{l}{\scriptsize Pooled over $N$: noisy-OR $-$ calibrated joint $-0.10$ [$-0.77$, $+0.64$]; noisy-OR $-$ joint $-0.04$ [$-0.74$, $+0.65$]; bit $-$ calibrated joint $+0.11$ [$-0.50$, $+0.68$]} \\
\midrule
\multicolumn{7}{l}{\emph{ScanNet, LLaVA-OneVision-7B}} \\
4 & Joint call & 91.9 & 92.2 & 91.6 & --- & --- \\
 & Calibrated joint call & 92.0 & 93.7 & 90.3 & --- & --- \\
 & OR of yes/no & 89.8 & 94.9 & 84.7 & --- & --- \\
 & $k$ of $N$ & 89.8 & 94.9 & 84.7 & --- & --- \\
\rowcolor{oursbg} & Calibrated bit (max-P) & 91.2 & 90.3 & 92.1 & $-0.7$ {\scriptsize[$-2.1$, $+0.5$]} & $-0.8$ {\scriptsize[$-2.2$, $+0.4$]} \\
\rowcolor{oursbg} & Noisy-OR & 91.4 & 91.6 & 91.2 & $-0.5$ {\scriptsize[$-1.6$, $+0.6$]} & $-0.6$ {\scriptsize[$-1.7$, $+0.4$]} \\
 & \multicolumn{6}{l}{\scriptsize AUC: noisy-OR 96.7, $\max_i p_i$ 97.1, joint call's $p_J$ 97.2} \\
\addlinespace[2pt]
8 & Joint call & 92.1 & 94.0 & 90.1 & --- & --- \\
 & Calibrated joint call & 92.2 & 94.4 & 90.1 & --- & --- \\
 & OR of yes/no & 88.3 & 97.4 & 79.1 & --- & --- \\
 & $k$ of $N$ & 87.6 & 88.7 & 86.5 & --- & --- \\
\rowcolor{oursbg} & Calibrated bit (max-P) & 92.1 & 93.1 & 91.1 & $+0.0$ {\scriptsize[$-0.9$, $+1.1$]} & $-0.2$ {\scriptsize[$-1.0$, $+0.8$]} \\
\rowcolor{oursbg} & Noisy-OR & 91.4 & 92.5 & 90.3 & $-0.6$ {\scriptsize[$-1.7$, $+0.3$]} & $-0.8$ {\scriptsize[$-1.7$, $+0.1$]} \\
 & \multicolumn{6}{l}{\scriptsize AUC: noisy-OR 96.8, $\max_i p_i$ 97.5, joint call's $p_J$ 97.5} \\
\addlinespace[2pt]
16 & Joint call & 92.3 & 95.2 & 89.4 & --- & --- \\
 & Calibrated joint call & 92.8 & 93.4 & 92.2 & --- & --- \\
 & OR of yes/no & 85.3 & 98.8 & 71.9 & --- & --- \\
 & $k$ of $N$ & 90.3 & 90.4 & 90.3 & --- & --- \\
\rowcolor{oursbg} & Calibrated bit (max-P) & 92.4 & 90.8 & 94.1 & $+0.1$ {\scriptsize[$-0.9$, $+1.2$]} & $-0.4$ {\scriptsize[$-1.2$, $+0.5$]} \\
\rowcolor{oursbg} & Noisy-OR & 90.9 & 92.4 & 89.4 & $-1.4$ {\scriptsize[$-2.6$, $-0.3$]} & $-2.0$ {\scriptsize[$-3.1$, $-0.8$]} \\
 & \multicolumn{6}{l}{\scriptsize AUC: noisy-OR 97.0, $\max_i p_i$ 97.9, joint call's $p_J$ 97.8} \\
\addlinespace[2pt]
\multicolumn{7}{l}{\scriptsize Pooled over $N$: noisy-OR $-$ calibrated joint $-1.13$ [$-1.90$, $-0.42$]; noisy-OR $-$ joint $-0.86$ [$-1.62$, $-0.08$]; bit $-$ calibrated joint $-0.44$ [$-1.30$, $+0.30$]} \\
\bottomrule
\end{tabular}
\caption{\textbf{Probability fusion on real scans, two model families} (150 ScanNet scans, existence; per-view and joint
probabilities re-collected with the prompts of the answer runs). Thresholds fitted per $N$ by 5-fold CV by scan; 95\%
bootstrap over scans (1{,}000 resamples per $N$, 2{,}000 pooled). Present: accuracy on present kinds that a frame sees.}
\label{tab:s-scanprob}
\end{table*}

%% file: tables/s_auc.tex
\begin{table*}[tp]
\centering
\footnotesize
\setlength{\tabcolsep}{4pt}
\begin{tabular}{@{}lrllllll@{}}
\toprule
 & & \multicolumn{2}{c}{Present and seen vs absent} & \multicolumn{2}{c}{Exclusive vs absent} & \multicolumn{2}{c}{Shared vs absent} \\
\cmidrule(lr){3-4}\cmidrule(lr){5-6}\cmidrule(l){7-8}
Testbed, model & $N$ & joint / team & joint $-$ team & joint / team & joint $-$ team & joint / team & joint $-$ team \\
\midrule
Thor-PV, 8B & 4 & 98.4 / 98.5 & $-0.1$ {\scriptsize[$-0.4$, $+0.2$]} & 96.1 / 96.0 & $+0.1$ {\scriptsize[$-0.8$, $+0.7$]} & 99.2 / 99.4 & $-0.2$ {\scriptsize[$-0.5$, $+0.0$]} \\
Thor-PV, 8B & 8 & 99.0 / 99.0 & $-0.0$ {\scriptsize[$-0.2$, $+0.1$]} & 97.7 / 97.3 & $+0.4$ {\scriptsize[$-0.4$, $+1.5$]} & 99.2 / 99.4 & $-0.2$ {\scriptsize[$-0.3$, $-0.0$]} \\
Thor-PV, 8B & 16 & 99.3 / 99.2 & $+0.1$ {\scriptsize[$-0.0$, $+0.3$]} & 97.6 / 97.6 & $+0.0$ {\scriptsize[$-1.1$, $+1.0$]} & 99.4 / 99.3 & $+0.1$ {\scriptsize[$-0.1$, $+0.2$]} \\
Thor-PV, 8B & 32 & 99.4 / 99.2 & $+0.2$ {\scriptsize[$+0.1$, $+0.4$]} & 97.4 / 95.9 & $+1.6$ {\scriptsize[$-0.5$, $+3.9$]} & 99.4 / 99.2 & $+0.2$ {\scriptsize[$+0.0$, $+0.3$]} \\
\midrule
Thor-PV, 4B & 4 & 98.9 / 99.0 & $-0.0$ {\scriptsize[$-0.3$, $+0.2$]} & 96.7 / 97.0 & $-0.3$ {\scriptsize[$-1.0$, $+0.3$]} & 99.7 / 99.7 & $+0.0$ {\scriptsize[$-0.1$, $+0.2$]} \\
Thor-PV, 4B & 8 & 99.3 / 99.3 & $-0.0$ {\scriptsize[$-0.1$, $+0.1$]} & 96.4 / 97.0 & $-0.6$ {\scriptsize[$-1.4$, $+0.3$]} & 99.7 / 99.7 & $-0.0$ {\scriptsize[$-0.1$, $+0.1$]} \\
Thor-PV, 4B & 16 & 99.6 / 99.5 & $+0.1$ {\scriptsize[$-0.0$, $+0.2$]} & 96.4 / 97.3 & $-0.9$ {\scriptsize[$-2.9$, $+0.7$]} & 99.7 / 99.6 & $+0.1$ {\scriptsize[$+0.0$, $+0.2$]} \\
Thor-PV, 4B & 32 & 99.6 / 99.5 & $+0.2$ {\scriptsize[$+0.0$, $+0.3$]} & 94.2 / 95.8 & $-1.6$ {\scriptsize[$-5.7$, $+2.8$]} & 99.7 / 99.5 & $+0.2$ {\scriptsize[$+0.1$, $+0.2$]} \\
\midrule
AV2-PV, 8B & 4 & 91.2 / 91.2 & $+0.0$ {\scriptsize[$-0.7$, $+0.8$]} & 90.4 / 90.1 & $+0.3$ {\scriptsize[$-0.6$, $+1.2$]} & 94.4 / 95.5 & $-1.2$ {\scriptsize[$-2.4$, $-0.1$]} \\
AV2-PV, 8B & 7 & 89.3 / 90.0 & $-0.7$ {\scriptsize[$-1.4$, $+0.1$]} & 86.4 / 87.3 & $-0.9$ {\scriptsize[$-1.9$, $+0.1$]} & 92.7 / 93.3 & $-0.7$ {\scriptsize[$-1.5$, $+0.2$]} \\
\bottomrule
\end{tabular}
\caption{\textbf{Threshold-free separation (AUC, \%).} Joint: the joint call's $p_J$. Team: the better of $\max_i p_i$ and the
noisy-OR of the agents' $p_i$ (picked on the evaluation data, which favours the team). CIs: 95\% bootstrap over rooms
(Thor-PV) or driving logs (AV2-PV), 300 resamples. At $N=32$ only 25 Thor-PV questions are exclusive.}
\label{tab:s-auc}
\end{table*}

%% file: tables/s_law.tex
\begin{table}[t]
\centering
\scriptsize
\setlength{\tabcolsep}{2.5pt}
\begin{tabular}{@{}lrrrrr@{}}
\toprule
 & & \multicolumn{3}{c}{OR team, absent kinds} & Joint call \\
\cmidrule(lr){3-5}
Testbed, model & $N$ & observed & law & indep. & absent \\
\midrule
Thor-PV, Qwen3-VL-8B & 4 & 79.9 & 80.2 & 78.4 & 92.6 \\
Thor-PV, Qwen3-VL-8B & 8 & 69.1 & 69.0 & 64.4 & 94.2 \\
Thor-PV, Qwen3-VL-8B & 16 & 60.6 & 56.1 & 48.3 & 95.3 \\
Thor-PV, Qwen3-VL-8B & 32 & 46.6 & 44.1 & 34.0 & 95.0 \\
\midrule
Thor-PV, Qwen3-VL-4B & 4 & 81.5 & 82.0 & 79.7 & 97.0 \\
Thor-PV, Qwen3-VL-4B & 8 & 71.1 & 72.2 & 67.2 & 98.3 \\
Thor-PV, Qwen3-VL-4B & 16 & 61.2 & 60.6 & 52.9 & 97.8 \\
Thor-PV, Qwen3-VL-4B & 32 & 48.2 & 47.9 & 39.1 & 97.2 \\
\midrule
AV2-PV, Qwen3-VL-8B & 4 & 79.2 & 79.3 & 78.6 & 92.1 \\
AV2-PV, Qwen3-VL-8B & 7 & 73.7 & 70.6 & 69.0 & 91.6 \\
\midrule
ScanNet, Qwen3-VL-8B & 4 & 83.1 & 83.1 & 79.9 & 91.4 \\
ScanNet, Qwen3-VL-8B & 8 & 76.7 & 74.3 & 65.8 & 90.3 \\
ScanNet, Qwen3-VL-8B & 16 & 69.3 & 64.0 & 48.1 & 89.2 \\
\midrule
ScanNet, Qwen3-VL-4B & 4 & 83.1 & 83.1 & 79.3 & 90.8 \\
ScanNet, Qwen3-VL-4B & 8 & 75.4 & 75.1 & 65.6 & 90.4 \\
ScanNet, Qwen3-VL-4B & 16 & 67.5 & 66.0 & 48.5 & 90.2 \\
\midrule
ScanNet, LLaVA-OV-7B & 4 & 84.7 & 84.7 & 81.8 & 91.6 \\
ScanNet, LLaVA-OV-7B & 8 & 79.1 & 76.5 & 69.4 & 90.1 \\
ScanNet, LLaVA-OV-7B & 16 & 71.9 & 66.0 & 53.4 & 89.4 \\
\midrule
ScanNet, Qwen2.5-VL-7B & 4 & 87.5 & 87.5 & 86.3 & 95.3 \\
ScanNet, Qwen2.5-VL-7B & 8 & 81.3 & 79.3 & 75.8 & 95.5 \\
ScanNet, Qwen2.5-VL-7B & 16 & 75.2 & 68.5 & 60.5 & 96.7 \\
\midrule
\multicolumn{6}{@{}l}{\emph{Earlier renders with numbered tags drawn on the objects (not used in the main text)}} \\
Thor-PV-tags, Qwen3-VL-8B & 4 & 70.0 & 70.2 & 68.7 & 87.6 \\
Thor-PV-tags, Qwen3-VL-8B & 8 & 58.1 & 55.6 & 52.0 & 89.0 \\
Thor-PV-tags, Qwen3-VL-8B & 16 & 46.3 & 40.8 & 34.6 & 91.7 \\
Thor-PV-tags, Qwen3-VL-8B & 32 & 34.2 & 28.7 & 21.3 & 91.5 \\
Thor-PV-tags, Qwen3-VL-4B & 4 & 76.9 & 77.0 & 74.8 & 92.6 \\
Thor-PV-tags, Qwen3-VL-4B & 8 & 64.7 & 65.3 & 60.6 & 93.1 \\
Thor-PV-tags, Qwen3-VL-4B & 16 & 53.4 & 51.6 & 44.8 & 96.7 \\
Thor-PV-tags, Qwen3-VL-4B & 32 & 41.0 & 37.5 & 29.8 & 95.0 \\
Thor-PV-tags, Idefics3-8B & 4 & 76.6 & 76.4 & 72.0 & 93.1 \\
Thor-PV-tags, Idefics3-8B & 8 & 63.4 & 65.8 & 56.5 & 94.5 \\
Thor-PV-tags, Mistral-Small-3.2 & 4 & 25.9 & 25.8 & 21.7 & 12.7 \\
Thor-PV-tags, Mistral-Small-3.2 & 8 & 18.5 & 17.1 & 12.8 & 21.2 \\
Thor-PV-tags, Mistral-Small-3.2 & 16 & 11.8 & 11.6 & 7.9 & 27.8 \\
Thor-PV-tags, GPT-5.4-mini & 4 & 25.2 & 26.3 & 21.3 & 27.5 \\
Thor-PV-tags, GPT-5.4-mini & 8 & 10.7 & 16.6 & 11.1 & 30.5 \\
Thor-PV-tags, GPT-5.4-mini & 16 & 4.6 & 10.8 & 7.3 & 31.3 \\
\bottomrule
\end{tabular}
\caption{\textbf{The model of Eq.~(1) for every run.} Fitted per kind on the flags of the first four cameras (frames) of the
training rooms, scans or scenes and predicted on held-out ones (5-fold CV by room or scan; by driving log on AV2-PV).
$N=4$ is the camera count the fit uses, but its predictions are still for held-out rooms: those rows check the fit on new rooms
(maximum likelihood on the training rooms need not match the held-out zero-flag rate), while the rows at $N>4$ are forecasts to
larger camera counts. GPT-5.4-mini (100 rooms) flags 45\% of absent kinds per view, so its OR team is near the floor.}
\label{tab:s-law}
\end{table}

%% file: tables/s_new.tex

\begin{table*}[t]\centering\small
\caption{Answer-passing protocols and text readers in the rendered rooms, balanced accuracy (\%) at $N=4/8/16/32$. $\dagger$: one threshold per $N$ fitted with labels on the training folds. Debate was run at $N=4$ and 16 only.}
\label{tab:s-protocols}
\begin{tabular}{l cccc cccc}
\toprule
 & \multicolumn{4}{c}{Qwen3-VL-8B} & \multicolumn{4}{c}{Qwen3-VL-4B} \\
Rule & 4 & 8 & 16 & 32 & 4 & 8 & 16 & 32 \\
\midrule
OR of yes/no answers & 89.1 & 84.3 & 80.2 & 73.3 & 90.1 & 85.3 & 80.5 & 74.0 \\
Conservative prompt, OR & 92.9 & 88.8 & 85.3 & 80.3 & 92.7 & 88.7 & 85.0 & 78.4 \\
Debate, two rounds, majority & 88.0 & -- & 83.3 & -- & 84.1 & -- & 80.3 & -- \\
Verbal confidence, noisy-OR$^\dagger$ & 90.4 & 88.8 & 86.9 & 83.9 & 94.5 & 93.7 & 94.5 & 93.5 \\
Verbal confidence, max$^\dagger$ & 90.6 & 88.9 & 86.6 & 82.9 & 94.7 & 93.7 & 91.8 & 87.5 \\
Conservative prompt, vote$^\dagger$ & 92.9 & 93.0 & 93.3 & 94.1 & 92.7 & 93.4 & 93.6 & 94.6 \\
\midrule
Reader of answers & 89.1 & 84.3 & 80.5 & 73.4 & 78.0 & 70.4 & 80.5 & 74.0 \\
Reader of probabilities & 82.5 & 74.7 & 68.4 & 63.9 & 82.0 & 75.5 & 72.1 & 65.6 \\
\quad told how to read them & 88.6 & 84.3 & 78.2 & 73.4 & 81.8 & 85.3 & 82.0 & 74.9 \\
\quad eight labelled examples & 82.8 & 73.1 & 74.7 & 68.8 & 89.6 & 79.1 & 79.7 & 76.8 \\
\quad brief reasoning & 85.6 & 88.1 & 85.7 & 79.7 & 85.7 & 88.3 & 87.2 & 86.2 \\
Reader of answers + probabilities$^\dagger$ & 88.8 & 88.1 & 87.6 & 84.0 & 90.5 & 85.2 & 80.1 & 73.8 \\
Reader of log-odds$^\dagger$ & 87.7 & 85.3 & 82.6 & 77.9 & 87.3 & 88.7 & 86.4 & 82.7 \\
\midrule
\rowcolor{oursbg}Calibrated bit (max rule)$^\dagger$ & 94.6 & 95.7 & 94.9 & 94.8 & 94.8 & 96.4 & 96.3 & 96.0 \\
\rowcolor{oursbg}Noisy-OR$^\dagger$ & 94.6 & 95.7 & 96.0 & 96.6 & 95.6 & 95.6 & 96.7 & 96.3 \\
Joint call (greedy) & 93.5 & 94.8 & 95.6 & 95.9 & 95.3 & 96.9 & 97.4 & 97.4 \\
\bottomrule
\end{tabular}
\end{table*}

\begin{table}[t]\centering\scriptsize\setlength{\tabcolsep}{3.0pt}
\caption{The max rule under fixed cuts, one cut optimised for all $N$ on the training folds, and the per-$N$ fit (rooms). Last column: common minus per-$N$, pooled over $N$, thresholds refitted in every bootstrap resample. The last two rows of each block use exact score breakpoints, so the per-$N$ row can differ from the other tables (a 200-quantile grid) by up to 0.4.}
\label{tab:s-cuts}
\begin{tabular}{ll cccc l}
\toprule
Model & Cut & 4 & 8 & 16 & 32 & pooled \\
\midrule
8B & fixed 0.5 & 88.7 & 83.6 & 79.4 & 72.6 & \\
8B & fixed 0.9 & 94.1 & 90.8 & 85.7 & 80.8 & \\
8B & fixed 0.99 & 94.5 & 94.8 & 92.5 & 87.6 & \\
8B & fixed 0.999 & 92.2 & 95.2 & 95.2 & 92.6 & \\
8B & fixed 0.9999 & 88.4 & 93.1 & 94.8 & 94.9 & \\
8B & one optimised cut & 92.2 & 95.3 & 95.2 & 92.2 & $-1.3$ $[-1.8, -0.5]$ \\
8B & fitted per $N$ & 94.6 & 95.7 & 95.1 & 94.8 & \\
\midrule
4B & fixed 0.5 & 89.6 & 84.6 & 79.8 & 73.9 & \\
4B & fixed 0.9 & 93.3 & 90.1 & 85.7 & 78.1 & \\
4B & fixed 0.99 & 95.0 & 94.0 & 90.3 & 85.4 & \\
4B & fixed 0.999 & 95.6 & 96.3 & 93.9 & 90.5 & \\
4B & fixed 0.9999 & 94.4 & 96.4 & 95.9 & 93.3 & \\
4B & one optimised cut & 94.3 & 96.6 & 96.4 & 93.8 & $-0.6$ $[-1.7, -0.1]$ \\
4B & fitted per $N$ & 94.8 & 96.6 & 96.2 & 95.7 & \\
\midrule
InternVL3 & one optimised cut & 81.0 & 83.8 & 85.1 & 84.7 & $-0.8$ $[-1.7, +0.4]$ \\
InternVL3 & fitted per $N$ & 83.0 & 84.5 & 85.3 & 85.3 & \\
Idefics3 & one optimised cut & 79.5 & 84.3 & 85.4 & 83.1 & $-1.3$ $[-3.0, -0.3]$ \\
Idefics3 & fitted per $N$ & 83.5 & 84.8 & 84.4 & 84.9 & \\
\bottomrule
\end{tabular}
\end{table}

\begin{table*}[t]\centering\footnotesize\setlength{\tabcolsep}{3.0pt}
\caption{The max rule with its threshold fitted to the joint call's answers instead of labels (no ground truth on the calibration side): all training scenes, and twenty scenes (mean over 40 draws, 5th to 95th percentile).}
\label{tab:s-labelfree}
\begin{tabular}{l l cccc}
\toprule
Setting & Threshold fitted to & 4 & 8 & 16 & 32 \\
\midrule
Rooms, 8B & labels & 94.6 & 95.7 & 94.9 & 94.8 \\
 & joint call, all scenes & 94.4 & 95.3 & 94.8 & 94.6 \\
 & joint call, twenty scenes & 94.5 [94.0, 94.8] & 94.9 [94.3, 95.4] & 94.7 [94.2, 95.2] & 94.1 [93.4, 94.7] \\
\midrule
Rooms, 4B & labels & 94.8 & 96.4 & 96.3 & 96.0 \\
 & joint call, all scenes & 95.3 & 96.0 & 96.3 & 96.1 \\
 & joint call, twenty scenes & 94.8 [94.0, 95.3] & 95.5 [94.5, 96.3] & 95.7 [95.0, 96.3] & 95.7 [94.9, 96.2] \\
\midrule
Rooms, InternVL3 & labels & 82.9 & 84.2 & 85.5 & 85.2 \\
 & joint call, all scenes & 82.2 & 84.3 & 85.7 & 85.4 \\
 & joint call, twenty scenes & 81.6 [80.1, 82.5] & 83.6 [82.2, 84.7] & 85.1 [84.2, 86.0] & 85.2 [84.5, 85.8] \\
\midrule
Rooms, Idefics3 & labels & 83.2 & 84.6 & 84.3 & 84.5 \\
 & joint call, all scenes & 81.1 & 84.1 & 82.5 & 81.9 \\
 & joint call, twenty scenes & 81.2 [79.9, 82.1] & 82.5 [80.6, 84.1] & 81.8 [79.9, 83.6] & 81.4 [78.4, 83.6] \\
\midrule
ScanNet, 8B & labels & 92.7 & 93.1 & 93.3 & -- \\
 & joint call, all scenes & 92.0 & 92.3 & 92.9 & -- \\
 & joint call, twenty scenes & 92.1 [91.8, 92.5] & 91.9 [91.2, 92.6] & 92.9 [92.3, 93.3] & -- \\
\midrule
ScanNet, LLaVA-OV & labels & 91.2 & 92.1 & 92.4 & -- \\
 & joint call, all scenes & 91.7 & 91.8 & 92.0 & -- \\
 & joint call, twenty scenes & 91.3 [90.8, 91.8] & 91.7 [91.0, 92.3] & 91.6 [91.0, 92.3] & -- \\
\midrule
Rooms, 8B & 16-image budget joint call (itself 94.1), twenty scenes & -- & -- & -- & 93.9 [93.3, 94.4] \\
Rooms, 8B & 8-image budget joint call (itself 90.4), twenty scenes & -- & -- & -- & 93.3 [92.4, 94.0] \\
Rooms, 8B & 4-image budget joint call (itself 82.7), twenty scenes & -- & -- & -- & 91.5 [89.1, 93.3] \\
\bottomrule
\end{tabular}
\end{table*}

\begin{table*}[t]\centering\small
\caption{Lone-witness sets: exactly one of $N$ cameras sees the object (363 absent questions). Balanced accuracy with thresholds refitted on these sets and with the thresholds of the ordinary questions (deployment), and recall of lone witnesses with thresholds matched to the joint call's false-alarm rate on the training folds (joint minus rule, paired bootstrap over rooms).}
\label{tab:s-needle}
\begin{tabular}{l r c cc c ccc}
\toprule
 & & & \multicolumn{2}{c}{bit} & noisy-OR & \multicolumn{3}{c}{recall at the joint call's false-alarm rate} \\
$N$ & present & joint & refit & deploy & deploy & joint & joint $-$ bit & joint $-$ noisy-OR \\
\midrule
\multicolumn{9}{l}{\emph{Qwen3-VL-8B}} \\
4 & 1,632 & 88.9 & 88.6 & 88.7 & 88.7 & 87.4 & $+0.4$ $[-0.8, +1.5]$ & $+0.3$ $[-0.8, +1.4]$ \\
8 & 1,554 & 88.7 & 87.4 & 88.4 & 85.5 & 84.7 & $+1.4$ $[+0.1, +2.6]$ & $+3.6$ $[+2.1, +5.1]$ \\
16 & 1,293 & 88.1 & 87.3 & 86.4 & 79.0 & 83.4 & $+5.6$ $[+3.9, +7.2]$ & $+11.1$ $[+9.2, +13.4]$ \\
32 & 671 & 87.9 & 87.0 & 87.2 & 79.2 & 82.1 & $+0.0$ $[-2.0, +1.8]$ & $+10.3$ $[+7.7, +13.1]$ \\
\multicolumn{9}{l}{\emph{Qwen3-VL-4B}} \\
4 & 1,632 & 90.0 & 90.4 & 90.2 & 89.2 & 84.4 & $-1.0$ $[-2.6, +0.6]$ & $+0.6$ $[-1.0, +2.1]$ \\
8 & 1,554 & 86.6 & 90.9 & 88.8 & 84.1 & 76.3 & $-2.7$ $[-4.7, -0.8]$ & $+2.6$ $[+0.6, +4.7]$ \\
16 & 1,293 & 87.8 & 87.9 & 88.2 & 78.8 & 79.1 & $+2.8$ $[+0.9, +4.7]$ & $+13.6$ $[+11.4, +15.8]$ \\
32 & 671 & 88.1 & 86.5 & 87.0 & 81.9 & 78.4 & $+4.2$ $[+1.5, +6.9]$ & $+13.0$ $[+9.9, +15.9]$ \\
\bottomrule
\end{tabular}
\end{table*}

\begin{table*}[t]\centering\small
\caption{Other model families and a larger model in the rooms (same questions, cameras and folds). Per-view rates: share of views answered yes when the kind is absent / when the view shows it.}
\label{tab:s-families}
\begin{tabular}{ll cccc}
\toprule
Model & Rule & 4 & 8 & 16 & 32 \\
\midrule
InternVL3-8B & OR & 74.1 & 69.7 & 62.3 & 58.9 \\
InternVL3-8B & fitted vote$^\dagger$ & 77.9 & 78.4 & 78.9 & 80.0 \\
\rowcolor{oursbg}InternVL3-8B & noisy-OR$^\dagger$ & 83.5 & 82.9 & 84.2 & 82.6 \\
\rowcolor{oursbg}InternVL3-8B & bit$^\dagger$ & 82.9 & 84.2 & 85.5 & 85.2 \\
InternVL3-8B & joint call & 81.8 & 83.2 & 84.4 & 84.7 \\
InternVL3-8B & calibrated joint call$^\dagger$ & 84.3 & 85.3 & 86.6 & 87.7 \\
\multicolumn{6}{l}{per-view yes rate: 19.0\% absent, 79.7\% shown; bit minus greedy joint call, pooled: $+0.9$ $[-0.4, +2.4]$} \\
\midrule
Idefics3-8B & OR & 81.8 & 79.3 & 73.5 & 68.1 \\
Idefics3-8B & fitted vote$^\dagger$ & 81.8 & 82.6 & 83.0 & 82.0 \\
\rowcolor{oursbg}Idefics3-8B & noisy-OR$^\dagger$ & 84.0 & 85.5 & 85.7 & 86.5 \\
\rowcolor{oursbg}Idefics3-8B & bit$^\dagger$ & 83.2 & 84.6 & 84.3 & 84.5 \\
Idefics3-8B & joint call & 79.8 & 81.3 & 81.1 & 81.2 \\
Idefics3-8B & calibrated joint call$^\dagger$ & 80.5 & 81.0 & 83.0 & 83.2 \\
\multicolumn{6}{l}{per-view yes rate: 9.7\% absent, 72.7\% shown; bit minus greedy joint call, pooled: $+3.3$ $[+2.0, +4.4]$} \\
\midrule
Qwen3-VL-32B & OR & 89.4 & 84.6 & 79.1 & 74.7 \\
Qwen3-VL-32B & fitted vote$^\dagger$ & 88.6 & 91.0 & 90.8 & 91.0 \\
\rowcolor{oursbg}Qwen3-VL-32B & noisy-OR$^\dagger$ & 94.0 & 94.6 & 94.6 & 95.0 \\
\rowcolor{oursbg}Qwen3-VL-32B & bit$^\dagger$ & 93.1 & 95.5 & 95.7 & 95.2 \\
Qwen3-VL-32B & joint call & 93.9 & 95.5 & 96.7 & 96.8 \\
Qwen3-VL-32B & calibrated joint call$^\dagger$ & 93.2 & 95.7 & 96.9 & 97.5 \\
\multicolumn{6}{l}{per-view yes rate: 7.5\% absent, 92.4\% shown; bit minus greedy joint call, pooled: $-0.9$ $[-1.8, +0.1]$} \\
\bottomrule
\end{tabular}
\end{table*}

\begin{table}[t]\centering\small
\caption{OR team's accuracy on absent kinds (\%) for the three further room models: observed, the beta-binomial forecast from four cameras (Eq.~1 of the main text) and the independent-views forecast, held-out rooms.}
\label{tab:s-law9}
\begin{tabular}{ll ccc}
\toprule
Model & & 8 & 16 & 32 \\
\midrule
Qwen3-VL-32B & observed & 70.0 & 58.7 & 49.6 \\
Qwen3-VL-32B & beta-binomial & 70.7 & 59.6 & 47.7 \\
Qwen3-VL-32B & independent & 66.0 & 51.4 & 37.3 \\
\midrule
InternVL3-8B & observed & 42.7 & 26.4 & 18.5 \\
InternVL3-8B & beta-binomial & 40.6 & 27.9 & 19.5 \\
InternVL3-8B & independent & 31.6 & 16.0 & 7.4 \\
\midrule
Idefics3-8B & observed & 62.8 & 48.5 & 36.9 \\
Idefics3-8B & beta-binomial & 61.7 & 49.3 & 38.4 \\
Idefics3-8B & independent & 56.2 & 40.6 & 27.6 \\
\bottomrule
\end{tabular}
\end{table}

%% file: tables/s_prior.tex
\begin{table*}[t]\centering\footnotesize\setlength{\tabcolsep}{3.5pt}
\caption{Fusion rules of prior multi-agent and classifier-combination work applied to the agents of the main text (Thor-PV, balanced accuracy in \%, $N=4/8/16/32$). Last columns: rule minus \method, pooled over $N$, 95\% room-bootstrap interval. $\dagger$: threshold or rates fitted on the training folds. Debate was run at $N=4$ and 16.}
\label{tab:s-prior}
\begin{tabular}{l cccc cccc ll}
\toprule
 & \multicolumn{4}{c}{Qwen3-VL-8B} & \multicolumn{4}{c}{Qwen3-VL-4B} & \multicolumn{2}{c}{minus \method, pooled} \\
Rule & 4 & 8 & 16 & 32 & 4 & 8 & 16 & 32 & 8B & 4B \\
\midrule
Majority vote~\cite{du2024improving} & 78.7 & 80.3 & 81.0 & 80.4 & 79.3 & 82.1 & 82.8 & 83.1 & $-15.6$ $[-16.7, -14.4]$ & $-14.2$ $[-15.4, -13.0]$ \\
Optimal fusion of hard decisions~\cite{chair1986optimal}$^\dagger$ & 89.6 & 90.7 & 91.5 & 92.2 & 91.1 & 91.5 & 91.8 & 92.3 & $-4.7$ $[-5.6, -3.8]$ & $-4.4$ $[-5.3, -3.6]$ \\
$k$ of $N$ vote$^\dagger$ & 88.4 & 89.8 & 91.1 & 91.9 & 89.6 & 91.2 & 91.3 & 92.4 & $-5.4$ $[-6.4, -4.4]$ & $-5.0$ $[-6.0, -4.0]$ \\
Debate, two rounds, majority~\cite{du2024improving} & 88.0 & -- & 83.3 & -- & 84.1 & -- & 80.3 & -- & $-9.6$ $[-11.2, -8.1]$ & $-14.0$ $[-15.7, -12.3]$ \\
Confidence-weighted vote, verbal~\cite{chen2024reconcile} & 73.8 & 71.6 & 70.0 & 68.1 & 61.5 & 57.9 & 56.1 & 55.2 & $-24.8$ $[-26.1, -23.6]$ & $-38.4$ $[-39.3, -37.3]$ \\
Confidence-weighted vote, first-token $p_i$ & 81.1 & 82.5 & 82.7 & 81.9 & 81.5 & 83.9 & 84.0 & 84.3 & $-13.6$ $[-14.8, -12.5]$ & $-12.6$ $[-13.8, -11.5]$ \\
\midrule
Sum rule, cut 0.5~\cite{kittler1998combining} & 85.6 & 84.2 & 83.8 & 82.1 & 87.0 & 86.0 & 85.1 & 84.3 & $-11.8$ $[-12.8, -10.7]$ & $-10.4$ $[-11.4, -9.4]$ \\
Product rule, cut 0.5 & 86.2 & 84.5 & 84.3 & 83.0 & 87.8 & 86.8 & 86.5 & 85.4 & $-11.2$ $[-12.3, -10.1]$ & $-9.4$ $[-10.4, -8.3]$ \\
Median rule, cut 0.5 & 85.3 & 83.7 & 83.6 & 81.9 & 87.1 & 85.9 & 85.0 & 84.3 & $-12.1$ $[-13.1, -11.0]$ & $-10.5$ $[-11.5, -9.5]$ \\
Max rule, cut 0.5 & 88.7 & 83.6 & 79.4 & 72.6 & 89.6 & 84.6 & 79.8 & 73.9 & $-14.7$ $[-16.6, -12.8]$ & $-14.1$ $[-16.0, -12.4]$ \\
Min rule, cut 0.5 & 64.9 & 56.8 & 53.1 & 51.6 & 66.0 & 57.8 & 53.4 & 51.7 & $-39.1$ $[-40.0, -38.1]$ & $-38.8$ $[-39.8, -37.8]$ \\
Sum rule$^\dagger$ & 93.2 & 92.0 & 91.4 & 92.2 & 93.4 & 92.5 & 91.3 & 91.1 & $-3.5$ $[-4.4, -2.6]$ & $-4.0$ $[-5.3, -3.0]$ \\
Product rule$^\dagger$ & 89.2 & 89.3 & 89.8 & 89.0 & 90.5 & 89.0 & 88.9 & 89.0 & $-6.4$ $[-7.7, -5.2]$ & $-6.7$ $[-7.6, -5.8]$ \\
Median rule$^\dagger$ & 89.1 & 86.9 & 85.9 & 84.7 & 89.5 & 87.5 & 86.8 & 85.5 & $-9.1$ $[-10.3, -8.0]$ & $-8.7$ $[-9.7, -7.6]$ \\
\midrule
\rowcolor{oursbg}Calibrated bit (max rule$^\dagger$) & 94.6 & 95.7 & 94.9 & 94.8 & 94.8 & 96.4 & 96.3 & 96.0 & $-0.7$ $[-1.3, -0.2]$ & $-0.2$ $[-0.9, +0.4]$ \\
\rowcolor{oursbg}\method (noisy-OR$^\dagger$) & 94.6 & 95.7 & 96.0 & 96.6 & 95.6 & 95.6 & 96.7 & 96.3 &  &  \\
\bottomrule
\end{tabular}
\end{table*}

%% file: tables/s_prior_real.tex
\begin{table*}[t]\centering\scriptsize\setlength{\tabcolsep}{2.6pt}
\caption{The fusion rules of Table~\ref{tab:s-prior} on the real testbeds, from the stored outputs (balanced accuracy in \%; ScanNet at $N=4/8/16$ frames, AV2-PV at $N=4/7$ cameras). $\Delta$: rule minus \method, pooled over $N$, with a 95\% bootstrap interval over scans or driving logs. $\dagger$: threshold or rates fitted on the training folds.}
\label{tab:s-prior-real}
\begin{tabular}{l ccc l ccc l cc l}
\toprule
 & \multicolumn{4}{c}{ScanNet, Qwen3-VL-8B} & \multicolumn{4}{c}{ScanNet, LLaVA-OV-7B} & \multicolumn{3}{c}{AV2-PV, Qwen3-VL-8B} \\
Rule & 4 & 8 & 16 & $\Delta$ & 4 & 8 & 16 & $\Delta$ & 4 & 7 & $\Delta$ \\
\midrule
Majority vote & 69.7 & 68.9 & 69.5 & $-23.4$ $[-25.7, -21.2]$ & 67.7 & 65.7 & 67.3 & $-24.3$ $[-26.4, -22.2]$ & 57.8 & 59.4 & $-24.0$ $[-26.2, -21.7]$ \\
Optimal fusion of hard decisions$^\dagger$ & 84.1 & 88.8 & 88.6 & $-5.6$ $[-6.8, -4.5]$ & 89.8 & 88.5 & 87.3 & $-2.7$ $[-3.5, -1.8]$ & 82.7 & 74.0 & $-4.2$ $[-5.4, -3.0]$ \\
$k$ of $N$ vote$^\dagger$ & 89.6 & 88.8 & 89.7 & $-3.4$ $[-4.5, -2.5]$ & 89.8 & 87.6 & 90.3 & $-2.0$ $[-2.7, -1.2]$ & 82.7 & 80.8 & $-0.8$ $[-1.9, +0.4]$ \\
Confidence-weighted vote, first-token $p_i$ & 72.7 & 70.5 & 70.3 & $-21.6$ $[-23.8, -19.3]$ & 73.8 & 70.5 & 69.7 & $-19.9$ $[-22.0, -17.8]$ & 59.4 & 59.3 & $-23.2$ $[-25.6, -20.9]$ \\
Sum rule$^\dagger$ & 92.2 & 89.1 & 89.9 & $-2.4$ $[-3.3, -1.6]$ & 90.7 & 88.8 & 89.0 & $-1.7$ $[-2.3, -1.0]$ & 82.5 & 81.3 & $-0.7$ $[-1.3, +0.1]$ \\
Product rule$^\dagger$ & 87.8 & 87.4 & 86.8 & $-5.4$ $[-6.7, -4.1]$ & 89.1 & 86.5 & 87.3 & $-3.6$ $[-4.7, -2.6]$ & 77.4 & 75.1 & $-6.3$ $[-8.4, -4.3]$ \\
Median rule$^\dagger$ & 83.8 & 80.3 & 79.7 & $-11.5$ $[-13.2, -9.8]$ & 83.6 & 79.7 & 79.3 & $-10.4$ $[-12.0, -8.7]$ & 72.7 & 70.1 & $-11.2$ $[-13.7, -8.6]$ \\
\midrule
OR of yes/no & 89.6 & 87.3 & 84.5 & $-5.7$ $[-6.9, -4.5]$ & 89.8 & 88.3 & 85.3 & $-3.4$ $[-4.4, -2.4]$ & 82.7 & 80.8 & $-0.8$ $[-1.9, +0.4]$ \\
\rowcolor{oursbg}Calibrated bit$^\dagger$ & 92.7 & 93.1 & 93.3 & $+0.2$ $[-0.4, +0.8]$ & 91.2 & 92.1 & 92.4 & $+0.7$ $[+0.0, +1.3]$ & 82.6 & 81.9 & $-0.3$ $[-1.0, +0.3]$ \\
\rowcolor{oursbg}\method$^\dagger$ & 92.7 & 92.8 & 92.9 &  & 91.4 & 91.4 & 90.9 &  & 83.1 & 82.0 &  \\
\bottomrule
\end{tabular}
\end{table*}

%% file: tables/s_pope.tex
\begin{table*}[t]\centering\scriptsize\setlength{\tabcolsep}{4.2pt}
\caption{\textbf{3D-POPE~\cite{yang20253dgrand}, a public benchmark of object existence in ScanNet rooms.} Accuracy and F1 (\%) under its three ways of drawing absent objects, for the ScanNet200 classes (the benchmark's main table) and accuracy for the NYU40 classes. The published rows are copied from~\cite{yang20253dgrand} and answer from the 3D scene; our rows answer its first question wording from evenly spaced frames of the public 25k-frame subset, for the scenes that have that many frames, so the two groups are not strictly comparable. $\dagger$: threshold fitted per team size on training scenes (five folds by scene). The sampled row is the best of four rules on five sampled answers per agent. Best frame-based value per block in bold.}
\label{tab:s-pope}
\begin{tabular}{l cc cc cc ccc}
\toprule
 & \multicolumn{6}{c}{ScanNet200 classes} & \multicolumn{3}{c}{NYU40 classes (accuracy)} \\
\cmidrule(lr){2-7}\cmidrule(l){8-10}
 & \multicolumn{2}{c}{Random} & \multicolumn{2}{c}{Popular} & \multicolumn{2}{c}{Adversarial} & Random & Popular & Adversarial \\
Method & Acc. & F1 & Acc. & F1 & Acc. & F1 & & & \\
\midrule
\multicolumn{10}{l}{\emph{Published, input: the 3D scene (all 140 scenes)}} \\
3D-LLM & 50.1 & 66.7 & 49.9 & 66.6 & 49.9 & 66.6 & 50.0 & 50.0 & 50.0 \\
3D-VisTA & 49.7 & 51.8 & 49.5 & 49.5 & 51.1 & 51.2 & 50.1 & 50.3 & 50.4 \\
LEO & 52.9 & 62.3 & 47.3 & 59.6 & 47.5 & 59.8 & 54.0 & 48.9 & 49.8 \\
3D-GRAND, zero-shot & 89.1 & 88.6 & 76.6 & 78.3 & 74.0 & 76.4 & 85.7 & 81.7 & 82.1 \\
\midrule
\multicolumn{10}{l}{\emph{Qwen3-VL-8B, input: eight frames (127 scenes)}} \\
Joint call & 86.1 & 86.5 & 74.0 & 77.5 & 70.5 & 75.1 & 82.8 & \textbf{80.0} & 78.9 \\
OR of yes/no & 83.4 & 85.2 & 70.1 & 76.1 & 67.1 & 74.3 & 78.8 & 74.8 & 74.1 \\
$k$ of $N$ say yes$^\dagger$ & 83.4 & 85.2 & 70.1 & 76.1 & 67.1 & 74.3 & 78.8 & 74.8 & 74.1 \\
Five sampled answers per agent$^\dagger$ & 84.4 & 85.8 & 71.6 & 76.8 & 67.8 & 74.5 & 80.0 & 76.2 & 75.0 \\
\rowcolor{oursbg}Calibrated bit$^\dagger$ & 86.9 & 86.9 & 74.5 & 77.3 & 70.9 & 74.9 & \textbf{83.2} & \textbf{80.0} & \textbf{79.3} \\
\rowcolor{oursbg}\method$^\dagger$ & \textbf{87.2} & \textbf{87.3} & \textbf{74.8} & \textbf{77.7} & \textbf{71.4} & \textbf{75.5} & 82.7 & 79.4 & 78.9 \\
\midrule
\multicolumn{10}{l}{\emph{Qwen3-VL-8B, input: sixteen frames (61 scenes)}} \\
Joint call & 87.3 & 88.0 & 72.4 & 77.2 & 69.6 & 75.5 & 80.8 & 79.2 & 78.2 \\
OR of yes/no & 79.2 & 82.5 & 66.4 & 74.5 & 63.6 & 72.9 & 74.3 & 71.4 & 71.3 \\
$k$ of $N$ say yes$^\dagger$ & 81.7 & 82.6 & 67.5 & 72.8 & 64.4 & 71.0 & 75.4 & 74.6 & 75.0 \\
Five sampled answers per agent$^\dagger$ & 82.4 & 83.1 & 68.3 & 73.2 & 65.5 & 71.5 & 76.7 & 75.6 & 75.8 \\
\rowcolor{oursbg}Calibrated bit$^\dagger$ & \textbf{88.3} & 88.1 & \textbf{75.5} & \textbf{78.0} & \textbf{72.8} & \textbf{76.2} & \textbf{84.1} & \textbf{81.5} & \textbf{81.0} \\
\rowcolor{oursbg}\method$^\dagger$ & 87.8 & \textbf{88.4} & 72.2 & 77.0 & 68.6 & 74.8 & 81.4 & 79.1 & 78.6 \\
\midrule
\multicolumn{10}{l}{\emph{Qwen3-VL-4B, input: eight frames (127 scenes)}} \\
Joint call & 86.7 & \textbf{86.9} & 74.3 & 77.5 & 70.2 & 74.8 & 81.9 & 79.7 & 78.9 \\
OR of yes/no & 84.6 & 86.1 & 70.8 & 76.6 & 67.4 & 74.5 & 77.9 & 74.5 & 73.5 \\
$k$ of $N$ say yes$^\dagger$ & 84.0 & 85.2 & 70.8 & 75.9 & 67.4 & 73.8 & 77.7 & 74.5 & 73.9 \\
Five sampled answers per agent$^\dagger$ & 84.9 & 86.2 & 71.5 & 76.8 & 67.8 & 74.6 & 78.7 & 74.8 & 74.0 \\
\rowcolor{oursbg}Calibrated bit$^\dagger$ & 86.9 & 86.3 & \textbf{77.2} & \textbf{78.5} & \textbf{74.0} & \textbf{76.1} & 82.9 & \textbf{80.2} & \textbf{80.1} \\
\rowcolor{oursbg}\method$^\dagger$ & \textbf{87.1} & 86.8 & 76.6 & 78.5 & 73.0 & 76.0 & \textbf{83.1} & 79.4 & 78.6 \\
\midrule
\multicolumn{10}{l}{\emph{Qwen3-VL-4B, input: sixteen frames (61 scenes)}} \\
Joint call & 87.2 & 87.8 & 73.2 & 77.4 & 69.4 & 75.0 & 80.6 & 79.2 & 78.7 \\
OR of yes/no & 81.4 & 84.1 & 66.2 & 74.4 & 63.1 & 72.7 & 73.0 & 70.1 & 70.0 \\
$k$ of $N$ say yes$^\dagger$ & 84.6 & 85.4 & 70.6 & 75.4 & 67.7 & 73.6 & 77.1 & 75.9 & 75.2 \\
Five sampled answers per agent$^\dagger$ & 84.9 & 85.4 & 71.4 & 75.6 & 67.9 & 73.3 & 77.5 & 76.5 & 75.6 \\
\rowcolor{oursbg}Calibrated bit$^\dagger$ & \textbf{88.3} & \textbf{88.2} & \textbf{76.0} & \textbf{78.5} & \textbf{73.2} & \textbf{76.6} & \textbf{81.0} & \textbf{79.7} & \textbf{79.7} \\
\rowcolor{oursbg}\method$^\dagger$ & 87.4 & 87.6 & 73.4 & 77.0 & 70.7 & 75.2 & 80.2 & 78.6 & 77.8 \\
\bottomrule
\end{tabular}
\end{table*}

%% file: tables/s_pope_n.tex
\begin{table}[t]\centering\scriptsize\setlength{\tabcolsep}{3.0pt}
\caption{\textbf{3D-POPE by team size} (the 61 scenes with sixteen frames; 1{,}880 distinct questions of both class sets, 496 present). Balanced accuracy and accuracy on absent objects (\%). $\dagger$: threshold fitted per $N$ on training scenes.}
\label{tab:s-pope-n}
\begin{tabular}{l ccc ccc}
\toprule
 & \multicolumn{3}{c}{Balanced accuracy} & \multicolumn{3}{c}{Absent objects} \\
\cmidrule(lr){2-4}\cmidrule(l){5-7}
Method \hfill $N=$ & 4 & 8 & 16 & 4 & 8 & 16 \\
\midrule
\multicolumn{7}{l}{\emph{Qwen3-VL-8B}} \\
Joint call & 74.2 & 77.8 & 80.6 & 73.2 & 70.6 & 68.0 \\
Calibrated joint call$^\dagger$ & 73.6 & 77.5 & 80.6 & 78.5 & 71.4 & 74.9 \\
OR of yes/no & 73.7 & 75.3 & 73.5 & 66.9 & 57.9 & 49.3 \\
$k$ of $N$ say yes$^\dagger$ & 73.7 & 75.3 & 75.2 & 66.9 & 57.9 & 63.8 \\
Sum rule on $p_i$$^\dagger$ & 74.3 & 76.7 & 76.4 & 73.0 & 63.9 & 65.5 \\
Five sampled answers$^\dagger$ & 73.8 & 76.4 & 76.1 & 68.8 & 60.9 & 66.1 \\
\rowcolor{oursbg}Calibrated bit$^\dagger$ & 73.9 & 78.1 & 82.3 & 81.0 & 75.9 & 77.7 \\
\rowcolor{oursbg}\method$^\dagger$ & 74.2 & 79.2 & 80.5 & 75.9 & 74.0 & 68.3 \\
\rowcolor{oursbg}\method, joint-call teacher & 74.5 & 78.0 & 80.6 & 72.0 & 68.4 & 68.4 \\
\midrule
\multicolumn{7}{l}{\emph{Qwen3-VL-4B}} \\
Joint call & 74.1 & 77.9 & 80.4 & 74.2 & 71.2 & 69.0 \\
Calibrated joint call$^\dagger$ & 74.4 & 77.6 & 80.9 & 79.1 & 75.7 & 75.9 \\
OR of yes/no & 75.1 & 75.7 & 73.8 & 68.1 & 57.7 & 49.5 \\
$k$ of $N$ say yes$^\dagger$ & 75.1 & 75.7 & 77.3 & 68.1 & 57.7 & 65.4 \\
Sum rule on $p_i$$^\dagger$ & 75.6 & 77.6 & 78.2 & 71.0 & 63.9 & 68.9 \\
Five sampled answers$^\dagger$ & 75.0 & 76.0 & 77.7 & 69.8 & 59.8 & 67.8 \\
\rowcolor{oursbg}Calibrated bit$^\dagger$ & 77.0 & 79.0 & 81.8 & 82.9 & 79.6 & 76.2 \\
\rowcolor{oursbg}\method$^\dagger$ & 75.7 & 78.8 & 80.2 & 83.2 & 76.5 & 71.8 \\
\rowcolor{oursbg}\method, joint-call teacher & 74.9 & 78.9 & 80.4 & 72.8 & 70.3 & 67.6 \\
\bottomrule
\end{tabular}
\end{table}

%% file: tables/s_samp.tex
\begin{table}[t]\centering\scriptsize\setlength{\tabcolsep}{2.4pt}
\caption{\textbf{Sampled answers do not replace probabilities} (Thor-PV, the 260 rooms of Table~1 of the main text). Five answers per agent sampled at temperature 0.7 for the first 16 cameras, and three sampled joint calls (8B). Balanced accuracy (\%); calls per question. $\dagger$: threshold fitted per $N$ on training rooms.}
\label{tab:s-samp}
\begin{tabular}{l c ccc ccc}
\toprule
 & & \multicolumn{3}{c}{Qwen3-VL-8B} & \multicolumn{3}{c}{Qwen3-VL-4B} \\
\cmidrule(lr){3-5}\cmidrule(l){6-8}
Method \hfill $N=$ & Calls & 4 & 8 & 16 & 4 & 8 & 16 \\
\midrule
Joint call & 1 & 93.5 & 94.8 & 95.6 & 95.3 & 96.9 & 97.4 \\
Joint call, majority of 3 samples & 3 & 93.3 & 95.5 & 96.0 & -- & -- & -- \\
\midrule
OR of yes/no & $N$ & 89.1 & 84.3 & 80.2 & 90.1 & 85.3 & 80.5 \\
$k$ of $N$ say yes$^\dagger$ & $N$ & 88.4 & 89.8 & 91.1 & 89.6 & 91.2 & 91.3 \\
\midrule
OR of majority answers & $5N$ & 88.9 & 83.4 & 79.1 & 89.7 & 84.7 & 79.7 \\
$k$ of $N$ on majority answers$^\dagger$ & $5N$ & 87.7 & 90.4 & 90.2 & 89.2 & 90.6 & 91.0 \\
Threshold on all $5N$ votes$^\dagger$ & $5N$ & 90.9 & 90.4 & 91.3 & 90.1 & 91.7 & 91.3 \\
Noisy-OR of yes shares$^\dagger$ & $5N$ & 91.6 & 90.4 & 92.5 & 90.9 & 92.0 & 92.9 \\
\midrule
\rowcolor{oursbg}Calibrated bit$^\dagger$ & $N$ & 94.6 & 95.7 & 94.9 & 94.8 & 96.4 & 96.3 \\
\rowcolor{oursbg}\method$^\dagger$ & $N$ & 94.6 & 95.7 & 96.0 & 95.6 & 95.6 & 96.7 \\
\bottomrule
\end{tabular}
\end{table}